\documentclass[10pt,twocolumn,letterpaper]{article}

\usepackage[pagenumbers]{cvpr} 

\definecolor{cvprblue}{rgb}{0.21,0.49,0.74}
\usepackage[pagebackref,breaklinks,colorlinks,allcolors=cvprblue]{hyperref}
\usepackage{algorithm}
\usepackage{algpseudocode}

\newcommand{\MOD}[1]{#1}

\usepackage{amsmath,amsfonts,bm,amssymb,mathtools,amsthm}
\usepackage{color,xcolor,placeins}
\usepackage{thm-restate}

\usepackage{hyperref}       
\hypersetup{
  colorlinks,
  linkcolor={blue!50!black},
  citecolor={blue!50!black},
  urlcolor={blue!80!black}
}
\definecolor{orange}{HTML}{ff6c0c}
\definecolor{blue}{HTML}{1f77b4}
\definecolor{Gray}{gray}{0.85}
\definecolor{LightCyan}{rgb}{0.88,1,1}

\usepackage{url}
\usepackage{xkcdcolors}
\usepackage{mdframed}

\usepackage{microtype}
\usepackage{graphicx}
\usepackage{subcaption}
\usepackage{latexsym}
\usepackage{sidecap}
\usepackage{caption}
\usepackage{booktabs} 
\usepackage{amsfonts}       
\usepackage{nicefrac}       
\usepackage{comment}
\usepackage{enumitem}
\usepackage{wrapfig,tikz}
\usepackage{longtable}
\usepackage{bigstrut, tabularx, multirow, makecell, diagbox}
\usepackage[export]{adjustbox}
\usepackage{float}
\usepackage{stackrel}
\usepackage{color}
\usepackage{colortbl}
\usepackage{theoremref}
\usepackage{cases}
\usepackage{stmaryrd}
\usepackage{mathabx}
\usepackage{dsfont}
\usepackage{arydshln}

\makeatletter
\usepackage{xspace}
\def\@onedot{\ifx\@let@token.\else.\null\fi\xspace}
\DeclareRobustCommand\onedot{\futurelet\@let@token\@onedot}

\definecolor{blue1}{RGB}{0,128,255}
\definecolor{blue3}{RGB}{0,0,128}
\definecolor{darkpastelgreen}{rgb}{0.01, 0.75, 0.24}
\definecolor{cerulean}{rgb}{0.0, 0.48, 0.65}

\def\eg{\emph{e.g}\onedot}

\def\ie{\emph{i.e}\onedot}

\def\vs{\emph{vs}\onedot}

\definecolor{darkgreen}{rgb}{0,0.6,0}

\newtheorem{prop}{Proposition}

\def\eqref#1{equation~\ref{#1}}

\def\1{\bm{1}}

\def\rvtheta{{\bm{\theta}}}

\def\rvc{{\mathbf{c}}}

\def\rvn{{\mathbf{n}}}

\def\rvp{{\mathbf{p}}}

\def\rvs{{\mathbf{s}}}

\def\rvw{{\mathbf{w}}}
\def\rvx{{\mathbf{x}}}
\def\rvy{{\mathbf{y}}}
\def\rvz{{\mathbf{z}}}

\def\vzero{{\bm{0}}}

\def\vtheta{{\bm{\theta}}}

\def\vc{{\bm{c}}}

\def\vs{{\bm{s}}}

\def\mF{{\bm{F}}}

\def\mI{{\bm{I}}}

\def\mP{{\bm{P}}}
\def\mQ{{\bm{Q}}}

\def\mS{{\bm{S}}}

\def\mSigma{{\bm{\Sigma}}}

\DeclareMathAlphabet{\mathsfit}{\encodingdefault}{\sfdefault}{m}{sl}
\SetMathAlphabet{\mathsfit}{bold}{\encodingdefault}{\sfdefault}{bx}{n}

\def\gA{{\mathcal{A}}}

\def\paperID{13917} 
\def\confName{CVPR}
\def\confYear{2025}

\title{Improving Diffusion Inverse Problem Solving with Decoupled Noise Annealing}

\author{
  Bingliang Zhang$^{*,1}$ \qquad\qquad Wenda Chu$^{*,1}$ \qquad\qquad Julius Berner$^{1}$\\ \,\, Chenlin Meng$^{2}$ \qquad\qquad Anima Anandkumar$^{1}$ \qquad \qquad Yang Song$^{3}$ \\
    $^{1}$California Institute of Technology\,\,\,\,\,\,\,\,$^{2}$Stanford University\,\,\,\,\,\,\,\,$^{3}$OpenAI\\
}

\begin{document}

\twocolumn[{%
\renewcommand\twocolumn[1][]{#1}%
\maketitle
\begin{center}
    \centering
    \captionsetup{type=figure}
    \includegraphics[width=\textwidth]{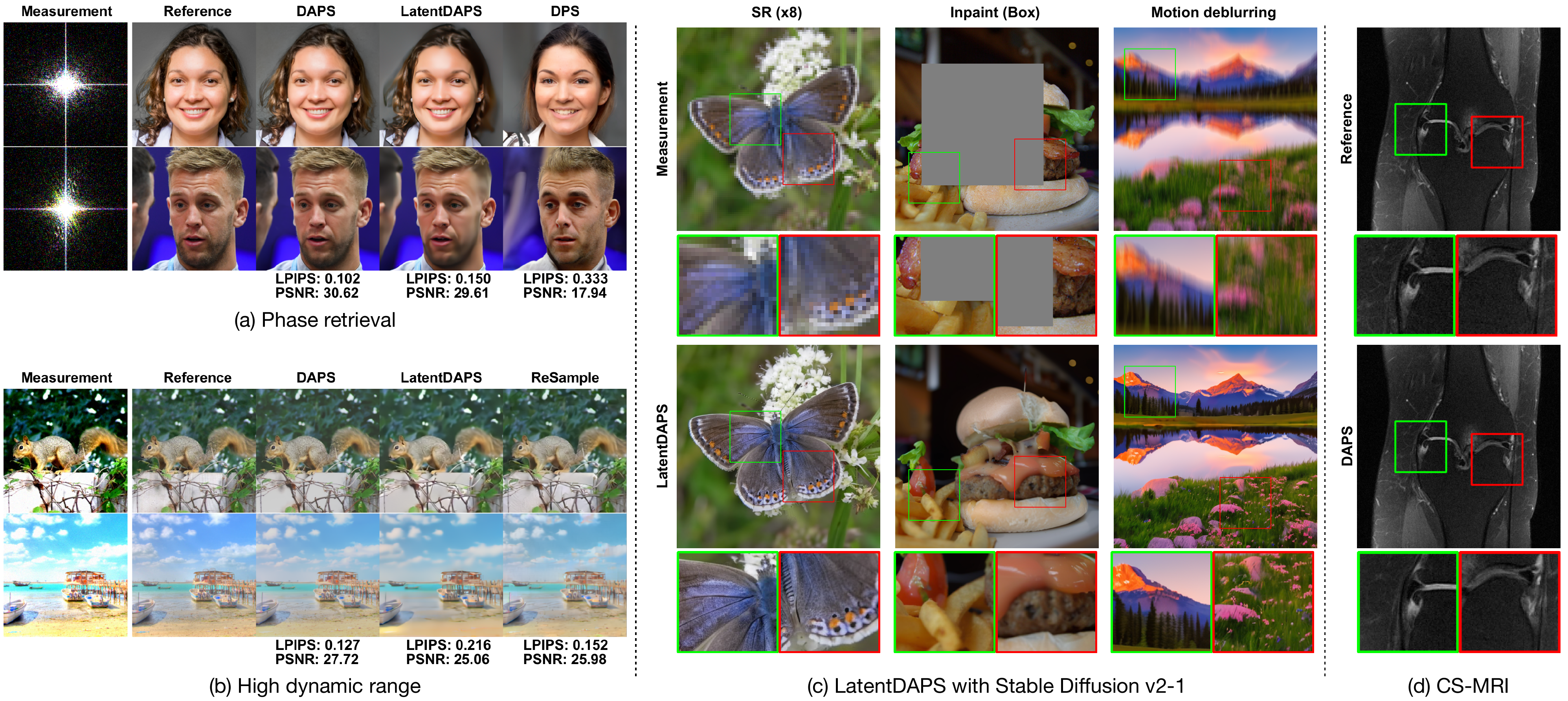}
    \captionof{figure}{\textbf{Overview of Decoupled Annealing Posterior Sampling (DAPS).} Our method provides a flexible and effective framework for solving inverse problems through a decoupled posterior sampling process. In (a)(b), we present DAPS visual results on FFHQ and ImageNet at a resolution of 256, and in (c), on natural images at a resolution of 768. In (d), we display DAPS results on compressed sensing multi-coil MRI (CS-MRI). DAPS effectively addresses nonlinear inverse problems as well as medical imaging MRI challenges. Additionally, DAPS can be enhanced using large-scale latent diffusion models (LDMs) \cite{rombach2022high}, as shown in (c).} 
     \label{main-comparison}
\end{center}
}]
\def\thefootnote{*}\footnotetext{These authors contributed equally to this work}\def\thefootnote{\arabic{footnote}}
\begin{abstract}
Diffusion models have recently achieved success in solving Bayesian inverse problems with learned data priors. Current methods build on top of the diffusion sampling process, where each denoising step makes small modifications to samples from the previous step. However, this process struggles to correct errors from earlier sampling steps, leading to worse performance in complicated nonlinear inverse problems, such as phase retrieval. To address this challenge, we propose a new method called Decoupled Annealing Posterior Sampling (DAPS) that relies on a novel noise annealing process. Specifically, we decouple consecutive steps in a diffusion sampling trajectory, allowing them to vary considerably from one another while ensuring their time-marginals anneal to the true posterior as we reduce noise levels. This approach enables the exploration of a larger solution space, improving the success rate for accurate reconstructions. We demonstrate that DAPS significantly improves sample quality and stability across multiple image restoration tasks, particularly in complicated nonlinear inverse problems. Our code is available at the GitHub repository \href{https://github.com/zhangbingliang2019/DAPS}{DAPS}.
\end{abstract}    
\vspace{-10pt}
\section{Introduction}\label{sec:intro}

Inverse problems are ubiquitous in science and engineering, with applications ranging from image restoration~\cite{song2021scorebased,kawar2022denoising,chung2023diffusion,saharia2023Image, lugmayr2022repaint,zhu2023denoising}, medical imaging~\cite{jalal2021robust,song2022solving,chung2022score, hung2023med, dorjsembe2024conditional,chung2022mr} to astrophotography~\cite{akiyama2019first,feng2023score,sun2020deep,sun2023provable}. Solving an inverse problem involves finding the underlying signal $\rvx_0$ from its partial, noisy measurement $\rvy$. Since the measurement process is typically noisy and many-to-one, inverse problems do not have a unique solution; instead, multiple solutions may exist that are consistent with the observed measurement. In the Bayesian inverse problem framework, the solution space is characterized by the posterior distribution $p(\rvx_0\mid \rvy) \propto p(\rvy \mid \rvx_0) p(\rvx_0)$, where $p(\rvy \mid \rvx_0)$ represents the noisy measurement process, and $p(\rvx_0)$ is the prior distribution. In this work, we aim to solve Bayesian inverse problems where the measurement process $p(\rvy \mid \rvx_0)$ is known, and the prior distribution $p(\rvx_0)$ is captured by a deep generative model trained on a corresponding dataset.

As score-based diffusion models~\cite{song2019generative,song2020improved,ho2020denoising,song2021scorebased, karras2022elucidating,karras2024analyzing} have risen to dominance in modeling high-dimensional data distributions like images, audio, and video, they have become the leading method for estimating the prior distribution $p(\rvx_0)$ in Bayesian inverse problems. A diffusion model generates a sample $\rvx_0$ by smoothly removing noise from an unstructured initial noise sample $\rvx_T$ through solving stochastic differential equations (SDEs). In particular, each step of the sampling process recursively converts a noisy sample $\rvx_{t+\Delta t}$, where $\Delta t> 0$ denotes the step size, to a slightly less noisy sample $\rvx_{t}$ until $t=0$. This iterative structure in the diffusion sampling process can be leveraged to facilitate Bayesian inverse problem solving. In fact, prior research~\cite{chung2023diffusion,song2023pseudoinverseguided,boys2023tweedie} has shown that given measurement $\rvy$ and a diffusion model prior $p(\rvx_0)$, we can sample from the posterior distribution $p(\rvx_0 \mid \rvy)$ by perturbing a reverse-time SDE using an approximate gradient of the measurement process, $\nabla_{\rvx_t} \log p(\rvy \mid \rvx_t)$, at every step of the SDE solver.

Despite the remarkable success of these methods in solving many real-world inverse problems, like image colorization~\cite{song2021scorebased,kawar2022denoising,liu2023improved}, image super-resolution~\cite{song2021scorebased,kawar2022denoising,chung2023diffusion,song2023pseudoinverseguided}, computed tomography~\cite{song2022solving,song2024solving}, and magnetic resonance imaging~\cite{song2022solving,jalal2021robust}, they face significant challenges in more complex inverse problems with nonlinear measurement processes, such as phase retrieval and nonlinear motion deblurring. This is partially because, in prior methods, each denoising step approximately samples from the distribution $p(\rvx_t \mid \rvx_{t + \Delta t}, \rvy)$. This causes $\rvx_t$ and $\rvx_{t+\Delta t}$ to be close to each other because of using a small step size $\Delta t$ in discretizing the reverse-time SDE. As a result, $\rvx_t$ can at most correct local errors in $\rvx_{t+\Delta t}$ but struggles to correct global errors that require significant modifications to the prior sample. This issue is exacerbated when the methods are applied to complicated, nonlinear inverse problems, such as phase retrieval, where they often converge to undesired samples that are consistent with the measurement but reside in low-probability areas of the prior distribution.

\begin{figure}[t]
    \centering
    \includegraphics[width=\linewidth]{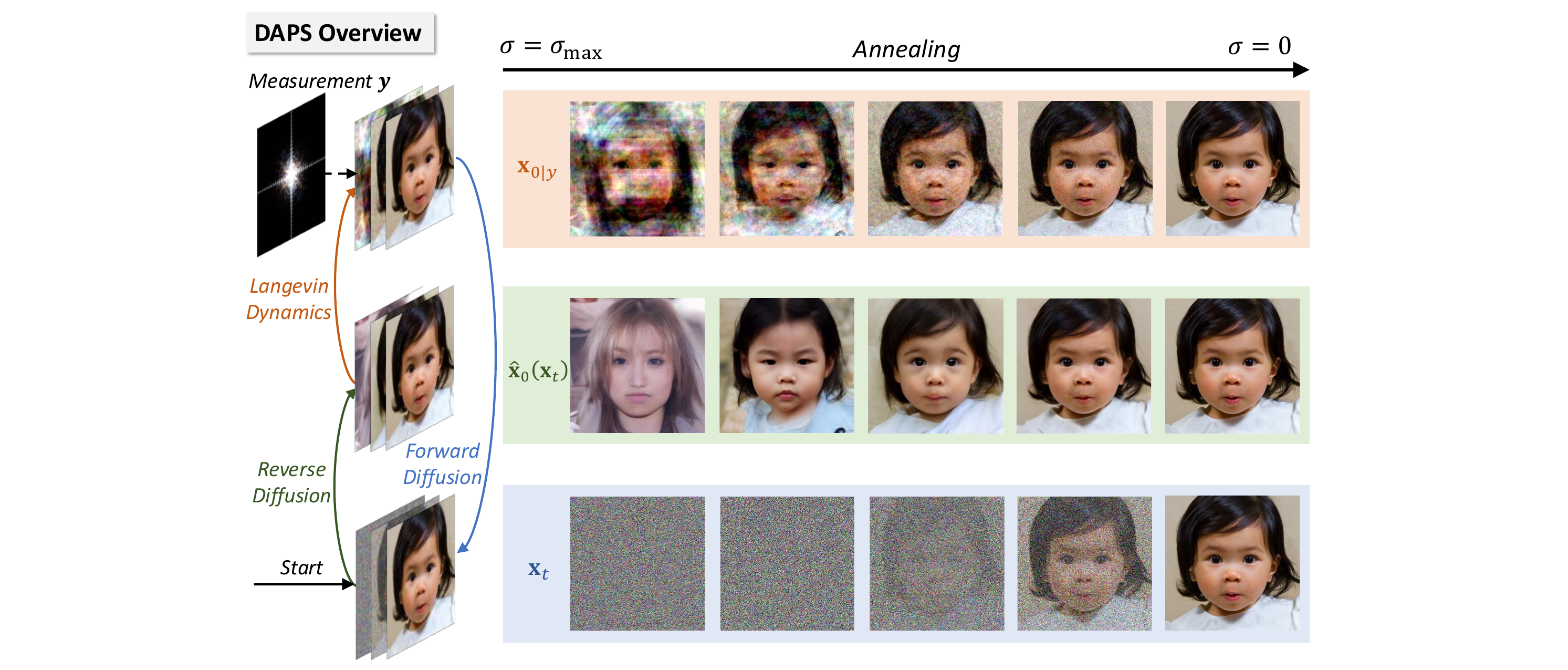}
    \caption{\textbf{An illustration of our method on phase retrieval}. Given a noisy sample $\rvx_t$, we first solve the reverse diffusion process to obtain $\hat \rvx_0( \rvx_t)$. Using this, we construct an approximation for $p(\rvx_0\mid\rvx_t,\rvy)$. Next, we sample $\rvx_{0\mid y} \sim p(\rvx_0\mid\rvx_t,\rvy)$ with multiple steps of Langevin dynamics, then perturb it with the forward diffusion process to obtain a sample with slightly less noise. We repeat this process until the noise reduces to zero.}
    \label{fig:method}
    \vspace{-10pt}
\end{figure}

To address this challenge, we propose a new framework for solving general inverse problems, termed \textbf{D}ecoupled \textbf{A}nnealing \textbf{P}osterior \textbf{S}ampling (DAPS). Our method employs a new noise annealing process inspired by the diffusion sampling process, where we decouple the consecutive samples $\rvx_{t+\Delta t}$ and $\rvx_t$ in the sampling trajectory, allowing samplers to correct large, global errors made in earlier steps. Instead of repetitively sampling from $p(\rvx_t \mid \rvx_{t + \Delta t}, \rvy)$ as in previous methods, which restricts the distances between consecutive samples $\rvx_{t+\Delta t}$ and $\rvx_t$, DAPS recursively samples from the marginal distribution $p(\rvx_t \mid \rvy)$. As illustrated in \cref{fig:method}, we factorize the time marginal $p(\rvx_t \mid \rvy)$ into three conditional distributions and sample from them in turn by solving the reverse diffusion process, simulating Langevin dynamics, and adding noise according to the forward diffusion process. We show that this creates approximate samples from corresponding marginal distributions. As the noise gradually anneals to zero, the time marginal $p(\rvx_t \mid \rvy)$ smoothly converges to the posterior distribution $p(\rvx_0 \mid \rvy)$, providing samples that approximately solve the Bayesian inverse problem.

Empirically, our method demonstrates significantly improved performance on various inverse problems compared to existing approaches, \MOD{ranging from image restoration to compressed sensing MRI.}
Our method can be combined with diffusion models in both raw pixel space and learned latent space, which we refer to as DAPS and LatentDAPS, respectively. As shown in \cref{main-comparison}, both methods provide superior reconstructions with improved visual perceptual quality across a wide range of nonlinear inverse problems. Our approach exhibits remarkable stability and sampling quality, particularly for challenging nonlinear inverse problems. On the FFHQ 256 dataset, we achieve PSNR values of 30.72 dB and 27.12 dB for the phase retrieval and high dynamic range tasks, representing improvements of 1.98 dB and 4.39 dB over the previous state-of-the-art results, respectively.
\MOD{We apply DAPS to large-scale text-conditioned latent diffusion models, showing its scalability to high-resolution image restoration tasks. Beyond image restoration tasks, DAPS achieves a PSNR of 31.49 dB, showing a 2.70 dB improvement over previous diffusion-based methods. We also demonstrate how DAPS can be applied to inverse problems on categorical data with pre-trained discrete diffusion models.} Additionally, DAPS performs well with a small number of neural network evaluations (approximately 100), striking a better balance between efficiency and sample quality.

\section{Background}
\subsection{Diffusion Models}
Diffusion models \cite{song2019generative,song2020improved,ho2020denoising,song2021scorebased,karras2022elucidating} generate data by reversing a predefined noising process. Let the data distribution be $p(\rvx_0)$. We can define a series of noisy data distributions $p(\rvx;\sigma)$ by adding Gaussian noise with a standard deviation of $\sigma$ to the data. These form the time-marginals of a stochastic differential equation (SDE) \cite{karras2022elucidating}, given by
    $\mathrm d \rvx_t = \sqrt{2\dot\sigma_t\sigma_t} \mathrm d\rvw_{t}$,
where $\sigma_t$ is the predefined noise schedule with $\sigma_0= 0$ and $\sigma_T = \sigma_{\max}$, $\dot \sigma_t$ is the time derivative of $\sigma_t$, and $\rvw_t$ is a standard Wiener process. We use $\rvx_t$ interchangeably with $\rvx_{\sigma_t}$. For a sufficiently large $\sigma_{\max}$, the distribution $p(\rvx;\sigma_{\max})$ converges to pure Gaussian noise $\mathcal N(\vzero, \sigma_{\max}^2 \mI)$. 

To sample from data distribution $p(\rvx_0)$, we first draw an initial sample from $\mathcal N(\vzero, \sigma_{\max}^2 \mI)$, then solve the reverse SDE:
    $\mathrm d\rvx_t = -2\dot\sigma_t\sigma_t\nabla_{\rvx_t}\log p(\rvx_t;\sigma_t)\mathrm dt + \sqrt{2\dot \sigma_t\sigma_t}\mathrm d\rvw_t,$
where $\nabla_{\rvx_t} \log p(\rvx_t; \sigma_t)$ is the time-dependent score function \cite{song2019generative,song2021scorebased}. Here $\nabla_{\rvx_t}\log p(\rvx_t;\sigma_t)$ is unknown, but can be approximated by training a diffusion model $\vs_\vtheta(\rvx_t,\sigma_t)$ such that $\vs_\vtheta(\rvx_t,\sigma_t) \approx \nabla_{\rvx_t}\log p(\rvx_t;\sigma_t)$. 

\subsection{Bayesian Inverse Problems with Diffusion}\label{sec:bip}
Inverse problems aim to recover data from partial, potentially noisy measurements. Formally, solving an inverse problem involves finding the inversion to a forward model that describes the measurement process. In general, a forward model takes the form of
   $ \rvy = \mathcal A(\rvx_0) + \rvn,$
where $\mathcal A$ is the measurement function, $\rvx_0$ represents the original data, $\rvy$ is the observed measurement, and $\rvn$ symbolizes the noise in the measurement process, often modeled as $\rvn\sim\mathcal N(\vzero, \beta_{\rvy}^2\mI)$. In a Bayesian framework, $\rvx_0$ comes from the posterior distribution \(p(\rvx_0\mid\rvy) \propto p(\rvx_0) p(\rvy \mid \rvx_0)\). Here $p(\rvx_0)$ is a prior distribution that can be estimated from a given dataset, and $p(\rvy \mid \rvx_0) = \mathcal{N}(\mathcal A (\rvx_0), \beta_{\rvy}^2 \mI)$ models the noisy measurement process.

When the prior $p(\rvx_0)$ is modeled by a pre-trained diffusion model, we can modify reverse SDE to approximately sample from the posterior distribution following Bayes' rule,

\vspace{-15pt}
\begin{align}
    \mathrm d\rvx_t = & -2\dot\sigma_t\sigma_t \nabla_{\rvx_t}\log p(\rvx_t;\sigma_t)\mathrm dt \nonumber\\ 
    & \hspace{-10pt} -2\dot\sigma_t\sigma_t  \nabla_{\rvx_t} \log p(\rvy\mid\rvx_t)\mathrm dt  +\sqrt{2\dot \sigma_t\sigma_t}\mathrm d \rvw_t.\label{eq:rev-sde}
\end{align}
Here, the noisy likelihood $\nabla_{\rvx_t} \log p(\rvy\mid\rvx_t)$ is generally intractable. Multiple methods have been proposed to estimate the noisy likelihood \cite{song2022solving,chung2022improving,song2023pseudoinverseguided,boys2023tweedie,meng2022diffusion,wang2022zero,cardoso2023monte,zhu2023denoising, rout2023firstordertweediesolvinginverse}. One predominant approach is the DPS algorithm \cite{chung2023diffusion}, which estimates $p(\rvy\mid\rvx_t) \approx p(\rvy\mid\rvx_0=\mathbb E[\rvx_0\mid\rvx_t])$. Another line of work \cite{kawar2022denoising,kawar2021snips,wang2022zero, cardoso2023montecarloguideddiffusion, dou2024diffusion} solves linear inverse problems by running the reverse diffusion in the spectral domain via singular value decomposition (SVD). Other methods bypass direct computation of this likelihood by interleaving optimization \cite{song2024solving,kamilov2023plugandplay,chen2023alternating,arvinte2021deep,zhu2023denoising, li2024decoupleddataconsistencydiffusion, wu2024principled,xu2024provably} or projection \cite{kawar2022denoising,chung2022score,jalal2021robust,wang2022zero} steps with normal diffusion sampling steps.

Despite promising empirical success, we find that this line of approaches faces challenges in solving more difficult inverse problems when the forward model is highly nonlinear. Accurately solving the reverse SDE in \cref{eq:rev-sde} requires the solver to take a very small step size $\Delta t > 0$, causing $\rvx_t$ and $\rvx_{t+\Delta t}$ to be very close to each other. Consequently, \(\rvx_t\) can only correct minor errors in \(\rvx_{t+\Delta t}\), but oftentimes fails to address larger, global errors that require substantial changes to $\rvx_{t+\Delta t}$. One such failure case is given in \cref{fig:syn}.

\section{Method}
\label{sec:method}
\vspace{-5pt}
\begin{figure}[t]
    \centering
    \begin{subfigure}[b]{.40\textwidth}
        \includegraphics[width=\textwidth]{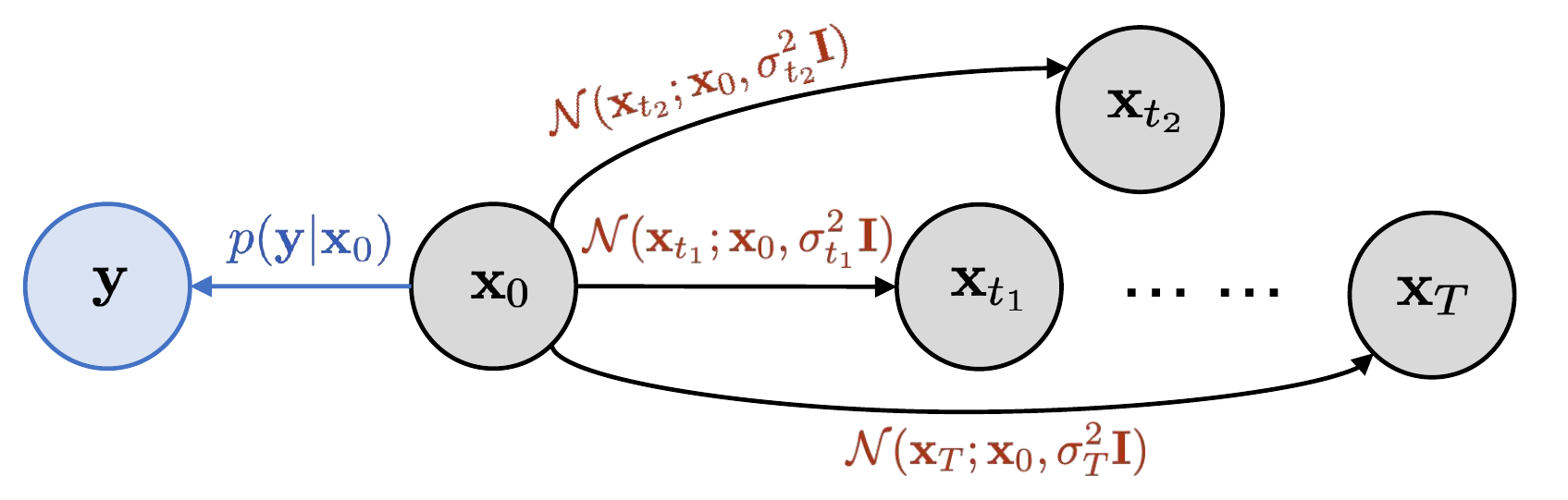}
    \caption{Diffusion models in pixel spaces.}
    \label{fig:factor_graph}
    \end{subfigure}
     \begin{subfigure}[b]{.40\textwidth}
        \centering
    \includegraphics[width=\textwidth]{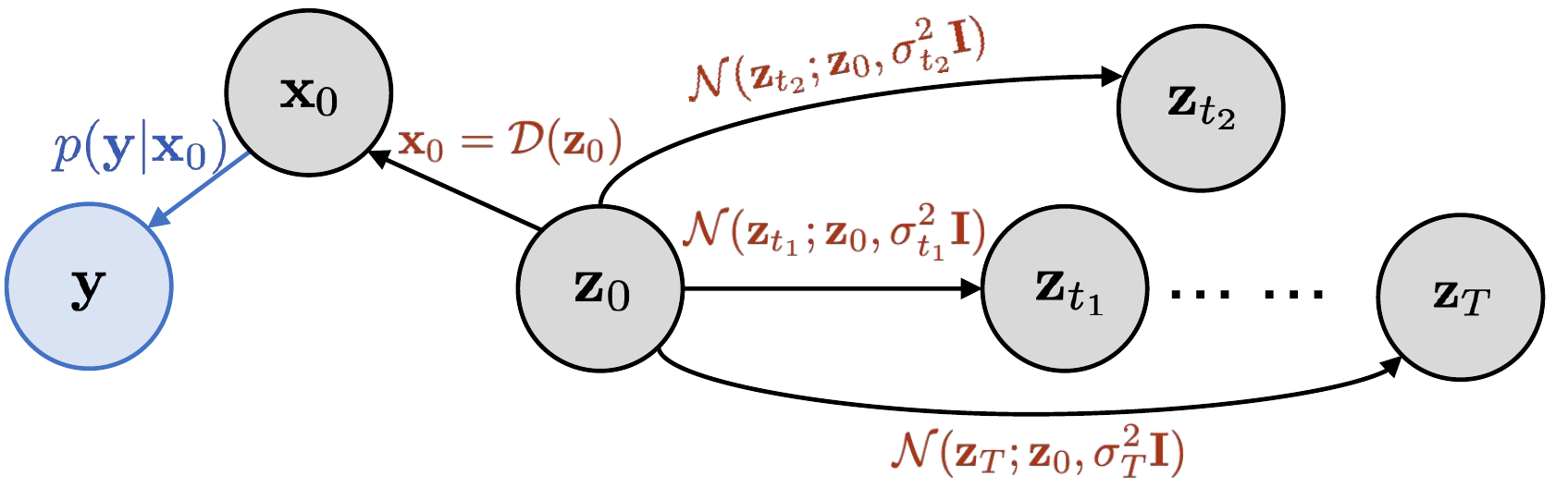}
    \caption{Latent diffusion models.}
    \label{fig:factor_graph_latent}
    \end{subfigure}
    \vspace{-8pt}
    \caption{\textbf{Probabilistic graphical models} of the decoupled noise annealing process in (a) pixel and (b) latent spaces respectively. Here $\rvx_{t}\sim p(\rvx_{t};\sigma_t)$, and $\rvy$ is the observed measurement. We factorize $p(\rvx_{t}\mid\rvy)$ based on these probabilistic graphs.}
    \vspace{-13pt}
\end{figure}

\subsection{Decoupled Annealing Posterior Sampling}
Instead of solving the reverse-time SDE in \cref{eq:rev-sde}, we propose a new noise annealing process that reduces the dependency between samples at consecutive time steps, as illustrated in \cref{fig:method,fig:factor_graph,fig:factor_graph_latent}. Unlike previous methods, we ensure $\rvx_{t}$ and $\rvx_{t + \Delta t}$ are conditionally independent given $\rvx_0$. To generate sample $\rvx_t$ from $\rvx_{t+\Delta t}$, we follow a two-step procedure: (1) sampling $\rvx_{0 \mid \rvy} \sim p(\rvx_0 \mid \rvx_{t+\Delta t}, \rvy)$, and (2) sampling $\rvx_{t} \sim \mathcal{N}(\rvx_{0 \mid \rvy}, \sigma_t^2 \mI)$. We repeat this process, gradually reducing noise until $\rvx_0$ is sampled. We call this process \textit{decoupled noise annealing}, which is justified by the proposition below.
\vspace{-5pt}
\begin{prop}
\label{prop:1}
    Suppose $\rvx_{t_1}$ is sampled from the  time-marginal $p(\rvx_{t_1}\mid\rvy)$, then
    \begin{equation}
        \rvx_{t_2} \sim \mathbb E_{\rvx_0 \sim p(\rvx_0\mid \rvx_{t_1}, \rvy)}[\mathcal N( \rvx_0, \sigma_{t_2}^2 \mI)]\label{eq:prop}
    \end{equation}
    satisfies the  time-marginal $p(\rvx_{t_2}\mid \rvy)$.
\end{prop}

\textbf{Remark.} The key idea of Proposition~\ref{prop:1} is to enable the sampling from $p(\rvx_{t_2}\mid\rvy)$ for any noise level $\sigma_{t_2}$ given any sample $x_{t_1}$ at another noise level $\sigma_{t_1}$. For a sufficiently large $\sigma_T$, one can assume $p(\rvx_{T}\mid \rvy)\approx p(
\rvx_{T};\sigma_T)\approx\mathcal N(\bm{0}, \sigma_T^2\mI)$. Starting from $\rvx_{T}$, we can iteratively sample from $p(\rvx_t\mid\rvy)$ with $\sigma_t$ annealed down from $\sigma_T$ to 0.    

The first step of our decoupled noise annealing requires sampling $\rvx_{0\mid \rvy} \sim p(\rvx_0 \mid \rvx_{t}, \rvy)$ where $\rvx_{t}$ and $\rvy$ are known. Since $\rvy$ is conditionally independent from $\rvx_t$ given $\rvx_0$, we can deduce from the Bayes' rule that
\begin{align}
    p(\rvx_0 \mid \rvx_t, \rvy) & = \frac{p(\rvx_0 \mid \rvx_t) p(\rvy \mid \rvx_0, \rvx_t)}{p(\rvy \mid \rvx_t)} \nonumber \\
    \ \  & \propto \ p(\rvx_0 \mid \rvx_t) p(\rvy \mid \rvx_0).
\end{align}
To sample $\rvx_{0\mid\rvy}$ from this unnormalized distribution, \MOD{MCMC methods such as Langevin dynamics~\cite{welling2011bayesian} and Hamiltonian Monte Carlo~\cite{betancourt2015hamiltonian} can be adopted (as explained in~\cref{appendix:hmc}). The updating rule of Langevin dynamics~\cite{welling2011bayesian} is given by} 

\vspace{-15pt}
\begin{align}
\label{eq:langevin}
    \rvx_0^{(j+1)} &= \rvx_0^{(j)} + \eta \nabla_{\rvx_0^{(j)}} \log p(\rvx_0^{(j)} \mid \rvx_t) \nonumber \\
    & + \eta \nabla_{\rvx_0^{(j)}} \log p(\rvy \mid \rvx_0^{(j)}) + \sqrt{2\eta} \bm{\epsilon}_j,
\end{align}
where $\eta > 0$ is the step size and $\bm{\epsilon}_j \sim \mathcal{N}(\vzero, \mI)$. When $\eta \to 0$ and $j \to \infty$, the sample $\rvx_0^{(j)}$ will be approximately distributed according to $p(\rvx_{0} \mid \rvx_t, \rvy)$.

\subsection{Practical Design Choice of DAPS}\label{sec:3.2}

\MOD{Following \cref{eq:langevin}, we discuss different ways to approximate the conditional distribution $p(\rvx_0\mid \rvx_t)$ due to its intractability. A line of previous works on diffusion posterior sampling \cite{chung2023diffusion,song2023pseudoinverseguided,boys2023tweedie} approximate $p(\rvx_0 \mid \rvx_t)$ by a Gaussian distribution,
\begin{equation}\label{eq:gaussian-approx}
    p(\rvx_0 \mid \rvx_t) \approx \mathcal{N}(\rvx_0; \hat{\rvx}_0(\rvx_t), r_t^2 \mI), 
\end{equation}
where $\hat{\rvx}_0(\rvx_t)$ is an estimator of $\rvx_0$ given $\rvx_t$ typically specified as $\mathbb E[\rvx_0\mid \rvx_t]$, and the variance $r_{t}^2$ is specified using heuristics.

Another possible way to estimate $p(\rvx_0\mid\rvx_t)$ is to decompose it via Bayes' rule, and substitute the intractable $p(\rvx_0)$ by the distribution $p(\rvx_0; \sigma_{t_{\min}})$ with a small Gaussian noise for numerical stability, i.e.,
\begin{align}
    \nabla_{\rvx_0}\log p(\rvx_0\mid \rvx_t) &= \nabla_{\rvx_0} \log p(\rvx_t \mid \rvx_0) + \nabla_{\rvx_0}\log p(\rvx_0)\nonumber \\
    & \hspace{-2em}\approx \nabla_{\rvx_0}\log p(\rvx_t \mid \rvx_0) + \vs_\vtheta(\rvx_0, t_{\min}).\label{eq:score_estimate}
\end{align}
Despite being a more accurate estimation, the \textit{diffusion-score estimation} \cref{eq:score_estimate} is much more expensive than using a \textit{Gaussian approximation} when applied to \cref{eq:langevin}, as it requires calling the score function multiple times. Empirical studies in \cref{sec:abla} show that using a \textit{Gaussian approximation} can achieve comparable results as the \textit{diffusion-score estimation}, but is significantly more time-efficient. In practice, we adopt the \textit{Gaussian approximation} and compute $\hat{\rvx}_0(\rvx_t)$ by solving the (unconditional) probability flow ODE starting at $\rvx_t$. We leave details in \cref{appendix:ode-solver}.}   

When the measurement noise is an isotropic Gaussian, \ie, $\rvn \sim \mathcal{N}(\vzero, \beta_\rvy^2 \mI)$, the update rule simplifies to:

\vspace{-10pt}
\begin{align}
\label{eq:langevin_gaussian}
    \rvx_0^{(j+1)} &= \rvx_0^{(j)} - \eta \nabla_{\rvx_0^{(j)}} \frac{\|\rvx_0^{(j)} - \hat{\rvx}_0(\rvx_t)\|^2}{2r_t^2} \nonumber \\
    &- \eta \nabla_{\rvx_0^{(j)}}\frac{\|\gA(\rvx_0^{(j)}) - \rvy\|^2}{2\beta_\rvy^2}  + \sqrt{2\eta} \bm{\epsilon}_j,
\end{align}
Combining \cref{eq:langevin_gaussian} with Proposition~\ref{prop:1}, we implement Decoupled Annealing Posterior Sampling (DAPS) as \cref{alg1} in \cref{appendix:pixel}. In particular, given a noise schedule $\sigma_t$ and a time discretization $\{t_i, i = 0, \dots, N_A\}$, we iteratively sample $\rvx_{t_i}$ from the measurement-conditioned time-marginal $p(\rvx_{t_i} \mid \rvy)$ for $i = N_A, N_A-1, \dots, 0$, following \cref{eq:prop} in Proposition~\ref{prop:1}. This algorithm provides an approximate sample $\rvx_0$ from the posterior distribution $p(\rvx_0 \mid \rvy)$.

The computational cost of Langevin dynamics mostly originates from evaluating the measurement function $\mathcal{A}$. In most image restoration tasks, this operation is significantly more efficient compared to evaluating the diffusion model. As demonstrated in \cref{appendix:efficiency}, the MCMC sampling introduces only a small overhead to the sampling process. 

\subsection{DAPS with Latent Diffusion Model}
\MOD{
Given a pretrained encoder $\mathcal E$ and decoder $\mathcal D$, the latent diffusion models (LDMs) \cite{rombach2022high} are trained to fit the latent distribution $p(\rvz_0)$, where $\rvz_0$ is given by the encoder $\rvz_0=\mathcal E(\rvx_0)$. One can recover $\rvx_0$ from the decoder $\rvx_0=\mathcal D(\rvz_0)$. We found the proposed method can be adapted to sampling with pre-trained latent diffusion models by factorizing the probabilistic graphical model for the latent diffusion process, as shown in \cref{fig:factor_graph_latent}. We refer to this as LatentDAPS in the following paper. 

Accordingly, we can derive the following Langevin dynamics updating rule from \cref{eq:langevin}:

\vspace{-17pt}
\begin{align}
\label{eq:langevin-latent}
    \rvz_0^{(j+1)} &= \rvz_0^{(j)} + \eta \nabla_{\rvz_0^{(j)}} \log p(\rvz_0^{(j)} \mid \rvz_t) \nonumber \\[-3pt]
    & + \eta \nabla_{\rvz_0^{(j)}} \log p(\rvy \mid \mathcal D(\rvz_0^{(j)})) + \sqrt{2\eta} \bm{\epsilon}_j.
\end{align}
\vspace{-12pt}

While LatentDAPS is a straightforward and natural extension of DAPS compared to recent latent diffusion inverse problem solvers \cite{rout2023solving, rout2024beyond, song2024solving}, it demonstrates comparable or even superior performance across several tasks in \cref{tab:results} and \cref{tab:sd-quant}. More detailed derivation and proofs are shown in \cref{appendix:latent}.}

\subsection{Discussion with Existing Methods}

\begin{figure}
    \centering
    \includegraphics[width=.44\textwidth]{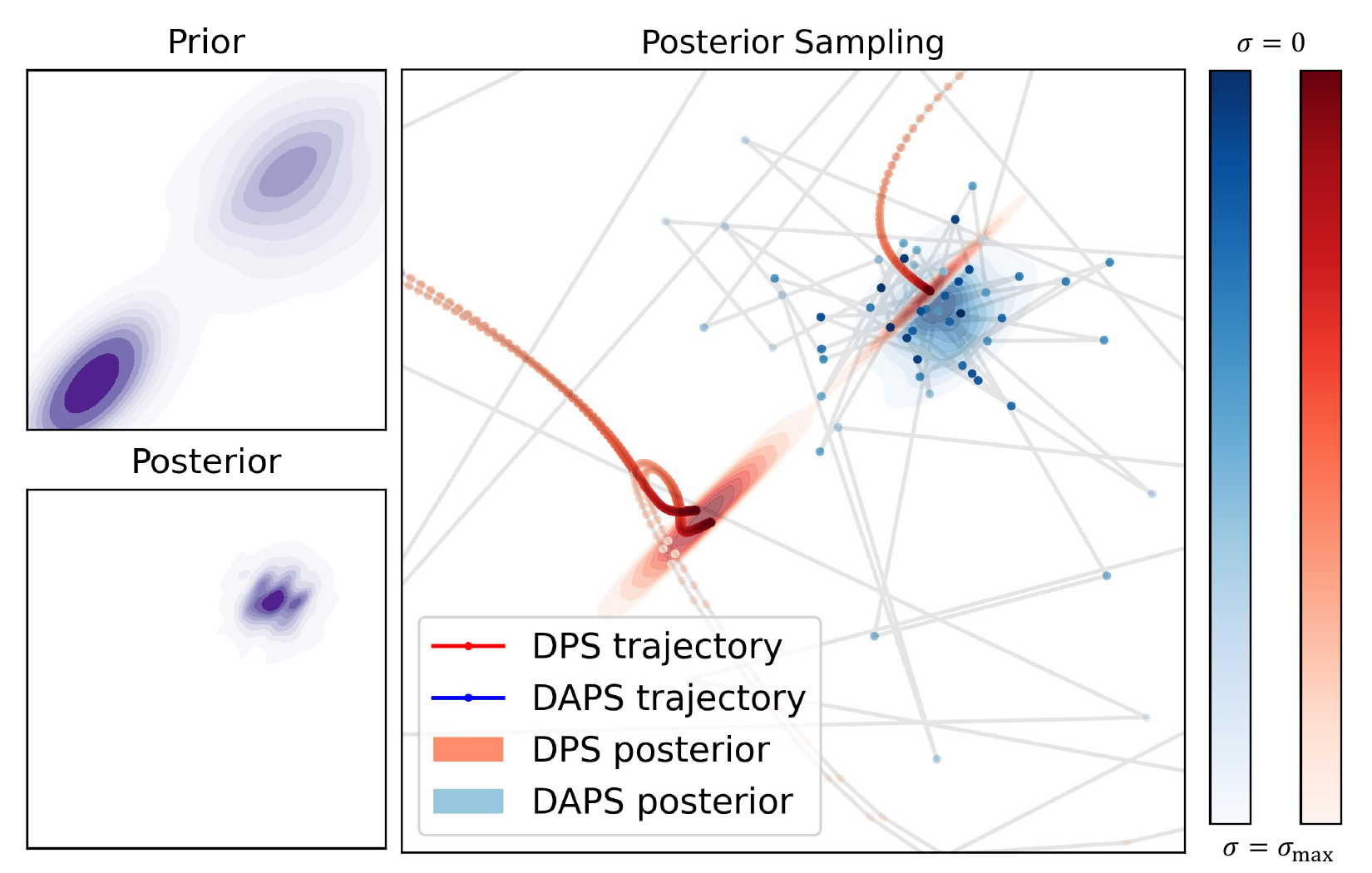}
    \caption{\textbf{DAPS vs. DPS on 2D synthetic data.} Consecutive sampling steps are close to each other for DPS but not for DAPS. Here DAPS approximates the posterior better than DPS.}
    \label{fig:syn}
    \vspace{-10pt}
\end{figure}

\textbf{Comparison with other posterior sampling methods.}
Unlike most previous algorithms, our sampling method does not solve a specific SDE/ODE; instead, we recursively sample from the time-marginal $p(\rvx_t\mid\rvy)$ with noise annealing to zero. We therefore decouple the dependency of $\rvx_t$ and $\rvx_{t+\Delta t}$ in the sampling process. We argue that decoupling helps correct the errors accumulated in early diffusion sampling steps by allowing non-local transitions. This is particularly important when the measurement function is nonlinear. As a concrete example, \cref{fig:syn} compares DAPS and DPS when solving a 2D nonlinear inverse problem with a Gaussian mixture prior. We visualize the trajectories of $\rvx_t$ from time $T$ to $0$ for both methods. For DAPS, points on the trajectory have significantly larger variations compared to those of DPS. As a result, DPS converges to wrong solutions, but DAPS is able to approximate the true posterior distribution. We provide more discussions in \cref{appendix:syn}.

\noindent\textbf{Comparison with optimization-based methods.} Although our method does not directly involve inner-loop optimization, we highlight that it connects with existing optimization-based solvers for inverse problems. For example, ReSample~\cite{song2024solving}, DiffPIR~\cite{zhu2023denoising}, and DCDP~\cite{li2024decoupleddataconsistencydiffusion} alternate between denoising optimizing, and resampling to solve inverse problems. However, these methods solve the MAP estimation problems, while DAPS draws samples from the posterior distribution. In particular, instantiating DAPS with Langevin dynamics and \textit{Gaussian approximation} resembles ~\cite{song2024solving} if we set standard deviation $\beta_\rvy \to 0$ and assume $\eta\beta_\rvy^2$ is constant in \cref{eq:latent}.

\section{Experiments}
\label{sec:experiment}

\subsection{Experimental Setup}
We evaluate our method using both pixel-space and latent diffusion models. For pixel-based diffusion experiments, we leverage the pre-trained diffusion models trained by \cite{chung2023diffusion} on the FFHQ dataset and the pre-trained model from \cite{dhariwal2021diffusion} on the ImageNet dataset. For latent diffusion models, we use the same pre-trained models as \cite{song2024solving}: the unconditional LDM-VQ4 trained on FFHQ and ImageNet by \cite{rombach2022high}. The autoencoder's downsample factor is $4$. We use the same time step discretization and noise schedule as EDM \cite{karras2022elucidating}.

\begin{figure*}
    \centering
    \begin{subfigure}[b]{.45\textwidth}
        \includegraphics[width=\textwidth]{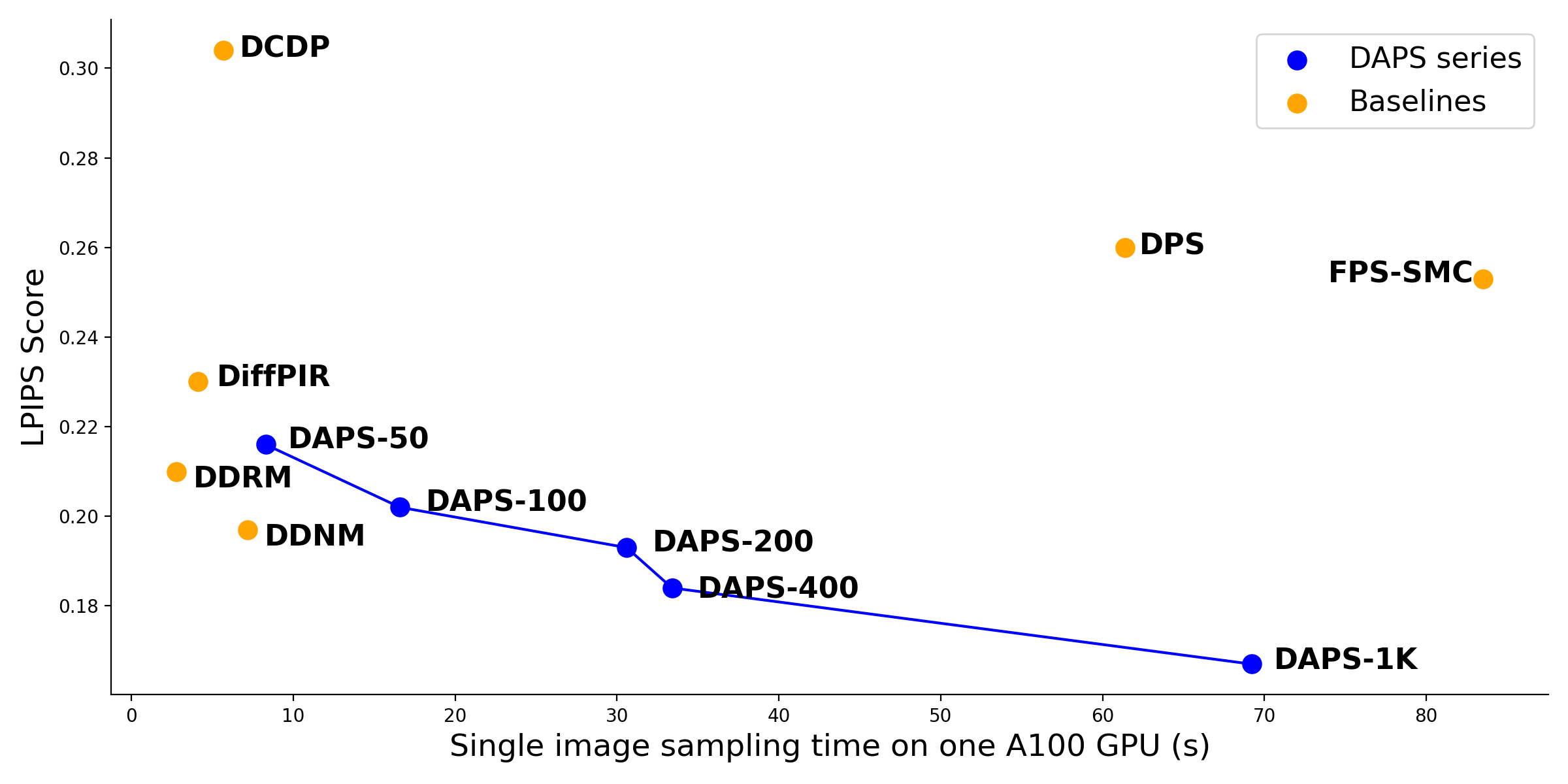}
    \caption{Gaussian deblur}
    \end{subfigure}
    \begin{subfigure}[b]{.45\textwidth}
        \includegraphics[width=\textwidth]{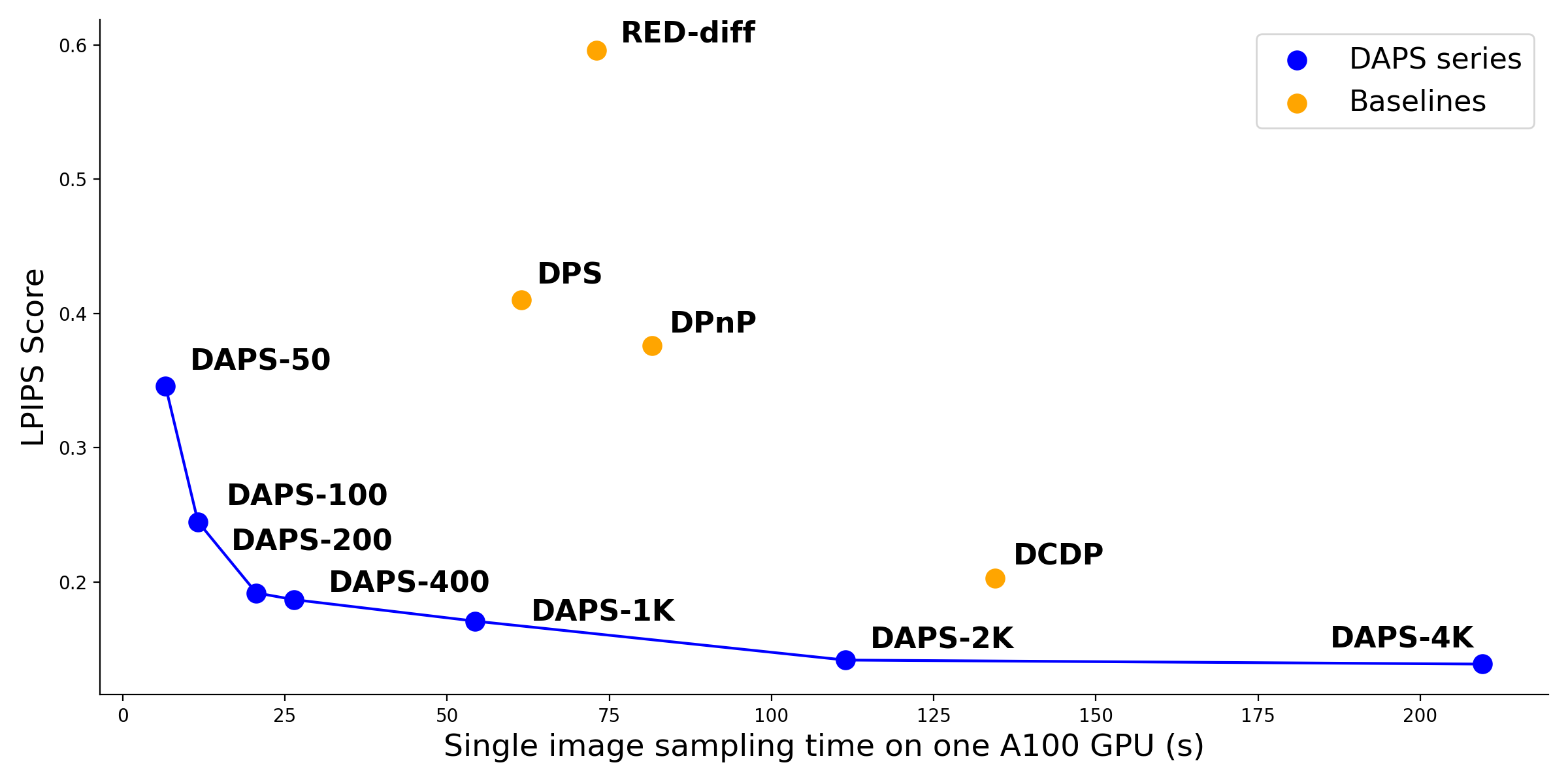}
    \caption{Phase retrieval}
    \end{subfigure}
    
    \caption{\small \textbf{Evaluation of time cost and sample quality}. The x-axis indicates the single image sampling time on an A100-SXM4-80GB GPU and y-axis shows the LPIPS. The evaluation uses 100 FFHQ images.}
    \label{fig:time_cost}
    \vspace{-5pt}
\end{figure*}

As mentioned in \cref{sec:method}, we implement a few-step Euler ODE solver to compute $\hat \rvx_0(\rvx_t)$, while maintaining the same number of neural function evaluations (NFE) for different noise levels. In our experiments, we use DAPS-1K for all linear tasks and DAPS-4K for all nonlinear tasks. DAPS-1K uses $4$ ODE solver NFE and $250$ annealing scheduling steps, while DAPS-4K uses $10$ NFE and $400$ steps, respectively. We discuss the effect of choosing these configurations in~\cref{sec:abla}. The corresponding latent version, LatentDAPS, follows the same settings but performs sampling in the latent space of VAEs. We use $100$ and $50$ Langevin steps per denoising iteration for DAPS and LatentDAPS respectively, and tune learning rates separately for each task.
Further details on model configurations, samplers, and other hyperparameters are provided in \cref{appendix:details}, along with more discussion on sampling efficiency in \cref{fig:time_cost} and \cref{appendix:efficiency}.

\textbf{Datasets and metrics.} Adopting the previous convention, we test our method on two image datasets, FFHQ $256\times 256$ \cite{karras2021style} and ImageNet $256\times 256$ \cite{deng2009imagenet}.  To evaluate our method, we use $100$ images from the validation set for both FFHQ and ImageNet. We include peak signal-to-noise-ratio (PSNR), structural similarity index measure (SSIM), Learned Perceptual Image Patch Similarity (LPIPS)\cite{zhang2018lpips} score and Fréchet inception distance (FID)\cite{heusel2018ganstrainedtimescaleupdate} as our main evaluation metrics. For both our method and baselines, we use the versions implemented in piq \cite{kastryulin2022piq} with all images normalized to the range $[0, 1]$. The replace-pooling option is enabled for LPIPS evaluation.

\textbf{Inverse problems.} We evaluate our method with a series of linear and nonlinear tasks. For linear inverse problems, we consider (1) super-resolution, (2) Gaussian deblurring, (3) motion deblurring, (4) inpainting (with a box mask), and (5) inpainting (with a 70\% random mask). For Gaussian and motion deblurring, kernels of size 61$\times$61 with standard deviations of 3.0 and 0.5, respectively, are used. In the super-resolution task, a bicubic resizer downscales images by a factor of 4. The box inpainting task uses a random box of size 128$\times$128 to mask the original images, while random mask inpainting uses a generated random mask where each pixel has a 70\% chance of being masked, following the settings in \cite{song2024solving}.

We consider three nonlinear inverse problems: (1) phase retrieval, (2) high dynamic range (HDR) reconstruction, and (3) nonlinear deblurring. Due to the inherent instability of phase retrieval, we adopt the strategy from DPS \cite{chung2023diffusion}, using an oversampling rate of 2.0 and reporting the best result out of four independent samples. The goal of HDR reconstruction is to recover a higher dynamic range image (factor of 2) from a low dynamic range image. For nonlinear deblurring, we use the default setting as described in \cite{m_Tran-etal-CVPR21}. All linear and nonlinear measurements are subject to white Gaussian noise with a standard deviation of $\beta_\rvy = 0.05$. Further details regarding the forward measuring functions for each task and their respective hyperparameters are provided in the \cref{appendix:details}.

\textbf{Baselines.} We compare our methods with the following baselines: DDRM~\cite{kawar2022denoising}, DPS~\cite{chung2023diffusion}, DDNM~\cite{wang2022zero}, DCDP~\cite{li2024decoupleddataconsistencydiffusion}, FPS-SMC~\cite{dou2024diffusion}, DiffPIR~\cite{zhu2023denoising}, DPnP~\cite{xu2024provably} for pixel-based diffusion model experiments. We compare the latent diffusion version of our methods with PSLD~\cite{rout2023solving} and ReSample~\cite{song2024solving}, which also operate in the latent space. Note that DDRM, DDNM, FPS-SMC, and PSLD cannot handle nonlinear inverse problems. We especially RED-diff~\cite{mardani2023variational} for nonlinear experiments and a traditional method HIO~\cite{fienup1987phase} for phase retrieval.

\begin{table*}
\centering
\begin{adjustbox}{width=0.87\textwidth}
\begin{tabular}{l!{\vrule}l!{\vrule}l!{\vrule}cccc!{\vrule}cccc}
\toprule
\multirow{2}{*}{\textbf{Task}} & \multirow{2}{*}{\textbf{Type}} & \multirow{2}{*}{\textbf{Method}} & \multicolumn{4}{c}{\textbf{FFHQ}} & \multicolumn{4}{c}{\textbf{ImageNet}} \\  

 &  &  &  \textbf{PSNR ($\uparrow$)} & \textbf{SSIM ($\uparrow$)} & \textbf{LPIPS ($\downarrow$)} & \textbf{FID ($\downarrow$)} & \textbf{PSNR ($\uparrow$)} & \textbf{SSIM ($\uparrow$)} & \textbf{LPIPS ($\downarrow$)} & \textbf{FID ($\downarrow$)} \\ \midrule

 \multirow{10}{*}{Super resolution 4×} & \multirow{7}{*}{\textit{\textit{Pixel}}} & DAPS (ours) & $\textbf{29.07}$ & $\textbf{0.818}$ & $\textbf{0.177}$ & $\underline{51.44}$& $\textbf{25.89}$ & $\underline{0.694}$ & $\textbf{0.276}$ & $\textbf{83.57}$ \\

& & DPS & $25.86$ & $0.753$ & $0.269$ & $81.07$ & $21.13$ & $0.489$ & $0.361$ & $106.32$ \\
& & DDRM & $26.58$ & $0.782$ & $0.282$ & $79.25$ & $22.62$ & $0.521$ & $0.324$ & $103.85$ \\
& & DDNM & $28.03$ & $0.795$ & $0.197$ & $64.62$ & $23.96$ & $0.604$ & $0.475$ & $98.62$\\
& & DCDP & $\underline{28.66}$ & $0.807$ & $\underline{0.178}$ & $53.81$ & - & - & - & -\\
& & FPS-SMC & $28.42$ & $\underline{0.813}$ & $0.204$ & $\textbf{49.25}$ & $\underline{24.82}$ & $\textbf{0.703}$ & $\underline{0.313}$ & $\underline{97.51}$\\
& & DiffPIR & $26.64$ & - & $0.260$ & $65.77$ & $23.18$ & - & $0.371$ & $106.32$ \\

\cmidrule{2-11}
& \multirow{3}{*}{\textit{\textit{Latent}}} & LatentDAPS(ours) & $\textbf{27.48}$ & $\textbf{0.801}$ & $\textbf{0.182}$ & $59.62$ & $\underline{25.06}$ & $\underline{0.673}$ & $\textbf{0.276}$ & $\textbf{84.37}$\\
& & PSLD & $\underline{24.35}$ & $\underline{0.649}$ & $\underline{0.287}$ & $74.36$ & $\textbf{25.42}$ & $\textbf{0.694}$ & $\underline{0.360}$ & $\underline{97.45}$\\
& & ReSample & $23.29$ & $0.594$ & $0.392$ & $93.18$ & $22.61$ & $0.576$ & $0.370$ & $113.42$\\
\midrule

\multirow{9}{*}{Inpaint (box)} & \multirow{6}{*}{\textit{\textit{Pixel}}} & DAPS(ours) & $24.07$ & $0.814$ & $\textbf{0.133}$ & $\textbf{43.10}$& $21.43$ & $0.725$ & $\underline{0.214}$ & $\underline{109.85}$ \\
& & DPS & $22.51$ & $0.792$ & $0.209$ & $61.27$ & $18.94$ & $0.722$ & $0.257$ & $126.52$\\
& & DDRM & $22.26$ & $0.801$ & $0.207$ & $78.62$ & $18.63$ & $0.733$ & $0.254$ & $116.37$\\
& & DDNM & $\underline{24.47}$ & $\textbf{0.837}$ & $0.235$ & $46.59$ & $\underline{21.64}$ & $\textbf{0.748}$ & $0.319$ & $\textbf{103.97}$\\
& & DCDP & $23.89$ & $0.760$ & $0.163$ & $\underline{45.23}$ & - & - & - & -\\
& & FPS-SMC & $\textbf{24.86}$ & $\underline{0.823}$ & $\underline{0.146}$ & $48.34$ & $\textbf{22.16}$ & $\underline{0.726}$ & $\textbf{0.208}$ & $111.58$ \\

\cmidrule{2-11}
& \multirow{3}{*}{\textit{Latent}} & LatentDAPS(ours) & $\underline{23.99}$ & $\underline{0.802}$ & $0.194$ & $\underline{46.52}$ & $17.19$ & $0.624$ & $\underline{0.340}$ & $\underline{145.63}$ \\
& & PSLD & $\textbf{24.22}$ & $\textbf{0.813}$ & $\textbf{0.158}$ & $\textbf{43.02}$ & $\textbf{20.10}$ & $\textbf{0.694}$ & $0.465$ & $146.53$\\
& & ReSample & $20.06$ & $0.749$ & $\underline{0.184}$ & $53.21$ & $\underline{18.29}$ & $\underline{0.631}$ & $\textbf{0.262}$ & $\textbf{127.84}$\\
\midrule

\multirow{8}{*}{Inpaint (random)} & \multirow{5}{*}{\textit{Pixel}} & DAPS(ours) & $\textbf{31.12}$ & $\textbf{0.844}$ & $\textbf{0.098}$ & $\textbf{32.17}$& $\underline{28.44}$ & $\underline{0.775}$ & $\textbf{0.135}$ & $\textbf{54.25}$\\
& & DPS & $25.46$ & $0.823$ & $0.203$ & $69.20$ & $23.52$ & $0.745$ & $0.297$ & $87.53$ \\
& & DDNM & $29.91$ & $0.817$ & $\underline{0.121}$ & $\underline{44.37}$ & $\textbf{31.16}$ & $\textbf{0.841}$ & $\underline{0.191}$ & $\underline{63.84}$\\
& & DCDP & $\underline{30.69}$ & $\underline{0.842}$ & $0.142$ & $52.51$ & - & - & - & -\\
& & FPS-SMC & $28.21$ & $0.823$ & $0.261$ & $61.23$ & $24.52$ & $0.701$ & $0.316$ & $79.12$\\

\cmidrule{2-11}
& \multirow{3}{*}{\textit{Latent}} & LatentDAPS(ours) & $\textbf{30.71}$ & $\textbf{0.813}$ & $\underline{0.141}$ & $\textbf{36.41}$ & $\underline{27.59}$ & $\underline{0.772}$ & $\underline{0.164}$ & $\underline{61.62}$ \\
& & PSLD & $\underline{30.31}$ & $\underline{0.809}$ & $0.221$ & $47.21$ & $\textbf{31.30}$ & $\textbf{0.783}$ & $0.337$ & $83.21$\\
& & ReSample & $29.61$ & $0.746$ & $\textbf{0.140}$ & $\underline{39.85}$ & $27.50$ & $0.756$ & $\textbf{0.143}$ & $\textbf{59.87}$ \\
\midrule

\multirow{10}{*}{Gaussian deblurring} & \multirow{7}{*}{\textit{\textit{Pixel}}} & DAPS(ours) & $\textbf{29.19}$ & $\textbf{0.817}$ & $\textbf{0.165}$ & $\textbf{53.33}$& $\underline{26.15}$ & $\underline{0.684}$ & $\textbf{0.253}$ & $\textbf{75.68}$ \\
& & DPS & $25.87$ & $0.764$ & $0.219$ & $79.75$ & $20.31$ & $0.598$ & $0.397$ & $116.42$\\
& & DDRM & $24.93$ & $0.732$ & $0.239$ & $92.43$ & $21.26$ & $0.564$ & $0.443$ & $146.89$ \\
& & DDNM & $\underline{28.20}$ & $\underline{0.804}$ & $\underline{0.216}$ & $\underline{57.83}$ & $\textbf{28.06}$ & $\textbf{0.703}$ & $\underline{0.278}$ & $\underline{81.43}$\\
& & DCDP & $27.50$ & $0.699$ & $0.304$ & $86.43$ & - & - & - & - \\
& & FPS-SMC & $26.54$ & $0.773$ & $0.253$ & $67.45$ & $23.91$ & $0.601$ & $0.387$ & $91.72$\\
& & DiffPIR & $27.36$ & - & $0.236$ & $59.65$ & $22.80$ & - & $0.355$ & $93.36$ \\

\cmidrule{2-11}
& \multirow{3}{*}{\textit{Latent}} & LatentDAPS(ours) & $\textbf{27.93}$ & $\textbf{0.764}$ & $\textbf{0.234}$ & $\textbf{64.52}$ & $25.05$ & $0.668 $ & $\underline{0.345}$ & $\underline{78.51}$ \\
& & PSLD & $23.27$ & $0.631$ & $0.316$ & $89.51$ & $\underline{25.86}$ & $\underline{0.688}$ & $0.390$ & $91.39$\\
& & ReSample & $\underline{26.39}$ & $\underline{0.714}$ & $\underline{0.255}$ & $\underline{71.69}$ & $\textbf{25.97}$ & $\textbf{0.703}$ & $\textbf{0.254}$ & $\textbf{65.35}$\\
\midrule

\multirow{8}{*}{Motion deblurring} & \multirow{5}{*}{\textit{\textit{Pixel}}} & DAPS(ours) & $\textbf{29.66}$ & $\textbf{0.847}$ & $\textbf{0.157}$ & $\textbf{39.49}$& $\textbf{27.86}$ & $\textbf{0.766}$ & $\textbf{0.196}$ & $\textbf{61.83}$ \\
& & DPS & $24.52$ & $0.801$ & $0.246$ & $65.23$ & $18.96$ & $0.629$ & $0.423$ & $137.81$\\
& & DCDP & $25.08$ & $0.512$ & $0.364$ & $125.13$ & - & - & - & -\\
& & FPS-SMC & $\underline{27.39}$ & $\underline{0.826}$ & $\underline{0.227}$ & $\underline{48.32}$ & $\underline{24.52}$ & $\underline{0.647}$ & $\underline{0.326}$ & $\underline{87.43}$\\
& & DiffPIR & $26.57$ & - & $0.255$ & $65.78$ & $24.01$ & - & $0.366$ & $94.63$ \\

\cmidrule{2-11}
& \multirow{3}{*}{\textit{Latent}} & LatentDAPS(ours) & $\underline{27.00}$ & $\underline{0.814}$ & $\underline{0.283}$ & $\underline{46.84}$ & $\underline{26.83}$ & $\textbf{0.745}$ & $\underline{0.296}$ & $\underline{73.62}$ \\
& & PSLD & $22.31$ & $0.678$ & $0.336$ & $96.15$ & $20.85$ & $0.594$ & $0.511$ & $124.67$\\
& & ReSample & $\textbf{27.41}$ & $\textbf{0.823}$ & $\textbf{0.198}$ & $\textbf{44.72}$ & $\textbf{26.94}$ & $\underline{0.738}$ & $\textbf{0.227}$ & $\textbf{66.89}$\\
\midrule

\multirow{7}{*}{Phase retrieval} & \multirow{4}{*}{\textit{\textit{Pixel}}} & DAPS(ours) & $\textbf{30.63}_{\pm 3.13}$ & $\textbf{0.851}_{\pm 0.072}$ & $\textbf{0.139}_{\pm 0.060}$ & $\textbf{42.71}$ & $\textbf{25.78}_{\pm 6.92}$ & $\textbf{0.743}_{\pm 0.084}$ & $\textbf{0.254}_{\pm 0.125}$ & $\textbf{82.67}$\\
& & DPS & $17.64_{\pm 2.97}$ & $0.441_{\pm 0.129}$ & $0.410_{\pm 0.090}$ & $104.52$ & $\underline{16.81}_{\pm 3.61}$ & $\underline{0.427}_{\pm 0.143}$ & $\underline{0.447}_{\pm 0.099}$ & $\underline{197.54}$ \\
& & RED-diff & $15.60_{\pm 4.48}$ & $0.398_{\pm 0.195}$ & $0.596_{\pm 0.092}$ & $167.43$ & $14.98_{\pm 3.75}$ & $0.386_{\pm 0.057}$ & $0.536_{\pm 0.129}$ & $212.24$ \\
& & DCDP & $\underline{28.65}_{\pm 8.09}$ & $\underline{0.781}_{\pm 0.217}$ & $\underline{0.203}_{\pm 0.196}$ & $\underline{68.13}$ & - & - & - & -\\

\cmidrule{2-11}
& \multirow{2}{*}{\textit{Latent}} & LatentDAPS(ours) & $\textbf{29.16}_{\pm 3.55}$ & $\textbf{0.796}_{\pm 0.089}$ & $\textbf{0.199}_{\pm 0.078}$ & $\textbf{54.26}$ & $\textbf{20.54}_{\pm 6.41}$ & $\textbf{0.612}_{\pm 0.114}$ & $\textbf{0.361}_{\pm 0.150}$ & $\textbf{129.54}$ \\
& & ReSample & $21.60_{\pm 8.10}$ & $0.648_{\pm 0.154}$ & $0.406_{\pm 0.224}$ & $84.32$ & $19.24_{\pm 4.21}$ & $0.618_{\pm 0.146}$ & $0.403_{\pm 0.174}$ & $130.47$\\

\cmidrule{2-11}
& \textit{Classical} & HIO & $13.53_{\pm 2.50}$ & $0.359_{\pm 0.093}$ & $0.726_{\pm 0.068}$ & $268.09$ & - & - & - & - \\
\midrule

\multirow{6}{*}{Nonlinear deblur} & \multirow{4}{*}{\textit{\textit{Pixel}}} & DAPS(ours) & $\underline{28.29}_{\pm 1.77}$ & $\underline{0.783}_{\pm 0.036}$ & $\textbf{0.155}_{\pm 0.032}$ & $\underline{49.38}$ & $\underline{27.73}_{\pm 3.23}$ & $\underline{0.724}_{\pm 0.048}$ & $\textbf{0.169}_{\pm 0.056}$ & $\underline{59.87}$\\
& & DPS & $23.39_{\pm 2.01}$ & $0.623_{\pm 0.082}$ & $0.278_{\pm 0.060}$ & $91.31$ & $22.49_{\pm 3.20}$ & $0.591_{\pm 0.101}$ & $0.306_{\pm 0.081}$ & $101.41$\\
& & RED-diff & $\textbf{30.86}_{\pm 0.51}$ & $\textbf{0.795}_{\pm 0.028}$ & $\underline{0.160}_{\pm 0.034}$ & $\textbf{43.84}$ & $\textbf{30.07}_{\pm 1.41}$ & $\textbf{0.754}_{\pm 0.023}$ & $\underline{0.211}_{\pm 0.083}$ & $\textbf{51.22}$\\
& & DCDP & $27.92_{\pm 2.64}$ & $0.779_{\pm 0.067}$ & $0.183_{\pm 0.051}$ & $51.96$ & - & - & - & -\\

\cmidrule{2-11}
& \multirow{2}{*}{\textit{Latent}} & LatentDAPS(ours) & $28.11_{\pm 1.75}$ & $0.713_{\pm 0.041}$ & $0.235_{\pm 0.049}$ & $53.63$ & $25.34_{\pm 3.44}$ & $0.615_{\pm 0.057}$ & $0.314_{\pm 0.080}$ & $76.73$\\
& & ReSample & $\textbf{28.24}_{\pm 1.69}$ & $0.742_{\pm 0.039}$ & $\textbf{0.185}_{\pm 0.039}$ & $\textbf{51.62}$ & $\textbf{26.20}_{\pm 3.71}$ & $\textbf{0.653}_{\pm 0.064}$ & $\textbf{0.206}_{\pm 0.057}$ & $\textbf{61.16}$\\
\midrule

\multirow{5}{*}{High dynamic range} & \multirow{3}{*}{\textit{\textit{Pixel}}} & DAPS(ours) & $\textbf{27.12}_{\pm 3.53}$ & $\textbf{0.752}_{\pm 0.041}$ & $\textbf{0.162}_{\pm 0.072}$ & $\textbf{42.97}$ & $\textbf{26.30}_{\pm 4.10}$ & $\textbf{0.717}_{\pm 0.067}$ & $\textbf{0.175}_{\pm 0.107}$ & $\textbf{64.19}$\\
& & DPS & $\underline{22.73}_{\pm 6.07}$ & $\underline{0.591}_{\pm 0.141}$ & $0.264_{\pm 0.156}$ & $112.82$ & $19.23_{\pm 2.52}$ & $0.582_{\pm 0.082}$ & $0.503_{\pm 0.106}$ & $146.23$\\
& & RED-diff & $22.16_{\pm 3.41}$ & $0.512_{\pm 0.083}$ & $\underline{0.258}_{\pm 0.089}$ & $\underline{108.32}$ & $\underline{22.03}_{\pm 5.90}$ & $\underline{0.601}_{\pm 0.094}$ & $\underline{0.274}_{\pm 0.198}$ & $\underline{113.48}$\\

\cmidrule{2-11}
& \multirow{2}{*}{\textit{Latent}} & LatentDAPS(ours) & $\textbf{25.94}_{\pm 2.87}$ & $\textbf{0.751}_{\pm 0.056}$ & $0.223_{\pm 0.080}$ & $74.83$ & $23.64_{\pm 4.10}$ & $0.609_{\pm 0.053}$ & $0.269_{\pm 0.099}$ & $93.51$\\
& & ReSample & $25.65_{\pm 3.57}$ & $0.732_{\pm 0.059}$ & $\textbf{0.182}_{\pm 0.085}$ & $\textbf{67.22}$ & $\textbf{25.11}_{\pm 4.21}$ & $\textbf{0.633}_{\pm 0.049}$ & $\textbf{0.198}_{\pm 0.089}$ & $\textbf{87.66}$\\
\bottomrule

\end{tabular}
\end{adjustbox}
\caption{\textbf{Quantitative evaluation on FFHQ and ImageNet on 5 linear and 3 nonlinear tasks.} Performance comparison of different methods on FFHQ (left) and ImageNet (right) in the image domain. For linear tasks, we report the mean performance (PSNR, SSIM and LPIPS) across 100 validation images. We provide both the mean and standard deviation for nonlinear tasks, where results exhibit higher instability. The FID scores are evaluated with the same $100$ validation images. The best and second-best results within each type of task are indicated by \textbf{bold} and \underline{underlined} marks, respectively. DAPS demonstrates superior performance on most tasks, with LatentDAPS showing competitive results. All tasks are using noisy measurement with noise level $\beta_{\rvy}=0.05$.}
\label{tab:results}
\end{table*}

\subsection{Main Results}
We show quantitative results for FFHQ and ImageNet dataset in \cref{tab:results}. The qualitative comparisons are provided in \cref{main-comparison}. As shown in \cref{tab:results}, our method achieves comparable or even better performance across all selected linear inverse problems. Moreover, our methods are remarkably stable when handling nonlinear inverse problems. For example, although existing methods such as DPS can recover high-quality images from phase retrieval measurements, they suffer from an extremely high failure rate, resulting in a relatively low PSNR and a high LPIPS. However, DAPS demonstrates superior stability across different samples and achieves a significantly higher success rate, resulting in a substantial improvement in PSNR and LPIPS as evidenced by the results in \cref{tab:results}. Moreover, DAPS captures and recovers much more fine-grained details in measurements compared to existing baselines. We include more samples and comparisons in \cref{appendix:addition} for further illustration.

\begin{figure*}
    \centering
    \includegraphics[width=\textwidth]{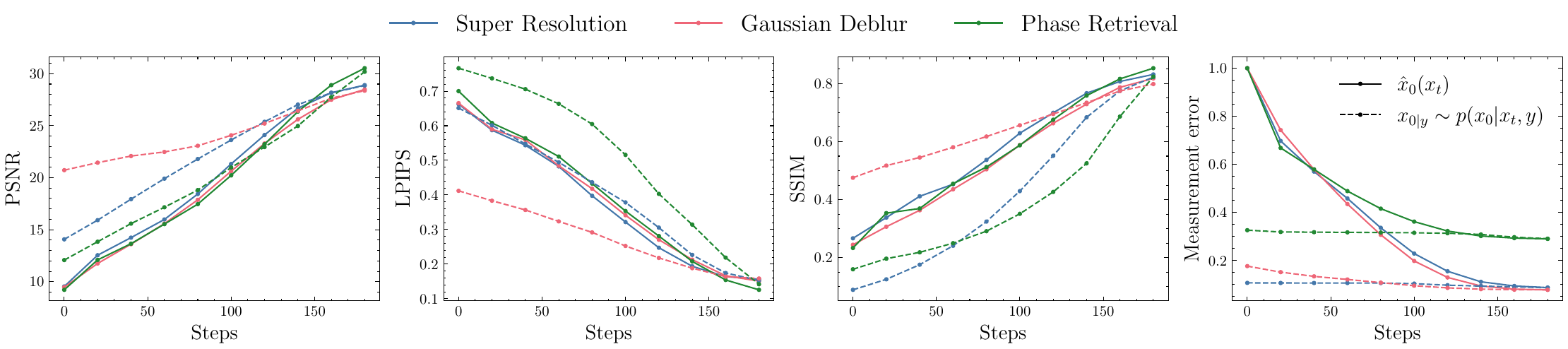}
    \vspace{-13pt}
    \caption{\textbf{Quantitative evaluations of image quality metrics and measurement error throughout the sampling process.} Here $\hat \rvx_0(\rvx_t)$ stands for the estimated mean of the Gaussian approximated distribution and $\rvx_{0\mid\rvy}\sim p(\rvx_0\mid\rvx_{t}, \rvy)$ means the samples from measurement conditioned time-marginal. The measurement errors are normalized between $[0, 1]$ for visualization.}
    \label{fig:quant-traj}
    \vspace{-10pt}
\end{figure*}

To understand the evolution of the noise annealing procedure, we evaluate two crucial trajectories in our method: 1) the estimated mean of \(p(\rvx_0\mid\rvx_t)\), denoted as \(\hat{\rvx}_0(\rvx_t)\); and 2) samples from \(p(\rvx_0\mid\rvx_t, \rvy)\) obtained through Langevin dynamics, denoted as \(\rvx_{0\mid\rvy}\). We assess image quality and measurement error across these trajectories, as depicted in \cref{fig:quant-traj}. The measurement error for \(\rvx_{0\mid\rvy} \sim p(\rvx_0\mid\rvx_t, \rvy)\) remains consistently low throughout the process, while the measurement error for \(\hat{\rvx}_0(\rvx_t)\) continuously decreases. Consequently, metrics such as PSNR and LPIPS exhibit stable improvement as the noise level is annealed.

\begin{figure}
    \begin{subfigure}[b]{.50\textwidth}
        \includegraphics[width=\textwidth]{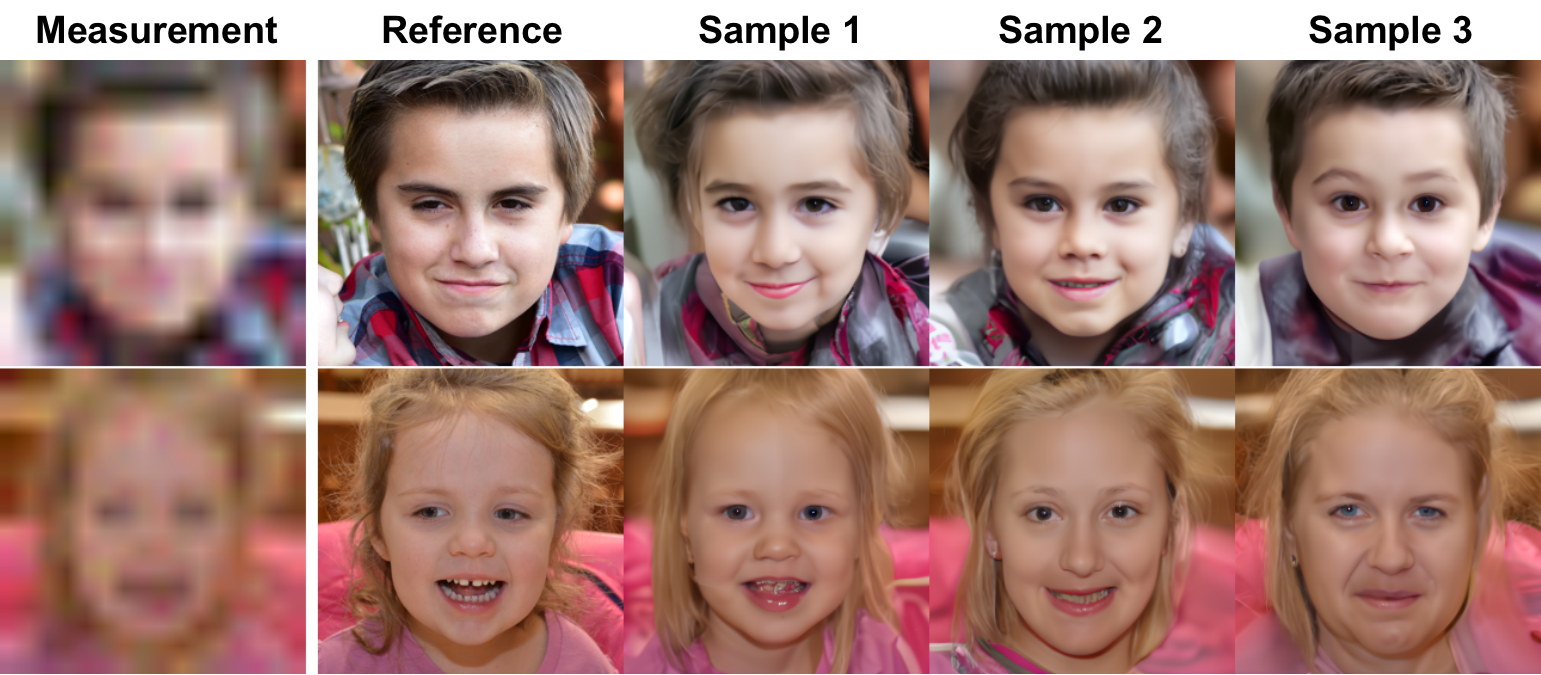}
    \caption{Super resolution of a factor $16$}
    \label{fig:sr-f16}
    \end{subfigure}
    \begin{subfigure}[b]{.50\textwidth}
        \includegraphics[width=\textwidth]{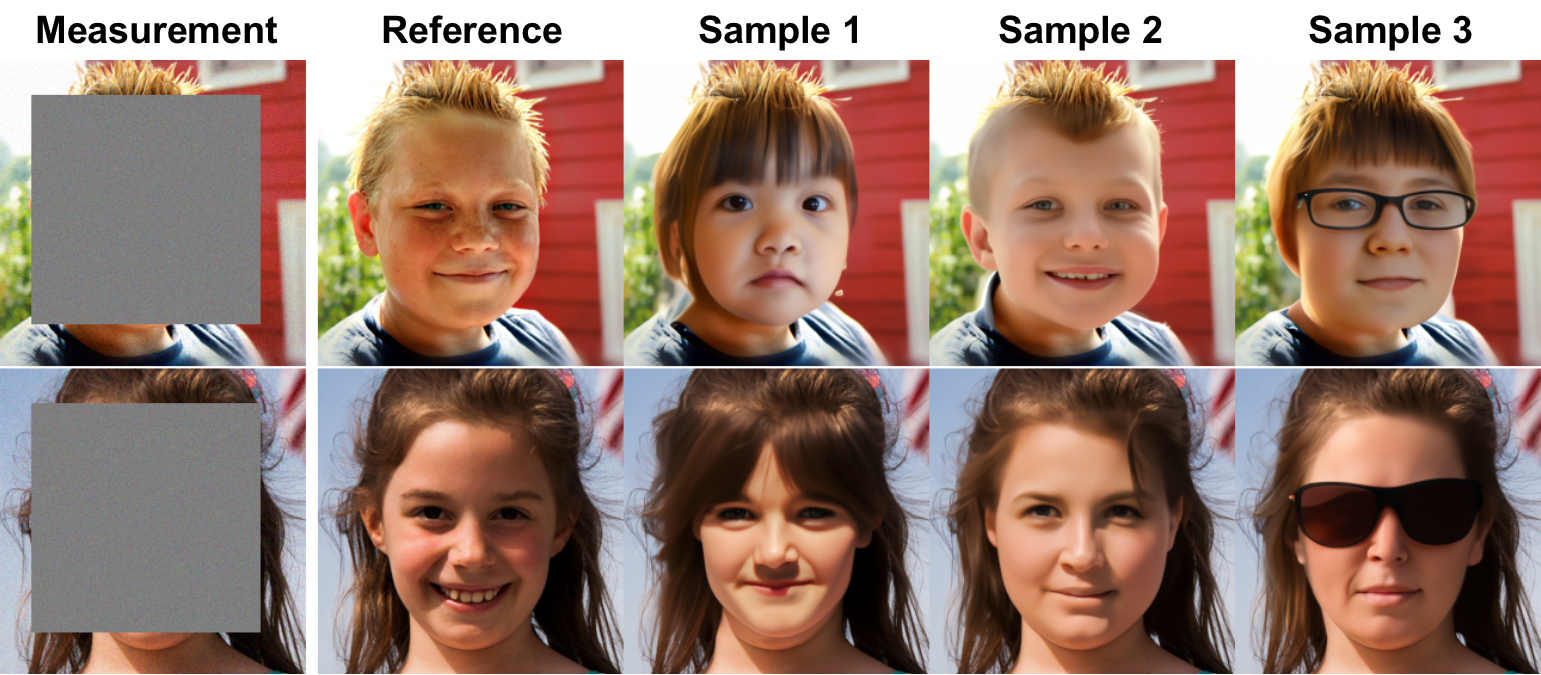}
    \caption{Inpainting of 192$\times$192 box}
    \label{fig:ipt-192}
    \end{subfigure}
\caption{\textbf{Sample diversity}. We present several diverse samples generated by the DAPS under two sparse measurements whose posterior distributions contain multiple modes. DAPS produces a variety of samples with distinct features, including differences in expression, wearings, and hairstyles.}
\vspace{-10pt}
\label{fig:diversity}
\end{figure}

In addition, our method can generate diversified samples when the measurements contain less information. To illustrate this, we selected two tasks where the posterior distributions may have multiple modes: (1) super-resolution by a factor of 16, and (2) box inpainting with a box size of 192$\times$192. Some samples from DAPS are shown in \cref{fig:diversity}, demonstrating DAPS's ability to produce diverse samples while preserving the measurement information. We show more samples in \cref{appendix:addition} for further demonstration.

\MOD{\noindent\textbf{DAPS with large-scale text-conditioned latent diffusion.} Due to the flexibility of DAPS, pretrained large-scale text-conditioned latent diffusion models (LDMs) \cite{rombach2022high, podell2023sdxlimprovinglatentdiffusion, ramesh2022hierarchicaltextconditionalimagegeneration} can also serve as the prior model $p(\rvx_0\mid \rvc)$ when given a text prompt $\rvc$. To demonstrate DAPS's potential, we use Stable Diffusion v2.1 \cite{rombach2022high} to present LatentDAPS results on three tasks with 768$\times$768 resolution images, as shown in \cref{main-comparison}. Additionally, we provide the text prompts used, along with a detailed discussion, in \cref{appendix:sd}, where we also include a quantitative evaluation.}

\MOD{\noindent\textbf{DAPS with discrete diffusion.} DAPS can be naturally extended to support posterior sampling with categorical prior distribution modeled by discrete diffusion models~\cite{austin2021structured,campbell2022continuous,lou2023discrete}. Instead of sampling $\rvx_{0|\rvy}$ by Langevin MC or Hamiltonian MC in continuous space, we use the Metropolis-Hasting algorithm to replace~\cref{eq:langevin}. We validate DAPS with discrete diffusion on a quantized MNIST dataset. We leave the details to \cref{appendix:discrete}.} 

\subsection{Further results on MRI}
\MOD{We further validate the effectiveness of DAPS on compressed sensing multi-coil MRI, which is a medical imaging technique that helps reduce the scan time of MRI via subsampling. We train a diffusion model using the pipeline from~\cite{karras2022elucidating} with fastMRI knee dataset~\cite{fastmri}. \cref{tab:MRI} shows that DAPS outperforms existing diffusion-based baselines. We include more qualitative results in~\cref{appendix:mri}.}

\begin{table}[t]
\centering
\begin{adjustbox}{width=\linewidth}
    \begin{tabular}{lccccccc}
    \toprule
    & \multicolumn{3}{c}{$\times 4$} & & \multicolumn{3}{c}{$\times 8$}\\
    \cline{2-4} \cline{6-8}\\[-10pt]
     & \textbf{PSNR ($\uparrow$)} & \textbf{SSIM ($\uparrow$)} & \textbf{Meas err ($\downarrow$)} & & \textbf{PSNR ($\uparrow$)} & \textbf{SSIM ($\uparrow$)} & \textbf{Meas err ($\downarrow$)}\\
    \midrule
    DAPS & $\textbf{31.48}$ & $\textbf{0.762}$ & $1.57$ & & $\textbf{29.01}$ & $\textbf{0.681}$ & $1.28$ \\
    DPS &  $26.13$ & $0.620$ & $9.90$ & & $20.82$ & $0.536$ & $6.74$\\
    DiffPIR & $28.31$ & $0.632$ & $10.55$ & & $26.78$ & $0.588$ & $7.79$\\
    ScoreMRI~\cite{chung2022score}  & $25.97$ & $0.468$ & $10.83$ & & $25.01$ & $0.405$ & $8.36$\\
    CSGM~\cite{jalal2021robust} & $28.78$ & $0.710$ & $\textbf{1.52}$ & & $26.15$ & $0.625$ & $\textbf{1.14}$\\
    \bottomrule
    \end{tabular}
    \end{adjustbox}
    \caption{\textbf{Quantitative results of compressed sensing MRI.} We consider $\times 4$ and $\times 8$ subsampling ratio. Meas err stands for $\ell_2$ data consistency error.}
    \vspace{-10pt}
    \label{tab:MRI}
\end{table}
\subsection{Ablation Studies}
\label{sec:abla}


\MOD{\noindent\textbf{Time efficiency of DAPS.} We run DAPS with various configurations to test its performance under different computing budgets. Specifically, we evaluate DAPS with the number of function evaluations (NFE) of the diffusion models ranging from 50 to 4K, with configurations specified in \cref{appendix:efficiency}. \cref{fig:time_cost} shows the relation between the sample quality of DAPS measured by LPIPS with its time cost, which indicates that DAPS is able to generate high-quality samples with a reasonable time cost.}

\MOD{\noindent\textbf{Effectiveness of Different Design choices of DAPS.} We conduct an ablation study to demonstrate the effectiveness of \textit{Gaussian approximation} compared with the more principled but more expensive \textit{diffusion-score estimation}. The results in \cref{tab:abla_approximation} demonstrate that \textit{Gaussian approximation} is $6\sim7$ times faster than \textit{diffusion-score estimation} with the cost of a slight performance drop, effectively trading off performance for efficiency.}

\begin{table}[t]
\centering
\begin{adjustbox}{width=\linewidth}
    \begin{tabular}{lccccccc}
    \toprule
    & \multicolumn{3}{c}{Gaussian deblurring} & & \multicolumn{3}{c}{High dynamic range}\\
    \cline{2-4} \cline{6-8}\\[-10pt]
     & \textbf{PSNR ($\uparrow$)} & \textbf{LPIPS ($\downarrow$)} & \textbf{Run time/s} & & \textbf{PSNR ($\uparrow$)} & \textbf{LPIPS ($\downarrow$)} & \textbf{Run time/s} \\
    \midrule
    \textit{Gaussian approx.} & $29.40$ & $0.170$ & $64.7$ & & $24.85 $ & $0.181$ & $61.4$ \\
    \textit{diffusion-score} &  $29.57$ & $0.160$ & $462.3$ & & $25.33$ & $0.193$ & $540.0$\\
    \textit{relative improvement} & $-0.575\%$ & $-6.25\% $& $614.53\%$ & & $-1.89\%$ & $-6.22\%$ & $779.48\%$  \\ 
    \bottomrule
    \end{tabular}
    \end{adjustbox}
    \caption{\textbf{Comparison of approaches to approximate $p(\rvx_0\mid \rvx_t)$.} Three tasks on $10$ FFHQ images with DAPS-1K consistently shown \textit{Gaussian approximation} achieve huge speed-up while slightly sacrificing the performance compared with \textit{diffusion-score estimation}.}
    \vspace{-10pt}
    \label{tab:abla_approximation}
\end{table}

\MOD{\noindent\textbf{More ablation studies.} We include more ablation studies of DAPS in \cref{appendix:ablation}, including the number of function evaluations, the number of ODE steps, the annealing noise scheduling, and different measurement noise levels.}


\section{Conclusion}
\label{sec:discussion}
In summary, we propose Decoupled Annealing Posterior Sampling (DAPS) for solving inverse problems, particularly those with complex nonlinear measurement processes such as phase retrieval. Our method decouples consecutive sample points in a diffusion sampling trajectory, allowing them to vary considerably, thereby enabling DAPS to explore a larger solution space. Empirically, we demonstrate that DAPS generates samples with better visual quality and stability compared to existing methods when solving a wide range of challenging inverse problems.


\section*{Acknowledgments}
We are grateful to Pika for providing the computing resources essential for this research. We also extend our thanks to the Kortschak Scholars Fellowship for supporting B.Z. and W.C. at Caltech. J.B. acknowledges support from the Wally Baer and Jeri Weiss Postdoctoral Fellowship. A.A. is supported in part by Bren endowed chair and by the AI2050 senior fellow program at Schmidt Sciences.

{
    \small
    \bibliographystyle{ieeenat_fullname}
    \bibliography{main}
}

\clearpage
\appendix
\setcounter{page}{1}
\onecolumn

    {\centering
    \Large
    \textbf{\thetitle}\\
    \vspace{0.5em}Supplementary Material \\
    \vspace{1.5em}
}

\section{DAPS with Pixel Diffusion Models}\label{appendix:pixel}
Pixel diffusion models learn to approximate the time-dependent score function $\vs_\vtheta(\rvx_t, t)\approx  \nabla_{\rvx_t}\log p(\rvx_t;\sigma_t)$, where $\sigma_t$ is a predefined noise schedule with $\sigma_0=0$ and $\sigma_T=\sigma_{\max}$. Langevin dynamics is used in DAPS due to its simplicity and flexibility. However, other advanced MCMC sampler such as Hamiltonian Monte Carlo (HMC) can be used to enhance the efficiency and quality. We will give a detailed derivation in \cref{appendix:hmc}. We also include a discussion of using the Metropolis Hasting algorithm, a gradient-free MCMC sampling method in \cref{appendix:mh}.

According to the Langevin dynamics updating rule \cref{eq:langevin}, HMC updating rule \cref{eq:hmc} and Metropolis Hasting updating rule \cref{eq:mh}, we summarize the algorithm of DAPS with pixel diffusion models in \cref{alg1} with Langevin dynamics, HMC or Metropolis Hasting. We also include an ablation study of different methods in \cref{appendix:addition}.

\subsection{Sampling with Hamiltonian Monte Carlo}
\label{appendix:hmc}
Hamiltonian Monte Carlo~\cite{betancourt2015hamiltonian} is designed to efficiently sample from complex, high-dimensional probability distributions by leveraging concepts from Hamiltonian dynamics to reduce random walk behavior and improve exploration of the target distribution. Recall our goal is to sample from the proposal distribution \( p(\rvx_0 \mid \rvx_t, \rvy) \). We define the Hamiltonian $H(\rvx_0, \rvp) = U(\rvx_0) + K(\rvp)$, where the potential energy $ U(\rvx_0) = -\log p(\rvx_0 \mid \rvx_t, \rvy)$ and the kinetic energy $ K(\rvp) = \frac{1}{2}\rvp^\top \rvp$. We then simulate Hamiltonian dynamics to propose new samples. 

\begin{enumerate}
    \item  Start with the estimator from probability flow ODE $ \hat{\rvx}_0^{(0)}=\hat\rvx_{0}(\rvx_{x_{t}})$ and sample a initial momentum $\rvp^{(0)} \sim \mathcal{N}(\vzero, \mI)$ at diffusion time step $t$.
    \item Simulate Hamiltonian Dynamics with step size $\eta_t$, momentum damping factor $\gamma_t$, and number of steps $N$, where the updating rule for $j=1,\cdots, N$,
    \begin{equation}\label{eq:hmc}
        (\hat{\rvx}_0^{(j+1)}, \rvp^{(j+1)})=\text{Hamiltonian-Dynamics}(\hat{\rvx}_0^{(j)}, \rvp^{(j)}),
    \end{equation}
    is given by:
     \begin{align}
   \rvp^{(j+1)} &= (1-\gamma_t\eta_t) \cdot \rvp^{(j)} - \eta_t \nabla_{x_k}U(x_k)+ \sqrt{2\gamma_t \eta_t}\epsilon,\quad \epsilon \sim \mathcal N(0, I), \\
   \hat{\rvx}_0^{(j+1)} &= \hat{\rvx}_0^{(j)} + \eta_t \rvp^{(j+\frac{1}{2})}.
   \end{align}
\end{enumerate}

Empirically, Hamiltonian Monte Carlo efficiently explores the target distribution $p(\rvx_0 \mid \rvx_t, \rvy) $ and requires less number of steps $N$ to achieve similar performance compared to Langevin dynamics, allowing for more efficient sampling. We find HMC can speed up LatentDAPS with large-scale text-conditioned LDMs quite a bit. We will discuss this more in \cref{appendix:sd}. 

\begin{algorithm}[t]
\caption{Decoupled Annealing Posterior Sampling (DAPS)}
\label{alg1}
\begin{algorithmic}

\Require Score model $\vs_{\vtheta}$, measurement $\rvy$, noise schedule $\sigma_t$, $(t_i)_{i\in \{0,\dots, N_A\}}$.
\State Sample $\rvx_{T} \sim \mathcal N(\vzero, \sigma_{T}^2\mI)$.
\For {$i = N_A, N_A-1, \dots, 1$}
\State {Initial $\rvp^{(0)}\sim \mathcal N(\vzero, \mI)$ for HMC only}
\State {Compute $\hat \rvx_0^{(0)}=\hat \rvx_0(\rvx_{t_i})$ by solving the probability flow ODE in \cref{eq:pfode} with $\vs_\vtheta$}
\For {$j = 0,\dots, N-1$}
\State {\textit{Langevin dynamics:}}
\begin{equation*}
    \hat \rvx_0^{(j+1)}  \leftarrow \hat \rvx_0^{(j)} + \eta_t \Big(\nabla_{\hat \rvx_0} \log p(\hat \rvx_0^{(j)}|\rvx_{t_i}) +  \nabla_{\hat \rvx_0} \log p(\rvy|\hat \rvx_0^{(j)})\Big)+ \sqrt{2\eta_t} \bm\epsilon_j, \ \bm\epsilon_j \sim \mathcal N(\vzero,\mI).
\end{equation*}

\State {\textit{or HMC:}}
 \begin{equation*}
        (\hat{\rvx}_0^{(j+1)}, \rvp^{(j+1)})\leftarrow\text{Hamiltonian-Dynamics}(\hat{\rvx}_0^{(j)}, \rvp^{(j)}),
    \end{equation*}
\State {\textit{or Metropolis Hasting:}}
 \begin{equation*}\hat{\rvx}_0^{(j+1)}\leftarrow\text{Metropolis-Hasting}(\hat{\rvx}_0^{(j)})
    \end{equation*}
\EndFor
\State Sample $\rvx_{t_{i-1}} \sim \mathcal N(\hat \rvx_0^{(N)},\sigma_{t_{i-1}}^2 \mI)$.
\EndFor
\State \textbf{Return} $\rvx_0$
\end{algorithmic}
\end{algorithm}

\subsection{Sampling with Metropolis Hasting algorithm}\label{appendix:mh}
The Metropolis-Hastings algorithm~\cite{metropolis1953equation} is a MCMC method used to sample from a target distribution $p(\rvx) $, especially when direct sampling is infeasible. It works by constructing a Markov chain whose stationary distribution corresponds to $ p(\rvx) $.  Here we discuss how to sample from $p(\rvx_0\mid \rvx_t, \rvy)$ using Metropolis Hasting under \textit{Gaussian approximation}. We adopt the Gaussian kernel as the proposal distribution which is symmetry,
\begin{equation}
    q(\rvx\to \rvx') = \mathcal N(\rvx'; \rvx, \eta_t^2 \mI)
\end{equation}
where $\eta_t$ is a hyperparameter to control the strength of the perturbation at diffusion time step $t$. Then we give the process of the Metropolis Hasting algorithm.
\begin{enumerate}
    \item Start with the estimator from probability flow ODE $ \hat{\rvx}_0^{(0)}=\hat\rvx_{0}(\rvx_{x_{t}})$.
    \item Perturb the current position by zero-centered Gaussian noise with standard deviation $\eta_t$, and number of steps $N$, for $j=1,\cdots, N$,
    \begin{equation}
        \rvx_{prop}^{(j)} = \hat\rvx_{0}^{(j)} +\eta_t\epsilon,\quad \epsilon\sim \mathcal N(\vzero, \mI) 
    \end{equation}
    \item Update position by,
    \begin{equation}\label{eq:mh}
        \hat\rvx_{0}^{(j+1)} =\text{Metropolis-Hasting}(\hat{\rvx}_0^{(j)})= \begin{cases}
            \rvx_{prop}^{(j)}\quad &\text{w.p. }\, \alpha(\hat\rvx_{0}^{(j)}\to \rvx_{prop}^{(j)})\\
            \hat\rvx_{0}^{(j)}\quad &\text{otherwise}
        \end{cases},
    \end{equation}
    where the acceptance probability $\alpha(\hat\rvx_{0}^{(j)}\to \rvx_{prop}^{(j)})$ is given by,
    \begin{equation}
        \alpha(\hat\rvx_{0}^{(j)}\to \rvx_{prop}^{(j)}) = \min\left(1, \dfrac{p(\rvx_{prop}^{(j)}\mid \rvx_t, \rvy)}{p(\hat\rvx_{0}^{(j)}\mid \rvx_t, \rvy)}\right)\approx \min\left(1, \dfrac{\mathcal N(\rvx_{prop}^{(j)};\rvx_t, \sigma_t^2\mI)\cdot \mathcal N(\rvy; \mathcal A(\rvx_{prop}^{(j)}), \beta_\rvy^2\mI)}{\mathcal N(\hat\rvx_{0}^{(j)};\rvx_t, \sigma_t^2\mI)\cdot \mathcal N(\rvy; \mathcal A(\hat\rvx_{0}^{(j)}), \beta_\rvy^2\mI)}\right)
    \end{equation}
\end{enumerate}

Empirically, Metropolis Hasting usually performs worse than Langevin dynamics and HMC and less efficient but more flexible to tasks that don't have access to the gradient of the data likelihood, $\nabla_{\rvx_0} \log p(\rvy\mid \rvx_0)$, \ie the forward function $\mathcal A$ is non-differentiable or its gradient is inaccessible. We will discuss more about such tasks in \cref{appendix:discrete}.

\newpage
\section{DAPS with Latent Diffusion Models}
\label{appendix:latent}

Latent diffusion models (LDMs) \cite{rombach2022high} operate the denoising process not directly on the pixel space, but in a low-dimensional latent space. LDMs have been known for their superior performance and computational efficiency in high-dimensional data synthesis. In this section, we show that our method can be naturally extended to sampling with latent diffusion models.

Let $\mathcal E: \mathbb R^n \to \mathbb R^k$ and $\mathcal D: \mathbb R^k \to \mathbb R^n$ be a pair of encoder and decoder. Let $\rvz_0 = \mathcal E(\rvx_0)$ where $\rvx_0\sim p(\rvx_0)$, and $p(\rvz; \sigma)$ be the noisy distribution of latent vector $\rvz$ by adding Gaussian noises of variance $\sigma^2$ to the latent code of clean data. We have the following Proposition according to the factor graph in \cref{fig:factor_graph_latent}.

\begin{prop}\label{prop:2}
    Suppose $\rvz_{t_1}$ is sampled from the measurement conditioned time-marginal $p(\rvz_{t_1}\mid\rvy)$, then
    \begin{equation}
        \rvz_{t_2} \sim \mathbb E_{\rvx_0 \sim p(\rvx_0\mid\rvz_{t_1}, \rvy)}[\mathcal N( \mathcal E(\rvx_0), \sigma_{t_2}^2 \mI)]
    \end{equation}
    satisfies the measurement conditioned time-marginal $p(\rvz_{t_2}\mid\rvy)$.
    Moreover,
    \begin{equation}
        \rvz_{t_2} \sim \mathbb E_{\rvz_0 \sim p(\rvz_0\mid\rvz_{t_1}, \rvy)}[ \mathcal N(\rvz_0,\sigma_{t_2}^2 \mI)].
    \end{equation}
    also satisfies the measurement conditioned time-marginal $p(\rvz_{t_2}\mid\rvy)$.
\end{prop}

\textbf{Remark.} We can efficiently sample from $p(\rvx_0\mid\rvz_{t_1}, \rvy)$ using similar strategies as in \cref{sec:method}, \ie,
\begin{equation}
\label{eq:latent}
    \rvx_0^{(j+1)} = \rvx_0^{(j)} + \eta \cdot \Big(\nabla_{\rvx_0^{(j)}} \log p(\rvx_0^{(j)} \mid \rvz_{t_1}) + \nabla_{\rvx_0^{(j)}} \log p(\rvy\mid\rvx_0^{(j)})\Big) + \sqrt{2\eta} \bm\epsilon_j.
\end{equation}
We further approximate $p(\rvx_0^{(j)}\mid\rvz_{t_1})$ by $\mathcal N(\rvx_0^{(j)}; \mathcal D(\hat \rvz_0(\rvz_{t_1})), r_{t_1}^2 \mI)$, where $\hat \rvz_0(\rvz_{t_1})$ is computed by solving the (unconditional) probability flow ODE with a latent diffusion model $\vs_\vtheta$ starting at $\rvz_{t_1}$. The Langevin dynamics can then be rewritten as
\begin{equation}
    \rvx_0^{(j+1)} = \rvx_0^{(j)} - \eta \cdot \nabla_{\rvx_0^{(j)}}\left( \frac{\|\rvx_0^{(j)} - \mathcal D(\hat \rvz_0(\rvz_{t_1}))\|^2}{2r_{t_1}^2} + \frac{\|\mathcal A(\rvx) - \rvy\|^2}{2\beta_\rvy^2}\right) + \sqrt{2\eta} \bm\epsilon_j.
\end{equation}

On the other hand, we can also decompose $p(\rvz_0\mid\rvz_{t_1},\rvy) \approx p(\rvz_0\mid\rvz_{t_1}) p(\rvy\mid\rvz_0)$ and run Langevin dynamics directly on the latent space,
\begin{equation}
    \rvz_0^{(j+1)} = \rvz_0^{(j)} + \eta \cdot \Big(\nabla_{\rvz_0^{(j)}} \log p(\rvz_0^{(j)}\mid\rvz_{t_1}) + \nabla_{\rvz_0^{(j)}} \log p(\rvy\mid\rvz_0^{(j)})\Big) + \sqrt{2\eta} \bm\epsilon_j.
\end{equation}
Assuming $p(\rvz_0^{(j)}\mid\rvz_{t_1})$ by $\mathcal N(\rvz_0^{(j)}; \hat \rvz_0(\rvz_t), r_{t_1}^2 \mI)$, we derive another Langevin MCMC updating rule in the latent space,
\begin{equation}
    \rvz_0^{(j+1)} = \rvz_0^{(j)} - \eta \cdot \nabla_{\rvz_0^{(j)}} \left( \frac{\|\rvz_0^{(j)} - \hat \rvz_0(\rvz_{t_1})\|^2}{2r_{t_1}^2} + \frac{\|\mathcal A(\mathcal D(\rvz_0^{(j)})) - \rvy\|^2}{2\beta_\rvy^2}\right) + \sqrt{2\eta} \bm\epsilon_j.
\end{equation}

Both approaches are applicable to our posterior sampling algorithm. We summarize DAPS with latent diffusion models in \cref{alg2}. It is worth mentioning that when employing the \textit{Gaussian approximation} for $p(\rvx_0 \mid \rvx_t)$, pixel-space Langevin dynamics typically results in higher approximation errors compared to latent-space Langevin dynamics but offers significantly faster computation. A practical approach is to utilize pixel-space Langevin dynamics during the early stages (\ie, for $N_A \geq i > M$) to balance approximation accuracy with efficiency, transitioning to latent-space Langevin dynamics in the later stages (\ie, for $M \geq i \geq 1$) to achieve improved sample quality, where $M$ is a hyperparameter that trade-off efficiency and approximation accuracy. For simplicity, we set $M = N_A$ throughout our experiments, and leave the exploration of this hyperparameter as a future direction.

\begin{algorithm}[t]
\caption{Latent Diffusion Decoupled Annealing Posterior Sampling (LatentDAPS) with Langevin Dynamics}
\label{alg2}
\begin{algorithmic} 
\Require Latent space score model $\vs_\rvtheta$, measurement $\rvy$, noise schedule $\sigma_t$, $t_{i\in \{0,\dots, N_A\}}$, encoder $\mathcal E$ and decoder $\mathcal D$.
\State Sample $\rvz_{T} \sim \mathcal N(\vzero, \sigma_{T}^2\mI)$.
\For {$i = N_A, N_A-1, \dots, 1$}
\State {Compute $\hat \rvz_0^{(0)} = \hat \rvz_0(\rvz_{t_i})$ by solving the probability flow ODE in \cref{eq:pfode} with $\rvs_\rvtheta$.}
\State {\textit{Pixel space Langevin dynamics:}}
\State {$\hat \rvx_0^{(0)} =  \mathcal D(\hat \rvz_0^{(0)})$}
\For {$j = 0,\dots, N-1$}
\begin{equation}
    \hat \rvx_0^{(j+1)} \leftarrow \hat \rvx_0^{(j)} + \eta_t \Big(\nabla_{\hat \rvx_0} \log p(\hat \rvx_0^{(j)}\mid\rvz_{t_i}) + \nabla_{\hat \rvx_0} \log p(\rvy\mid\hat \rvx_0^{(j)})\Big) + \sqrt{2\eta_t} \bm\epsilon_j,\ \bm\epsilon_j \sim \mathcal N(\vzero,\mI).
\end{equation}
\EndFor
\State {$\hat \rvz_0^{(N)} = \mathcal E(\hat \rvx_0^{(N)})$}

\State {\textit{Or, latent space Langevin dynamics:}}
\For  {$j = 0,\dots, N-1$}
\begin{equation}
    \hat \rvz_0^{(j+1)} \leftarrow \hat \rvz_0^{(j)} + \eta_t \Big(\nabla_{\hat \rvz_0^{(j)}} \log p(\hat \rvz_0^{(j)}\mid\rvz_{t_i}) + \nabla_{\hat \rvz_0^{(j)}} \log p(\rvy\mid\hat \rvz_0^{(j)})\Big) + \sqrt{2\eta_t} \bm\epsilon_j,\ \bm\epsilon_j \sim \mathcal N(\vzero,\mI).
\end{equation}
\EndFor
\State Sample $\rvz_{t_{i-1}} \sim \mathcal N(\hat \rvz_0^{(N)},\sigma_{t_{i-1}}^2 \mI)$.
\EndFor
\State \textbf{Return} $\mathcal D(\rvz_0).$
\end{algorithmic}
\end{algorithm}

\newpage

\subsection{Sampling with Large-scaled Text-conditioned Latent Diffusion Models}
\label{appendix:sd}
Large-scale text-conditioned latent diffusion models (LDMs) \cite{ramesh2022hierarchicaltextconditionalimagegeneration, rombach2022high} provide a diverse and powerful prior for solving inverse problems. Given a text prompt $\rvc$, our goal is to sample from $p(\rvx_0 \mid \rvc, \rvy) \propto p(\rvy \mid \rvx_0) p(\rvx_0 \mid \rvc)$, where $p(\rvx_0 \mid \rvc)$ is modeled by the large-scale text-conditioned LDMs. To achieve this, we can utilize \cref{alg2} for posterior sampling.

\begin{figure}[t]
    \centering
    \includegraphics[width=\linewidth]{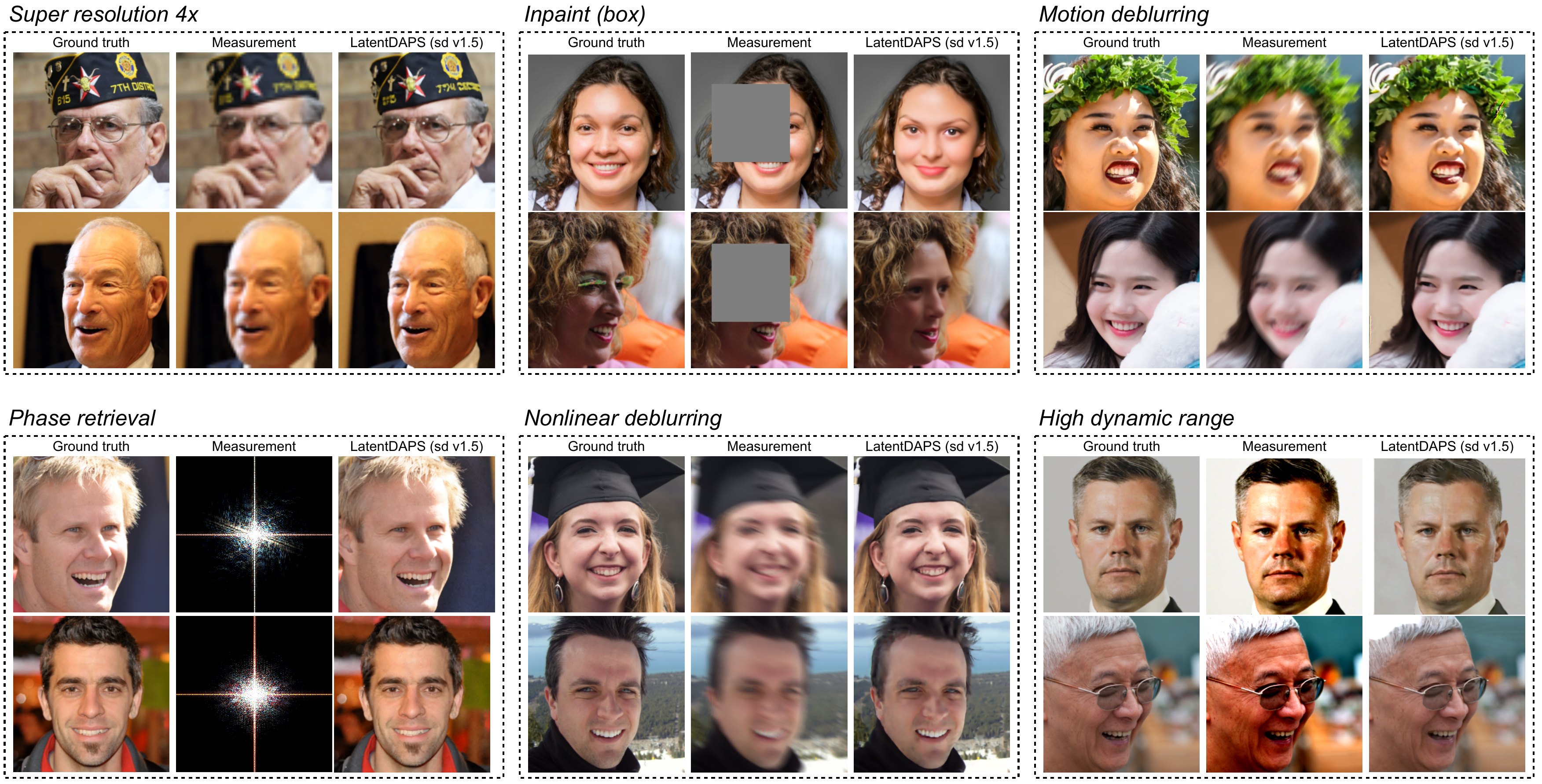}
    \caption{\textbf{Sampling results of LatentDAPS (SD v1.5) on FFHQ 256$\mathbf{\times}$256 images.} The sampling is enhanced with classifier-free guidance for text with guidance scale $7.5$. The used text prompt is shown in \cref{tab:sd-prompt}}
    \label{fig:sd-qual}
\end{figure}

\begin{figure}[t]
    \centering
    \includegraphics[width=\linewidth]{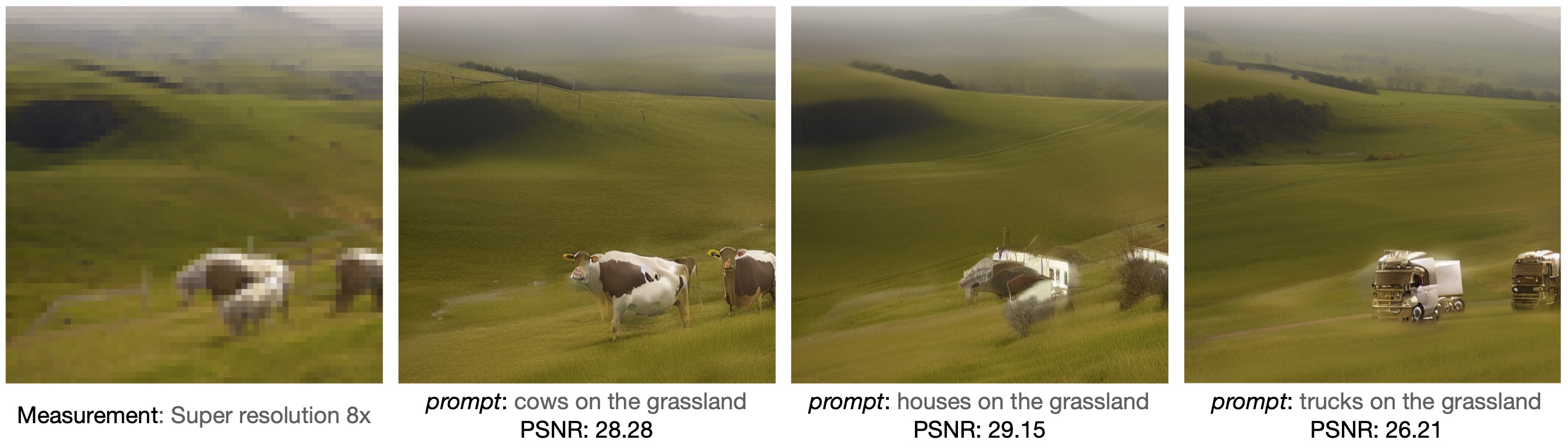}
    \caption{\textbf{Exploring different modes in the posterior distribution by text prompts.} Multiple sampling results using LatentDAPS (SD v1.5) were obtained for the super resolution 8× problem on a 512$\times$512 natural image. Text prompts provided additional control over the sampled results.}
    \label{fig:sd-text-multi}
\end{figure}

To evaluate the performance of DAPS with text-conditioned LDMs, we use Stable Diffusion v1.5 \cite{rombach2022high} and assess it on the same eight tasks from the FFHQ 256 dataset, as described in \cite{rout2023solving, rout2024beyond}. The sampling process is enhanced using classifier-free guidance for text conditioning with a guidance scale of $7.5$. The quantitative results are shown in \cref{tab:sd-quant}, and the qualitative results are presented in \cref{fig:sd-qual}. A significant advantage of text-conditioned LDMs is their ability to flexibly control posterior sampling using text prompts. As illustrated in \cref{fig:sd-text-multi}, text prompts enable effective exploration of different modes in the posterior distribution. Finally, we summarize the text prompts used for each task in \cref{tab:sd-prompt}.

\begin{table}[h]
\centering
\caption{\textbf{Quantitative evaluation on FFHQ 256$\mathbf{\times}$256 of LatentDAPS with Stable Diffusion v1.5.}. The value shows the mean over $100$ images, and all tasks assume the measurement noise level $\beta_\rvy=0.01$. Numbers for PSLD are copied from the original paper.}
\label{tab:sd-quant}
\resizebox{\linewidth}{!}{
{\small

\begin{tabular}{lcccccccccccccccc}
\toprule
Method & \multicolumn{2}{c}{Super Resolution 4×} & \multicolumn{2}{c}{Inpaint (Box)} & \multicolumn{2}{c}{Inpaint (Random)} & \multicolumn{2}{c}{Gaussian deblurring} & \multicolumn{2}{c}{Motion deblurring} &
\multicolumn{2}{c}{Phase retrieval} &
\multicolumn{2}{c}{Nonlinear deblurring} &
\multicolumn{2}{c}{High dynamic range} \\

\cline{2-3}  \cline{4-5}  \cline{6-7} \cline{8-9} \cline{10-11} \cline{12-13} \cline{14-15} \cline{16-17} \\[-8pt]
 & LPIPS$\downarrow$ & PSNR$\uparrow$ & LPIPS$\downarrow$ & PSNR$\uparrow$ & LPIPS$\downarrow$ & PSNR$\uparrow$ & LPIPS$\downarrow$ & PSNR$\uparrow$ & 
 LPIPS$\downarrow$ & PSNR$\uparrow$ & LPIPS$\downarrow$ & PSNR$\uparrow$ & LPIPS$\downarrow$ & PSNR$\uparrow$ & LPIPS$\downarrow$ & PSNR$\uparrow$ \\ 

\midrule
LatentDAPS (SD-v1.5, ours)  & $\textbf{0.109}$ & $\textbf{31.43}$ & $\textbf{0.133}$ & $23.33$ & $\textbf{0.064}$ & $\textbf{34.60}$ & $\textbf{0.160}$ & $\textbf{30.73}$ & $0.101$ & $33.84$ & $0.256$ & $27.47$ & $0.116$ & $31.67$ & $0.186$ & $24.52$\\
PSLD (SD-v1.5) & $0.201$ & $30.73$ & $0.167$ & - & $0.096$ & $30.31$ & $0.221$ & $30.10$ & - & - & - & - & -  & -  & -  & -   \\
\bottomrule
\normalsize
\end{tabular}}}

\end{table}

\newpage

\begin{table}[t]
    \centering
    \caption{\textbf{The text prompts} used in LatentDAPS (SD v1.5).}
    \label{tab:sd-prompt}
    {\small
    \begin{tabular}{cl}
        \toprule
         & Text Prompts \\
        \midrule
        FFHQ Evaluation & \textit{A natural looking human face.}\\
        \midrule
        \multirow{3}{*}{Teaser \cref{main-comparison}} & \textit{Blue butterfly on white flower, green blurred background.}\\
        &\textit{Juicy burger with toppings, fresh fries, blurred restaurant background.}\\
        &\textit{Sunlit mountain reflected in a serene lake, surrounded by trees.}\\
        \bottomrule
    \end{tabular}}

\end{table}

\newpage

\section{DAPS with Discrete Diffusion Models}
\label{appendix:discrete}

Discrete diffusion models~\cite{austin2021structured,campbell2022continuous,lou2023discrete} are designed to generate categorical data over a finite support. Similar to diffusion models in the continuous space, discrete diffusion models evolve a family of distributions $p_0(\rvx), \dots, p_T(\rvx)$ according to a continuous-time Markov chain. Specifically, the forward diffusion process is defined as
\begin{equation}
\label{eq:ctmc}
    \frac{\mathrm dp_t}{\mathrm dt} = \mQ_t p_t,
\end{equation}
where $\mQ_t$ are predefined transition matrices, such that $p_t$ converges to a simple distribution like uniform distribution or a special ``mask'' state, as $t \to \infty$. To reverse the forward process, it suffices to learn the ``concrete score'', $s(\rvx,t) = \Big[\frac{p_y(\rvy)}{p_t(\rvx)}\Big]_{\rvy\neq \rvx}$.
The reverse diffusion process is given by
\begin{equation}
\label{eq:reverse_discrete}
    \frac{\mathrm dp_{T-t}}{\mathrm dt} = \overline \mQ_{T-t} p_{T-t},
\end{equation}
where $\overline \mQ_t(\rvy,\rvx) = s(\rvx,t) \mQ_t(\rvx, \rvy)$ and $\overline \mQ_t(\rvx,\rvx) = - \sum_{y\neq x}\overline \mQ_t(\rvy,\rvx)$.

Given the similarity of continuous and discrete diffusion models, we find that DAPS can be extended to perform posterior sampling with discrete diffusion models. Instead of making the \textit{Gaussian approximation} as in continuous diffusion models, we approximate $p(\rvx_0\mid \rvx_t)$ with an exponential distribution over Hamming distance, i.e.,
\begin{equation}
    p(\rvx_0\mid \rvx_t) \approx \exp( - \|\rvx_0 - \hat \rvx_0(\rvx_t)\|_0 / r_t),
\end{equation}
where $r_t$ is determined heuristically.
We put the DAPS sampling algorithm for discrete diffusion models as in \cref{alg3}.

\begin{algorithm}[t]
\caption{Decoupled Annealing Posterior Sampling (DAPS) with Discrete Diffusion Models}
\label{alg3}
\begin{algorithmic}

\Require Score model $\vs_{\vtheta}$, measurement $\rvy$, $(t_i)_{i\in \{0,\dots, N_A\}}$.
\State Sample $\rvx_{T} \sim p_T$.
\For {$i = N_A, N_A-1, \dots, 1$}
\State {Compute $\hat \rvx_0^{(0)}=\hat \rvx_0(\rvx_{t_i})$ by solving the reverse continuous-time Markov chain in \cref{eq:reverse_discrete} with $\vs_\vtheta$}
\For {$j = 0,\dots, N-1$}
\State {\textit{Metropolis Hasting}}
\begin{equation*}
    \hat \rvx_0^{(j+1)}  \sim p(\rvx_0\mid \rvx_t, \rvy) \approx p(\rvy\mid \rvx_0) \exp(\|\rvx_0 - \hat\rvx_0(\rvx_{t_i})\|_0 / r_{t_i}).
\end{equation*}
\EndFor
\State Sample $\rvx_{t_{i-1}} \sim p(\rvx_{t_{i-1}}\mid \hat\rvx_0^{(N)})$ following \cref{eq:ctmc}.
\EndFor
\State \textbf{Return} $\rvx_0$
\end{algorithmic}
\end{algorithm}

\paragraph{Experiments on discretized MNIST dataset.}

We conduct experiments on the discretized MNIST dataset to demonstrate how DAPS can be applied to inverse problems on categorical data. We first discretize and flatten MNIST data to binary strings. We consider two inverse problems: 1) inpainting and 2) XOR operator. We mask out $50\%$ pixels for the inpainting task. For the XOR task, we draw $50\%$ random pairs from the MNIST binary strings and compute XOR over all the pairs, which serves as a highly nonlinear test case. We compare DAPS with best-of-N samples, and a recent work on discrete diffusion posterior sampling (SVDD-PM).


As shown in \cref{tab:discrete}, DAPS with discrete diffusion models is able to achieve very high classification accuracy on both inverse problem tasks, outperforming both baselines by a large margin. This suggests the effectiveness of DAPS in solving inverse problems for categorical data. We leave further exploration of DAPS in discrete-state space for future work.
\begin{table}[ht]
    \centering
    \caption{\textbf{Quantitative results on discretized MNIST on two discrete inverse problems.} Both SVDD-PM and best-of-N use 20 particles for sampling. The classification accuracy is computed using a simple ConvNet model trained using MNIST training dataset.}
    \label{tab:discrete}
    {\small
    \begin{tabular}{lccccc}
    \toprule
     & \multicolumn{2}{c}{Inpainting} & & \multicolumn{2}{c}{XOR} \\
     \cline{2-3} \cline{5-6}\\[-10pt]
     & PSNR $\uparrow$ & Accuracy ($\%$) $\uparrow$ & & PSNR $\uparrow$ & Accuracy ($\%$) $\uparrow$ \\
     \midrule
      DAPS   & $\textbf{18.82}_{\pm 2.03}$ & $\textbf{97.0}$ & & $\textbf{20.50}_{\pm 6.40}$ &  $\textbf{98.0}$\\
      SVDD-PM &  $11.84_{\pm 2.56}$ & $38.0$ & & $13.00_{\pm 2.88}$ & $59.0$\\
      Best-of-N   & $10.56_{\pm 1.11}$ &   $37.0$ & & $10.50_{\pm 1.13}$ & $36.0$\\
      \bottomrule
    \end{tabular}}

\end{table}

\newpage
\section{Proof for Propositions}
\label{appendix:proof}
\begingroup
\def\thetheorem{\ref{prop:1}}
\begin{prop}[Restated]
    Suppose $\rvx_{t_1}$ is sampled from the measurement conditioned time-marginal $p(\rvx_{t_1}\mid\rvy)$, then
    \begin{equation}
        \rvx_{t_2} \sim  \mathbb E_{\rvx_0 \sim p(\rvx_0\mid \rvx_{t_1}, \rvy)}\mathcal N( \rvx_0, \sigma_{t_2}^2 \mI) \label{eq:prop_appendix}
    \end{equation}
    satisfies the measurement conditioned time-marginal $p(\rvx_{t_2}\mid \rvy)$.
\end{prop}
\addtocounter{theorem}{-1}
\endgroup
\begin{proof}
We first factorize the measurement conditioned time-marginal $p(\rvx_{t_2}\mid\rvy)$ by
    \begin{align}
    p(\rvx_{t_2}\mid\rvy) &= \iint  p(\rvx_{t_2}, \rvx_0, \rvx_{t_1} \mid\rvy) \mathrm d\rvx_0 \mathrm d\rvx_{t_1}\\
    &= \iint p(\rvx_{t_1} \mid \rvy) p(\rvx_0\mid\rvx_{t_1}, \rvy) p(\rvx_{t_2}\mid\rvx_0, \rvx_{t_1}, \rvy)\mathrm d\rvx_0 \mathrm d\rvx_{t_1}.
\end{align}
Recall the probabilistic graphical model in \cref{fig:factor_graph}. $\rvx_{t_2}$ is independent of $\rvx_{t_1}$ and $\rvy$ given $\rvx_0$. Therefore,
\begin{equation}
    p(\rvx_{t_2}\mid\rvx_0, \rvx_{t_1}, \rvy) = p(\rvx_{t_2}\mid\rvx_0).
\end{equation}
As a result,

\begin{align}
        p(\rvx_{t_2}\mid\rvy) &= \iint p(\rvx_{t_1} \mid \rvy) p(\rvx_0\mid\rvx_{t_1}, \rvy) p(\rvx_{t_2}\mid\rvx_0)\mathrm d\rvx_0 \mathrm d\rvx_{t_1}\\
    &= \mathbb E_{\rvx_{t_1} \sim p(\rvx_{t_1}\mid\rvy)} \mathbb E_{\rvx_0 \sim p(\rvx_0\mid \rvx_{t_1}, \rvy)} p(\rvx_{t_2}\mid\rvx_0)\\
    &= \mathbb E_{\rvx_0 \sim p(\rvx_0\mid \rvx_{t_1}, \rvy)} \mathcal N(\rvx_{t_2};  \rvx_0, \sigma_{t_2}^2 \mI),
\end{align}
given $\rvx_{t_1}$ is drawn from the measurement conditioned time-marginal $p(\rvx_{t_1}\mid y)$.
\end{proof}

\begingroup
\def\thetheorem{\ref{prop:2}}
\begin{prop}[Restated]
    Suppose $\rvz_{t_1}$ is sampled from the measurement conditioned time-marginal $p(\rvz_{t_1}\mid\rvy)$, then
    \begin{equation}\label{eq:pixel-langevin}
        \rvz_{t_2} \sim \mathbb E_{\rvx_0 \sim p(\rvx_0\mid\rvz_{t_1}, \rvy)}\mathcal N( \mathcal E(\rvx_0), \sigma_{t_2}^2 \mI)
    \end{equation}
    satisfies the measurement conditioned time-marginal $p(\rvz_{t_2}\mid\rvy)$.
    Moreover,
    \begin{equation}\label{eq:latent-langevin}
        \rvz_{t_2} \sim \mathbb E_{\rvz_0 \sim p(\rvz_0\mid\rvz_{t_1}, \rvy)} \mathcal N(\rvz_0,\sigma_{t_2}^2 \mI).
    \end{equation}
    also satisfies the measurement conditioned time-marginal $p(\rvz_{t_2}\mid\rvy)$.
\end{prop}
\addtocounter{theorem}{-1}
\endgroup
\begin{proof}
    We first factorize the measurement conditioned time-marginal $p(\rvz_{t_2}\mid\rvy)$ by
\begin{align}
    p(\rvz_{t_2}\mid\rvy) & = \iint p(\rvz_{t_2}, \rvx_0, \rvz_{t_1}\mid\rvy) \mathrm d\rvx_0 \mathrm d \rvz_{t_1}\\
    &= \iint p(\rvz_{t_1}\mid\rvy) p(\rvx_0\mid\rvz_{t_1}, \rvy) p(\rvz_{t_2}\mid\rvx_0, \rvz_{t_1}, y) \mathrm d\rvx_0 \mathrm d \rvz_{t_1}.
\end{align}
Recall the probabilistic graphical model in \cref{fig:factor_graph_latent}. $\rvz_{t_2}$ is independent of $\rvz_{t_1}$ and $\rvy$ given $\rvz_0$, while $\rvz_0$ is determined only by $\rvx_0$. Therefore,

\begin{equation}
    p(\rvz_{t_2}\mid\rvx_0, \rvz_{t_1}, \rvy) = p(\rvz_{t_2}\mid\rvx_0) = \mathcal N(\rvz_{t_2}; \mathcal E(\rvx_0),\sigma_{t_2}^2 \mI).
\end{equation}
Hence,
\begin{align}
        p(\rvz_{t_2}\mid\rvy) &= \iint p(\rvz_{t_1} \mid \rvy) p(\rvx_0\mid\rvz_{t_1}, \rvy) p(\rvz_{t_2}\mid\rvx_0)\mathrm d\rvx_0 \mathrm d\rvx_{t_1}\\
    &= \mathbb E_{\rvz_{t_1} \sim p(\rvz_{t_1}\mid\rvy)} \mathbb E_{\rvx_0 \sim p(\rvx_0\mid \rvz_{t_1}, \rvy)} p(\rvz_{t_2}\mid\rvx_0)\\
    &= \mathbb E_{\rvx_0 \sim p(\rvx_0\mid \rvz_{t_1}, \rvy)} \mathcal N(\rvz_{t_2};  \mathcal E(\rvx_0), \sigma_{t_2}^2 \mI),
\end{align}
assuming $\rvz_{t_1}$ is drawn from $p(\rvz_{t_1}\mid\rvy)$.

Moreover, we can also factorize $p(\rvz_{t_2}\mid\rvy)$ by
\begin{align}
    p(\rvz_{t_2}\mid\rvy) &= \iint p(\rvz_{t_2},\rvz_{t_1}, \rvz_0\mid\rvy) \mathrm d\rvz_0 \mathrm d\rvz_{t_1}\\
    &= \iint p(\rvz_{t_1}\mid\rvy) p(\rvz_0\mid\rvz_{t_1},\rvy) p(\rvz_{t_2}\mid\rvz_0,\rvz_{t_1},\rvy) \mathrm d\rvz_0 \mathrm d\rvz_{t_1}\\
    &= \iint p(\rvz_{t_1}\mid\rvy) p(\rvz_0\mid\rvz_{t_1},\rvy) p(\rvz_{t_2}\mid\rvz_{t_1}) \mathrm d\rvz_0 \mathrm d\rvz_{t_1}.
\end{align}
The last equation is again derived directly from \cref{fig:factor_graph_latent}. Given that $\rvz_{t_1}$ is sampled from $p(\rvz_{t_1}\mid\rvy)$, we have that
\begin{equation}
    p(\rvz_{t_2}\mid\rvy) = \mathbb E_{\rvz_0 \sim p(\rvz_0\mid\rvz_{t_1},\rvy)} \mathcal N(\rvz_{t_2}; \rvz_0,\sigma_{t_2}^2 \mI).
\end{equation}
\end{proof}
\newpage 

\section{Discussions}
\subsection{Sampling Efficiency}
\label{appendix:efficiency}
The sampling efficiency is a crucial aspect of inverse problem solvers. The time cost of diffusion model-based methods is highly dependent on the number of neural function evaluations (NFE). Here in \cref{tab:time} we show the NFE of the default setting of some pixel space baseline methods and DAPS with different configurations. In \cref{fig:nfe}, we show the quantitative evaluation of DAPS with different NFE. As we can see, DAPS can achieve relatively much better performance than baselines with small NFE.

\begin{table}[H]
    \centering
     \caption{\textbf{Sampling time of DAPS on phase retrieval task with FFHQ 256.} The nonparallel single image sampling time on the FFHQ 256 dataset with 1 NVIDIA A100-SXM4-80GB GPU. The time depends may differ slightly in different runs.}
    \label{tab:time}
    {\small
    \begin{tabular}{l|c|c|c|c}
    \toprule
      Configuration  & ODE Steps & Annealing Steps & NFE & Seconds/Image \\
    \midrule 
    DPS & - & - & 1000 & 35 \\
    DDRM & - & - & 20 & 2\\
    RED-diff & - & - & 1000 & 47 \\ 
    \midrule
       DAPS-50 & 2 & 25 & 50 & 4 \\
        DAPS-100 & 2 & 50 & 100 & 7 \\
        DAPS-200 & 2 & 100 & 200 &  13\\
        DAPS-400 & 4 & 100 & 400 & 17 \\
        DAPS-1K & 5 & 200 & 1000 & 37 \\
        DAPS-2K & 8 & 250 & 2000 & 61 \\
        DAPS-4K & 10 & 400 & 4000 & 108 \\
        
    \bottomrule
    \end{tabular}}
   
\end{table}

\subsection{Limitations and Future Extension}
\label{appendix:limit}
Though DAPS achieves significantly better performance on inverse problems like phase retrieval, there are still some limitations. 

First, we only adopt a very naive implementation of the latent diffusion model with DAPS, referred to as LatentDAPS. However, some recent techniques~\cite{rout2023solving, song2024solving} have been proposed to improve the performance of posterior sampling with latent diffusion models. Specifically, one main challenge is that $\rvx_{0\mid \rvy}$ obtained by Langevin dynamics in pixel space might not lie in the manifold of clean images. This could further lead to a sub-optimal performance for autoencoders in diffusion models since they are only trained with clean data manifold. 

Furthermore, we only implement DAPS with a decreasing annealing scheduler, but the DAPS framework can support any scheduler function $\sigma_t^A$ as long as $\sigma_0^A = 0$. A non-monotonic scheduler has the potential of providing DAPS with more power to explore the solution space. 

Finally, we utilize fixed NFE for the ODE solver. However, one could adjust it automatically. For example, less ODE solver NFE for smaller $t$ in later sampling steps. We would leave the discussions above as possible future extensions.

\subsection{Broader Impacts}\label{sec:impacts}
We anticipate that DAPS can offer a new paradigm for addressing challenging real-world inverse problems using diffusion models. DAPS tackles these problems by employing a diffusion model as a general denoiser, which learns to model a powerful prior data distribution. This approach could significantly enrich the array of methods available to the inverse problem-solving community. However, it is important to note that DAPS might generate biased samples if the diffusion model is trained on biased data. Therefore, caution should be exercised when using DAPS in bias-sensitive scenarios.

\newpage

\section{Experimental Details}
\label{appendix:details}

\subsection{Inverse Problem Setup}
Most inverse problems are implemented in the same way as introduced in \cite{chung2023diffusion}. However, for inpainting with random pixel masks, motion deblurring, and nonlinear deblurring, we fix a certain realization for fair comparison by using the same random seeds for mask generation and blurring kernels. Moreover, for phase retrieval, we adopt a slightly different version as follows:
\begin{equation}
    \rvy\sim \mathcal N(|\mathbf F\mathbf P(0.5\rvx_0+0.5)|,\beta_\rvy^2\mI),
\end{equation}
which normalize the data to lies in range $[0, 1]$ first. Here $\mathbf F$ and $\mathbf P$ are discrete Fourier transformation matrices and oversampling matrices with ratio $k/n$. Same as \cite{chung2023diffusion}, we use an oversampling factor $k=2$ and $n=8$. We normalize input $x_0$ by shifting its data range from $[-1, 1]$ to $[0, 1]$ to better fit practical settings, where the measured signals are usually non-negative.

The measurement for high dynamic range reconstruction is defined as
\begin{equation}
    \rvy \sim \mathcal N(\mathrm{clip}(\alpha \rvx_0, -1,1), \beta_\rvy^2 \mI),
\end{equation}
where the scale $\alpha$ controls the distortion strength. We set $\alpha = 2$ in our experiments. 

\subsection{DAPS Implementation Details}
\paragraph{Euler ODE Solver}\label{appendix:ode-solver}
For any given increasing and differentiable noisy scheduler \(\sigma_t\) and any initial data distribution \(p(\rvx_0)\), we consider the forward diffusion SDE \( \mathrm d\rvx_t = \sqrt{2\dot{\sigma_t}\sigma_t } \, \mathrm d\rvw_t \), where \(\dot{\sigma_t}\) denotes the time derivative of \(\sigma_t\) and \(\mathrm d\rvw_t\) represents the standard Wiener process. This SDE induces a probability path of the marginal distribution \(\rvx_t\), denoted as \(p(\rvx_t;\sigma_t)\). As demonstrated in \cite{karras2022elucidating, song2021scorebased}, the probability flow ODE for the above process is given by:
\begin{equation}
    \mathrm d\rvx_t = -\dot{\sigma_t}\sigma_t  \nabla_{\rvx_t} \log p(\rvx_t;\sigma_t) \, \mathrm dt.\label{eq:pfode}
\end{equation}
By employing the appropriate preconditioning introduced in \cite{karras2021style}, we can transform the pre-trained diffusion model with parameter $\vtheta$ to approximate the score function of the above probability path: \(\vs_\vtheta(\rvx_t, \sigma_t) \approx \nabla_{\rvx_t} \log p(\rvx_t;\sigma_t)\). In DAPS, we compute $\hat\rvx_0(\rvx_t)$ by solving the ODE given $\rvx_t$ and time $t$ as initial values. 

Numerically, we use scheduler $\sigma_t=t$ and implement an Euler solver \cite{karras2022elucidating}, which evaluates $\frac{\mathrm d \rvx_t}{\mathrm d t}$ at $N_{\text{ode}}$ discretized time steps in interval $[0, t]$ and updates $\rvx_t$ by the discretized ODE. The time step $t_i$, $i=1,\cdots, N_{\text{ode}}$ are selected by a polynomial interpolation between $t$ and $t_{\min}$:
\begin{equation}\label{eq:polyint}
    t_i = \left(t^\frac{1}{\rho}+\dfrac{i}{N-1}\left(t_{\min}^{\frac{1}{\rho}}-t^\frac{1}{\rho}\right)\right)^\rho.
\end{equation}
We use $\rho=7$ and $t_{\min} = 0.02$ throughout all experiments.

\paragraph{Annealing Scheduler} To sample from the posterior distribution $p(\rvx_0\mid \rvy)$, DAPS adopts a noise annealing process to sample $\rvx_t$ from measurement conditioned time-marginals $p(\rvx_t\mid \rvy)$, where $\rvx_t$ is defined by noisy perturbation of $\rvx_0$: $\rvx_t = \rvx_0 +\sigma_t^A\bm\epsilon$, $\bm\epsilon\sim \mathcal N(\vzero, \mI)$, where $\sigma_t^A$ is the annealing scheduler. In practice, we start from time $T$, assuming $p(\rvx_{T}\mid \rvy)\approx \mathcal N(\vzero, \sigma_{\max}^2\mI)$, with $\sigma_{\max} = \sigma_{T}^A$. For simplicity, we adopt $\sigma_t^A=t$ and the same polynomial interpolation in \cref{eq:polyint} between $\sigma_0$ and $\sigma_{T}$ for total $N_{A}$ steps. 


\paragraph{Hyperparameters Overview} The hyperparameters of DAPS can be categorized into the following three categories.

(1) The ODE solver steps $N_{\text{ode}}$ and annealing scheduler $N_{A}$. These two control the total NFE of DAPS. Need to trade-off between cost and quality. For linear tasks, $N_{\text{ode}}=5$ and $N_{A}=200$. And for nonlinear tasks $N_{\text{ode}=10}$ and $N_{A} = 400$. For LatentDAPS, including the one with the Stable Diffusion model, we choose $N_{\text{ode}}=5$ and $N_{\text{A}}=50$ for linear tasks and $N_{\text{ode}}=10$ and $N_{\text{A}}=100$ for nonlinear tasks.

(2) The step size $\eta_t$ and total step $N$ in Langevin dynamics (also damping factor $\gamma_t$ in HMC). These two control the sample quality from $p(\rvx_0\mid \rvx_{t}, \rvy)$. For simplicity, we adopt a linear decay scheme $\eta_t = \eta_0[\delta+t/T(1-\delta)]$, where $\delta$ is the decay ratio and $T$ is the starting timestep. We include the final hyperparameters in \cref{tab:hyper}. Moreover, instead of using the true $\beta_\rvy=0.05$ in \cref{eq:langevin}, we regard $\beta_\rvy$ as a hyperparameter and set it to $0.01$ for better empirical performance. Moreover, we adopt HMC to speed up LatentDAPS with Stable Diffusion, where we report $\mu =1-\gamma_t\eta_t$ and $L=\eta_t^2$, which are set as fixed values for all timesteps.

(3) The $\sigma_{\max}$ and $\sigma_{\min}$ used in annealing process. We set $\sigma_{\max}=100$ and $10$ for DAPS and LatentDAPS and $\sigma_{\min}=0.1$ to make be more robust to noise in measurement.

\begin{table}[t]
    \centering
    \caption{\textbf{The hyperparameters} of experiments in paper for all tasks.}
    \label{tab:hyper}
    \resizebox{\linewidth}{!}{
    \begin{tabular}{c|c|cccccccc}
    \toprule
       \multirow{1}{*}{Algorithms} & Tasks  &  Super Resolution 4× & Inpaint (Box) & Inpaint (Random) & Gaussian deblurring & Motion deblurring & Phase retrieval & Nonlinear deblurring & High dynamic range\\
       \midrule
       \multirow{3}{*}{DAPS} & $\eta_0$ & 1e-4 & 5e-5 & 1e-4 & 1e-4&5e-5&5e-5&5e-5&2e-5\\

       &$\delta$ & 1e-2 &  1e-2 &  1e-2 &  1e-2& 1e-2& 1e-2& 1e-2& 1e-2\\

        &$N$ & 100 &100 &100 &100 &100 &100 &100 &100  \\
    \midrule
       
       \multirow{3}{*}{LatentDAPS} & $\eta_0$ &1e-4& 2e-6& 2e-6& 2e-6& 2e-6& 4e-6 & 2e-6& 6e-7 \\
        &$\delta$ & 1e-2 &  1e-2 &  1e-2 &  1e-2& 1e-2& 1e-2& 1e-2& 1e-2\\

        &$N$ & 50& 50& 50& 50& 50& 50& 50& 50  \\
       \midrule

       \multirow{3}{*}{LatentDAPS (SD v1.5)} & $L$ & 1e-4 & 1e-5 & 3e-5 &  1e-5 & 2e-5 & 7e-5 & 1e-5 & 1e-5 \\
        &$\mu$ & 0.45 &  0.60 &  0.60 &  0.90 & 0.85 & 0.70 & 0.80 & 0.70 \\

        &$N$ & 30& 20& 15& 40& 30& 100& 60 & 45  \\
       
    \bottomrule
    \end{tabular}}

\end{table}

\subsection{Baseline Details}

\paragraph{DPS} All experiments are conducted with the original code and default settings as specified in \cite{chung2023diffusion}. For high dynamic range reconstruction task, we use the $\xi_i = 1/\|\rvy-\mathcal A(\hat \rvx_0(\rvx_i)\|$

\paragraph{DDRM} We adopt the default setting of $\eta_B = 1.0$ and $\eta = 0.85$ with 20 DDIM steps as specified in \cite{kawar2022denoising}.

\paragraph{DDNM} We adopt the default setting of $\eta_B = 1.0$ and $\eta = 0.85$ with 100 DDIM steps as specified in \cite{wang2022zero}.

\paragraph{DCDP}We adopt the default setting in \cite{li2024decoupleddataconsistencydiffusion} and directly use the open-sourced code for all results.

\paragraph{FPS-SMC} We adopt the default setting of $M=20$ and $N=1000$ as specified in \cite{dou2024diffusion}.

\paragraph{DiffPIR} We adopt the same evaluation settings and report the results in \cite{zhu2023denoising}.

\paragraph{RED-diff} For a fair comparison, we use a slightly different RED-diff\cite{mardani2023variational} by initializing the algorithm with random noise instead of a solution from the pseudoinverse. This might lead to a worse performance compared with the original RED-diff algorithm. We use $\lambda=0.25$ and $lr=0.5$ for all experiments.
\paragraph{PSLD} We use the official implementation of PSLD~\cite{rout2023solving} with the default configurations. Specifically, we use Stable diffusion v1.5 for ImageNet experiments, which is commonly believed to be a stronger pre-trained model than LDM-VQ4 used in other experiments.

\paragraph{ReSample} All experiments are based on the official code of ReSample~\cite{song2024solving} with 500 steps DDIM sampler. 

\paragraph{DPnP} We direly use the reported time cost and results in DPnP~\cite{xu2024provably}.

\newpage

\section{Experiments on Synthetic Data Distributions}
\label{appendix:syn}

\cref{fig:syn} shows the sampling trajectories and predicted posterior distribution of DPS and DAPS on a synthetic data distribution. Specifically, we create a 2D Gaussian mixture as the prior distribution, \ie, $p(\rvx_0) = \frac{1}{2} \left(\mathcal N(\rvx_0; \vc_1,\mSigma_1) + \mathcal N(\rvx_0; \vc_2, \mSigma_2)\right)$. Let $\vc_1 = (-0.3, -0.4)$ and $\vc_2 = (0.6, 0.5)$, $\mSigma_1 = \mSigma_2 = \mathrm{diag}(0.01, 0.04)$. We draw 1000 samples from this prior distribution to create a small dataset, from which we can compute a closed-form empirical Stein score function at any noise level $\sigma$.
\begin{figure}[ht]
    \centering
    \includegraphics[width=\textwidth]{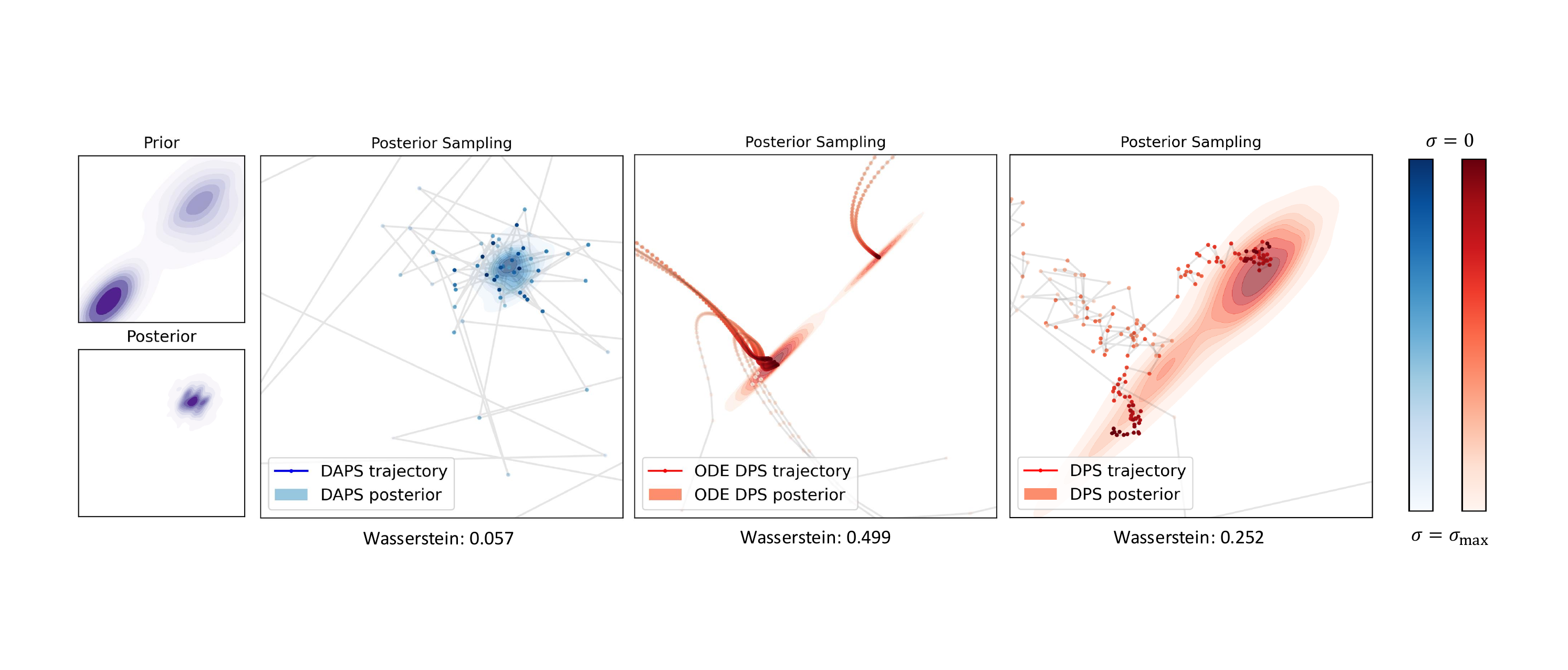}
    \caption{\textbf{DAPS and DPS (both SDE and ODE) on 2-dimensional synthetic data.} DAPS achieves much more accurate posterior sampling in terms of 2-Wasserstein distance.}
    \label{fig:syn_all}
\end{figure}

Moreover, we consider the simplest measurement function that contains two modes, \ie, $\rvy = \exp\left(-\frac{\|\rvx\|^2}{0.05}\right) + \exp\left(-\frac{\|\rvx-(0.5,0.5)\|^2}{0.05}\right) + \rvn$, where $\rvn \sim \mathcal N(\vzero,\beta_\rvy^2 \mI)$ with $\beta_\rvy = 0.3$. Let $\rvy = 0$, so that the likelihood $p(\rvy\mid\rvx_0)$ has two modes at $(0.5,0.5)$ and $(0,0)$. Since the prior distribution is large only at $(0.5,0.5)$, the posterior distribution is single-mode, as illustrated in \cref{fig:syn_all}. 
\begin{wrapfigure}{r}{0.5\textwidth}
    \centering
    \includegraphics[width=.45\textwidth]{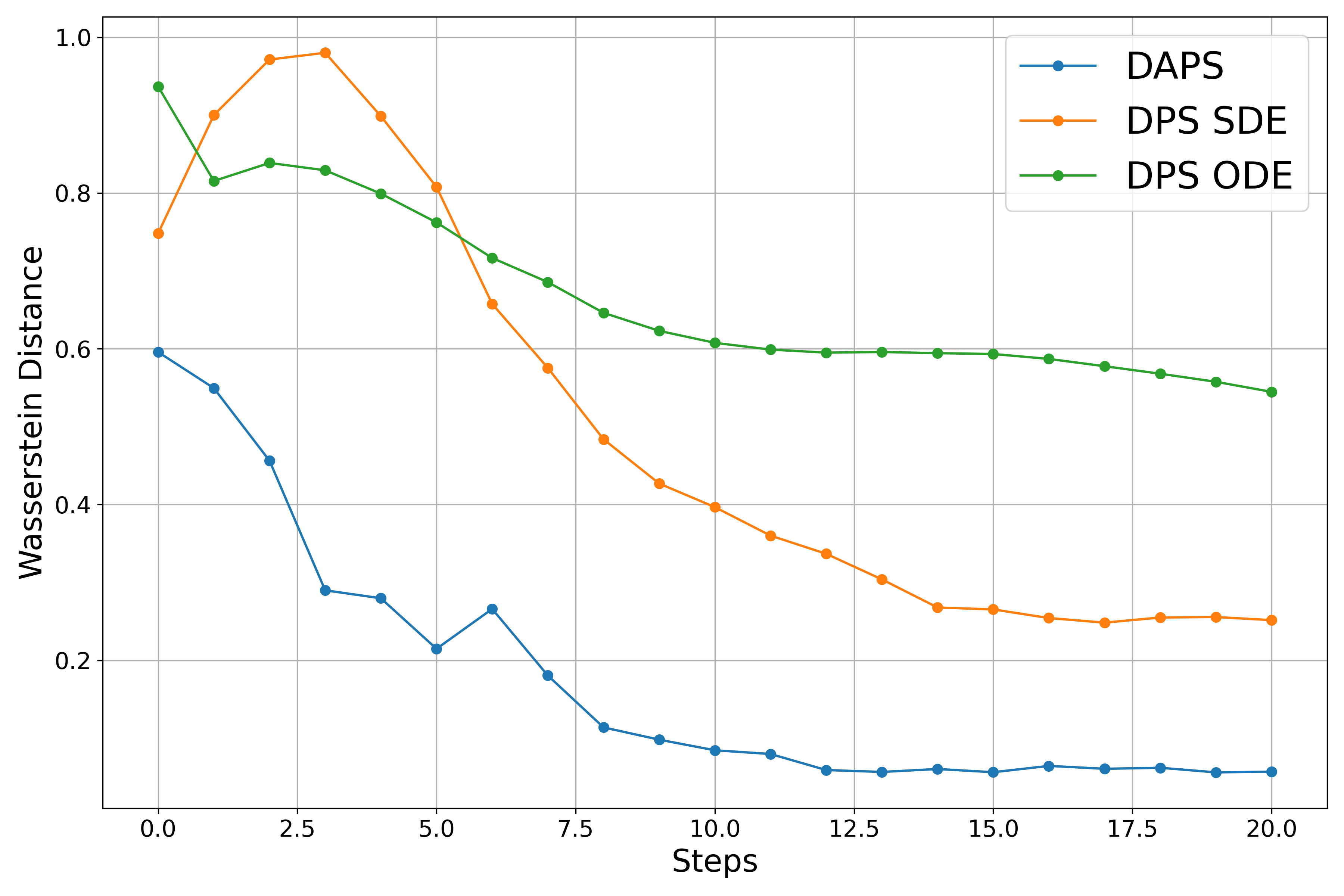}
    \caption{\textbf{Wasserstein distance} between estimated $\rvx_t \sim p(\rvx_t\mid\rvy)$ and ground truth $p(\rvx_t\mid\rvy)$.}
    \label{fig:wasserstein}
\end{wrapfigure}

We run both DPS and DAPS for 200 steps and 100 independent samples on this synthetic dataset. However, as shown in \cref{fig:syn_all}, both SDE and ODE versions of DPS converge to two different modes. This is because DPS suffers from large errors in estimating likelihood $p(\rvx_t\mid\rvy)$, especially in the early stages. These errors can hardly be corrected and are propagated along the SDE/ODE trajectory. DAPS, on the other hand, samples from a time-marginal distribution at each time step, and is able to recover the posterior distribution more accurately.

\vspace{1em}
We further investigate the performance of posterior estimation by computing the Wasserstein distance between samples $\rvx_t$ and ground truth posterior $p(\rvx_t\mid\rvy)$ for each step $t$. As shown in \cref{fig:wasserstein}, the Wasserstein distance for DAPS decreases quickly and remains small throughout the sampling process. This conforms with our theory that the distribution of $\rvx_t$ is ensured to be $p(\rvx_t\mid\rvy)$ for every noise level $\sigma_t$.

\newpage

\section{Experimental Results on Compressed Sensing Multi-coil MRI}
\label{appendix:mri}

Compressed Sensing Multi-coil magnetic resonance imaging (CS-MRI) is a medical imaging problem that aims to shorten the acquisition time of MRI scanning by subsampling. Specifically, CS-MRI takes in only a subset of the measurement space (k-space) and solves an inverse problem to reconstruct the whole source image in high resolution. Suppose the underlying source image is $\rvx_0 \in \mathbb C^n$. The forward function of CS-MRI can be written as
\begin{equation}
    \rvy = \mP \mF \mS \rvx_0 + \rvn,
\end{equation}
where $\mP$ is a subsampling operator with only zeros and ones on its diagonal, $\mF$ is the Fourier transform matrix, and $\mS$ is the multi-coil sensitivity map so that $\mS = [\mS_1,\dots, \mS_c]$ for $c$ coils.

\paragraph{Experimental setup.} We preprocess the fastMRI dataset~\cite{fastmri} by calculating the magnitude images of the minimal variance unbiased estimator (MVUE), and resize them into $320\times 320$ grayscale images. This pipeline is the same as~\cite{jalal2021robust}. We use a diffusion model trained with EDM framework~\cite{karras2022elucidating} on the preprocessed data. We compare DAPS with DPS~\cite{chung2023diffusion} and DiffPIR~\cite{zhu2023denoising} designed for general image restoration tasks, and also ScoreMRI~\cite{chung2022score} and CSGM~\cite{jalal2021robust} which are specifically designed for solving MRI with diffusion models. 

\begin{figure}[ht]
    \centering
    \includegraphics[width=\linewidth]{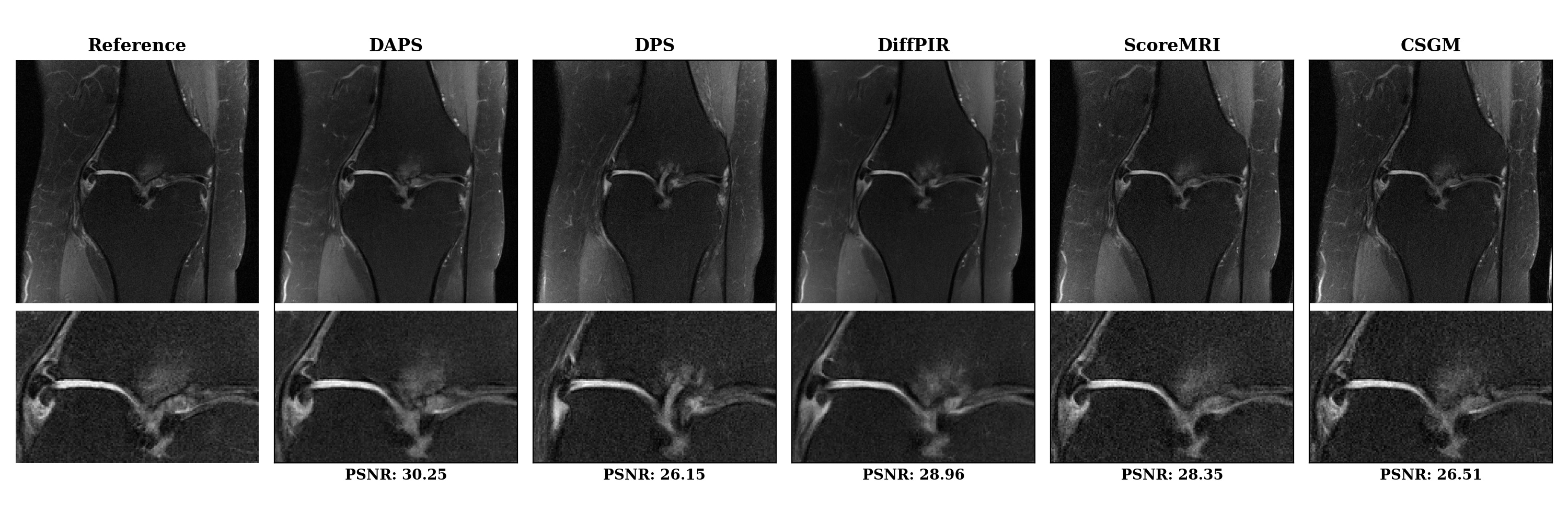}
    \includegraphics[width=\linewidth]{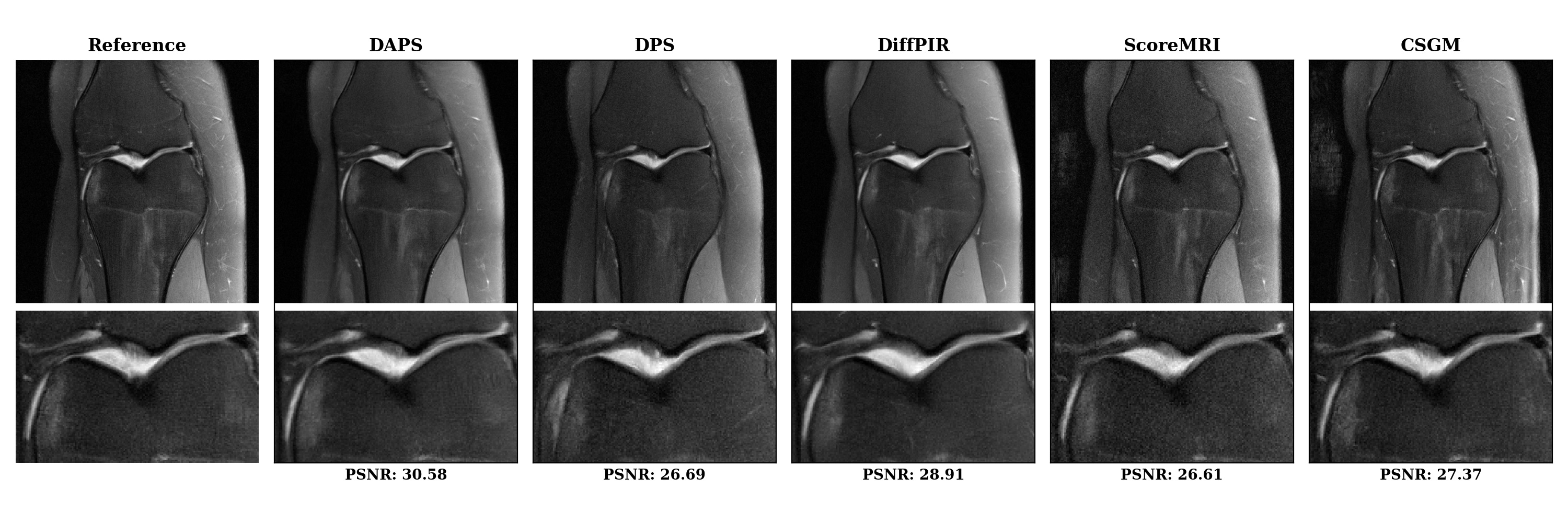}
    \label{fig:mri}
    \vspace{-20pt}
    \caption{\textbf{Qualitative results} of inverse problem solvers with diffusion models on CS-MRI.}
\end{figure}

\paragraph{Results.} We include the quantitative results in~\cref{tab:MRI} in the main text. Here, we provide some visual results of the reconstructions in~\cref{fig:mri}, which validates the capability of DAPS in solving real-world medical imaging inverse problems.

\newpage
\section{Additional Results}
\label{appendix:addition}

\subsection{More Ablation Study}
\label{appendix:ablation}
\paragraph{Effectiveness of different MCMC samplers.}
We conduct a detailed comparison of the three proposed MCMC samplers in DAPS. To represent linear and nonlinear inverse problems, we select super resolution 4× and high dynamic range, evaluating performance using FFHQ 256 images. Hyperparameter tuning is performed on 5 validation images, and results are reported on 10 test images. The comparison is illustrated in \cref{tab:mcmc-res}. HMC achieves the best balance between time efficiency and generation quality among the methods. Langevin dynamics, while simpler to implement and requiring fewer tunable hyperparameters, delivers competitive results. Although Metropolis Hastings shows slightly inferior performance, it has the advantage of being extendable to problems where gradient information is unavailable.

\begin{table}[ht]
    \centering
    \caption{\textbf{Quantitative comparison on different MCMC samplers.} HMC achieves the best balance between time efficiency and generation quality among the methods.}
    \label{tab:mcmc-res}
    \begin{tabular}{lcccccc}
    \toprule
     & \multicolumn{2}{c}{Super resolution 4×} & & \multicolumn{2}{c}{High dynamic range} & \multirow{2}{*}{Number of forward function callings} \\
     \cline{2-3} \cline{5-6}\\[-10pt]
     & PSNR $\uparrow$ & LPIPS $\downarrow$  & & PSNR $\uparrow$ & LPIPS $\downarrow$ &  \\
     \midrule
      DAPS+HMC  & $\textbf{28.17}$ & $\textbf{0.160}$ & & $\textbf{27.68}$ &  $0.163$ &  $10$\\
      DAPS+Langevin dynamics &  $28.15$ & $0.166$ & & $27.57$ & $\textbf{0.162}$ & 100 \\
      DAPS+Metropolis hasting   & $27.46$ &   $0.192$ & & $25.21$ & $0.224$ & 50000\\
      \bottomrule
    \end{tabular}
\end{table}

\paragraph{Effectiveness of the number of function evaluations.}
To better understand how the number of function evaluations (NFE) of the diffusion model influences the performance, we evaluate the performance of DAPS with different configurations. Recall that we use an ODE sampler in each inner loop to compute $\hat \rvx_0(\rvx_t)$, the total NFE for DAPS is the number of inner ODE steps times the number of noise annealing steps. We evaluate DAPS using NFE ranging from 50 to 4k, with configurations as specified in \cref{appendix:efficiency}. As indicated by \cref{fig:nfe}, DAPS achieves relatively decent performance even with small NFE.

\begin{figure}[ht]
    \centering
    \includegraphics[width=\textwidth]{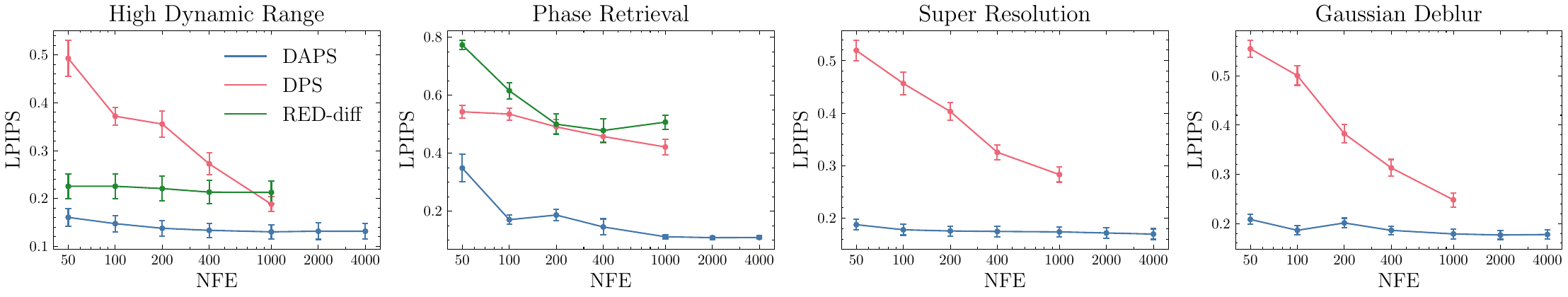}
    \caption{\textbf{Quantitative evaluations of image quality for different number of function evaluations (NFE).} Experiments are conducted on the FFHQ 256 dataset for four different tasks.}
    \label{fig:nfe}
\end{figure}

\begin{figure}[ht]
\centering
        \includegraphics[width=.7\textwidth]{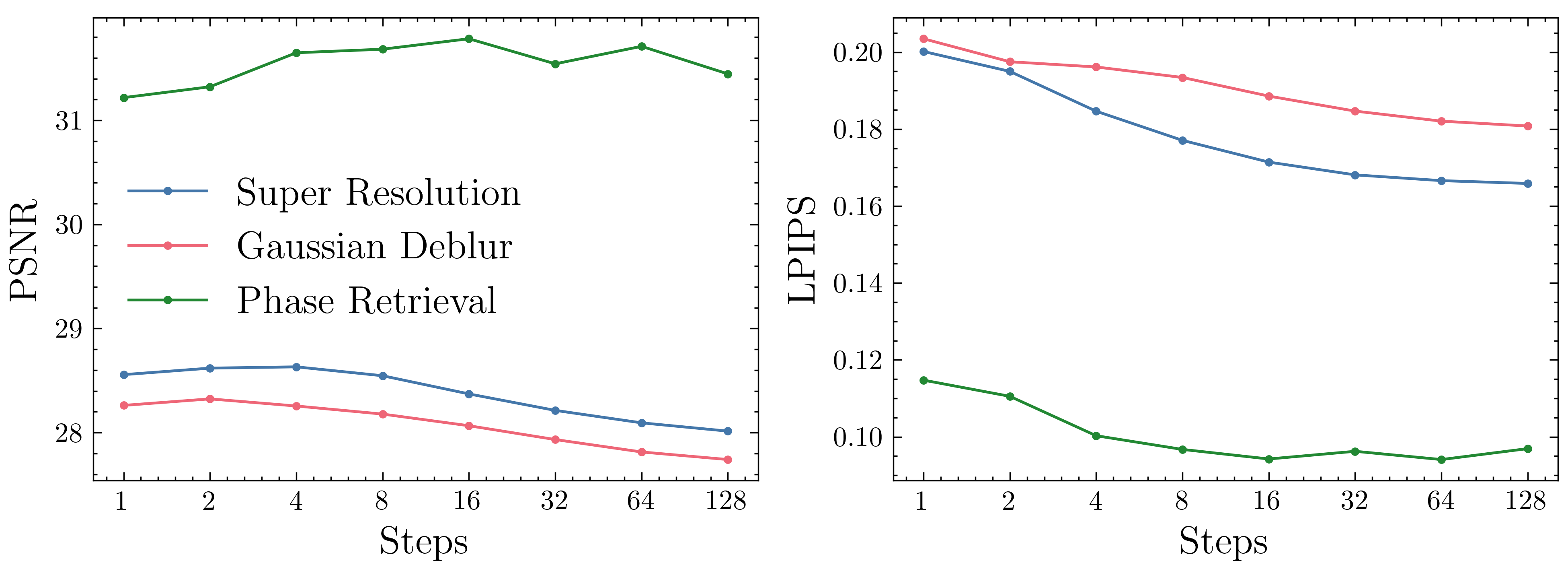}
    \caption{The effect of the \textbf{number of ODE steps} for denoisers.}
    \label{fig:ode_ablation}
\end{figure}

\begin{figure}[ht]
    \centering
    \begin{subfigure}[b]{.49\textwidth}
        \includegraphics[width=\textwidth]{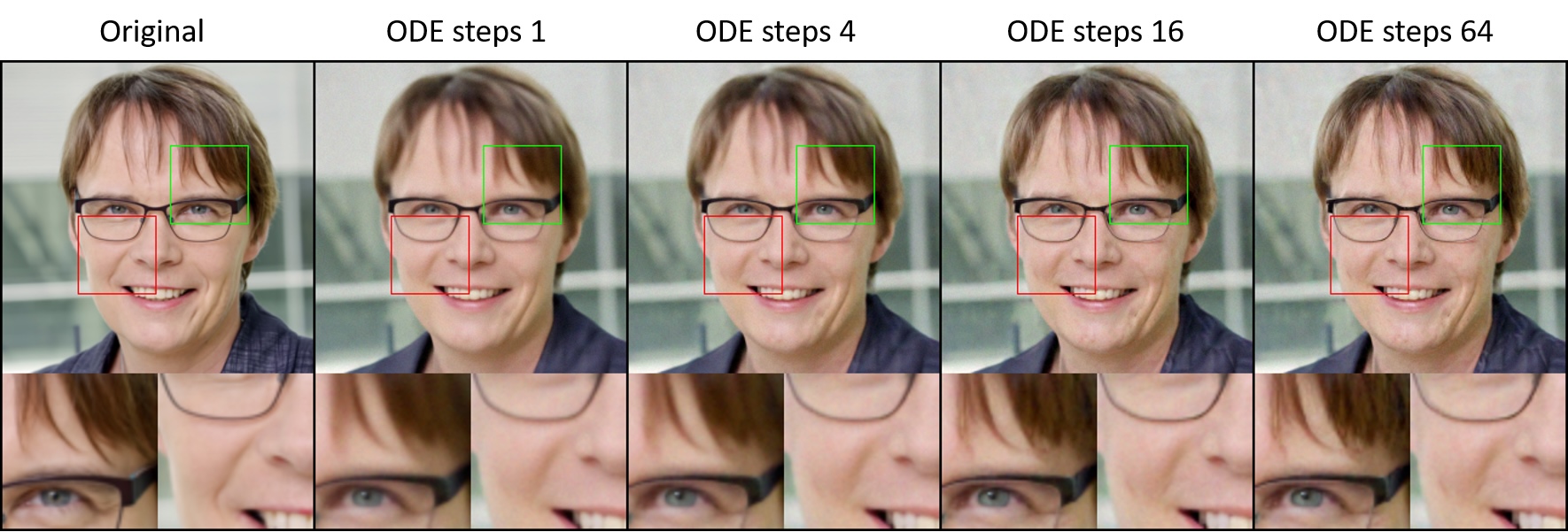}
    \end{subfigure}
    \begin{subfigure}[b]{.49\textwidth}
        \includegraphics[width=\textwidth]{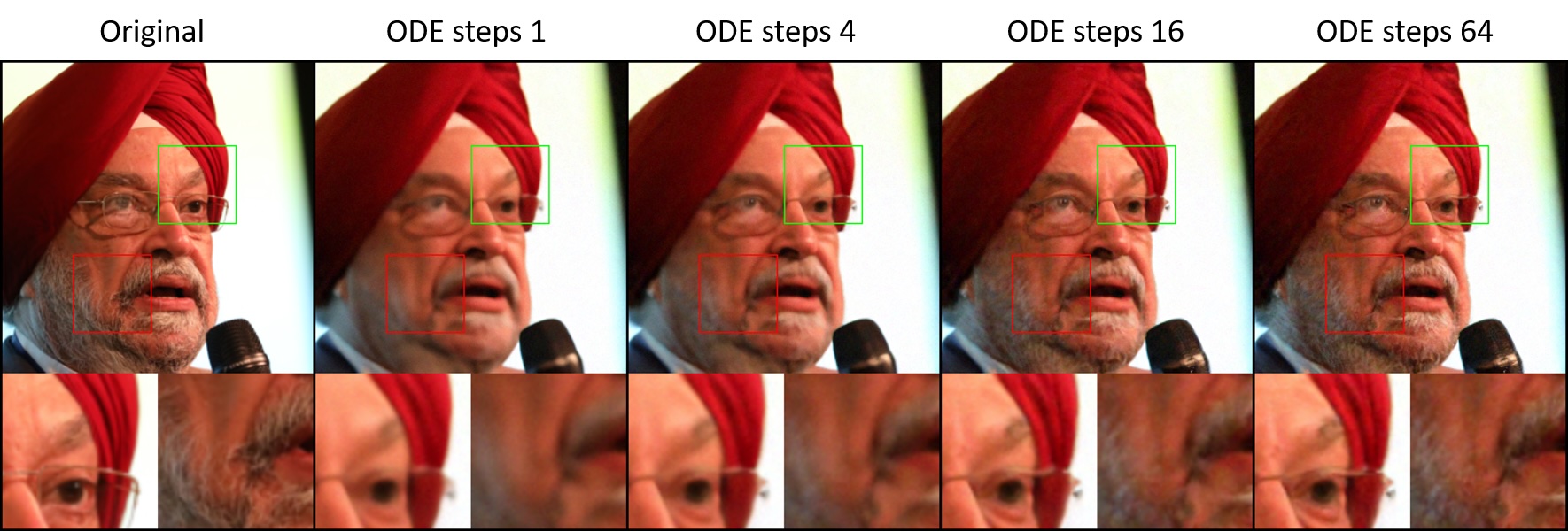}
    \end{subfigure}
    \caption{\small \textbf{Qualitative ablation studies on the number of ODE steps}. We run DAPS with different numbers of ODE steps on super-resolution 4× task. More details are observed as the number of ODE steps increases.}
    \label{fig:ode_step}
\end{figure}

\paragraph{Effectiveness of the number of ODE steps.}
Recall that we use an ODE sampler to compute $\hat \rvx_0(\rvx_t)$, the estimated mean of the approximated distribution $p(\rvx_0\mid \rvx_t)$. We use the same number of function evaluations in our ODE sampler throughout the entire algorithm. To test how the number of function evaluations (NFE) in the ODE sampler influences the performance, we try different NFE on two linear tasks and one nonlinear task. In particular, when NFE is $1$, the ODE sampler is equivalent to computing $\mathbb E[\rvx_0\mid\rvx_t]$ via Tweedie's formula. As shown in \cref{fig:ode_ablation}, increasing NFE in the ODE sampler consistently improves the overall image perceptual quality, \MOD{but also at the cost of a slightly lower PSNR}. \MOD{This trade-off between PSNR and LPIPS is also observed in \cite{mardani2023variational}. Perceptually, we notice that increasing ODE steps adds more fine-grained details to the produced images, which improves LPIPS but decreases PSNR. This finding is corroborated by~\cref{fig:ode_step}, where the reconstructed images appear less blurry and show high-frequency details as the number of ODE steps increases.}

\begin{figure}[ht]
\centering
\includegraphics[width=.7\textwidth]{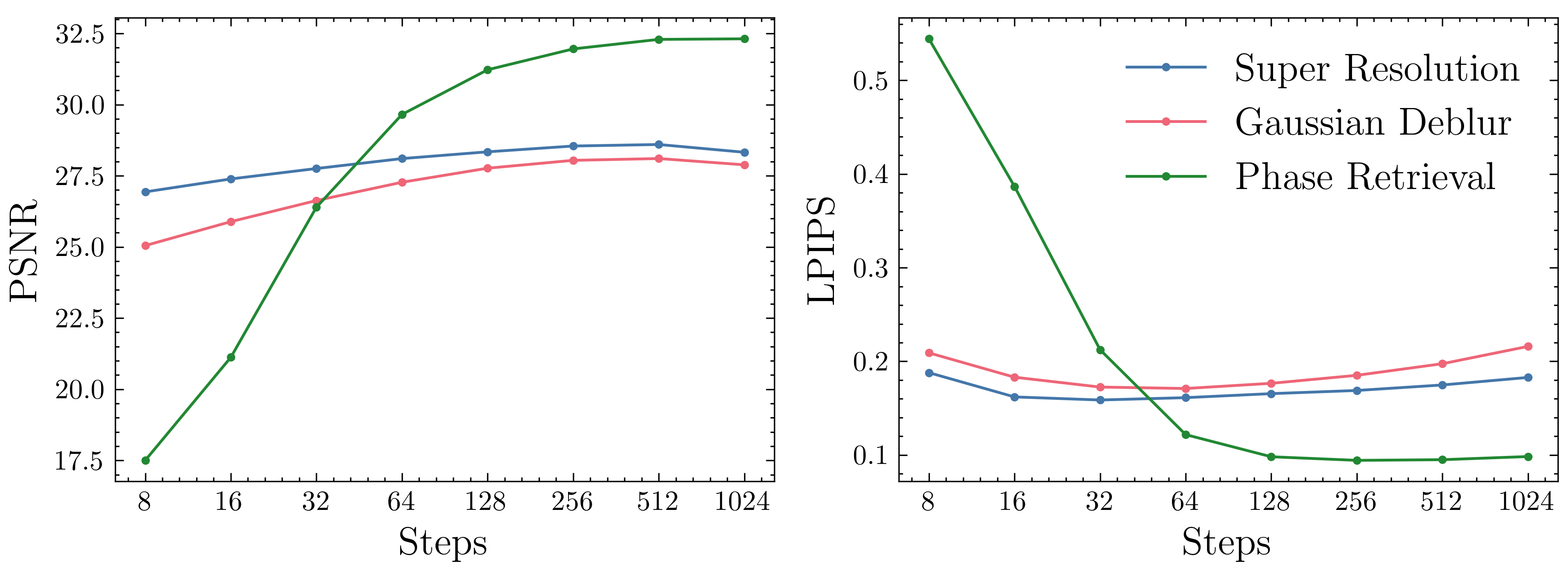}
    \caption{The effect of the number of \textbf{annealing noise scheduling steps}.}
    \label{fig:sampler_ablation}
\end{figure}

\paragraph{Effectiveness of annealing noise scheduling step.}
To better understand how the scheduling of sigma influences performance, we also evaluate the effects of sampling with varying noise scheduling steps. A larger number of scheduling steps implies a denser discretization grid between $\sigma_{\max}$ and $\sigma_{\min}$. The quantitative results are shown in \cref{fig:sampler_ablation}. The performance of DAPS on linear tasks slightly increases as the number of annealing noise scheduling steps increases, while its performance on nonlinear tasks (\eg, phase retrieval) increases dramatically with the number of scheduling steps. However, DAPS achieves a near-optimal sample quality when the number of noise scheduling steps is larger than 200.

\begin{figure}[ht]
\centering
    \includegraphics[width=.7\textwidth]{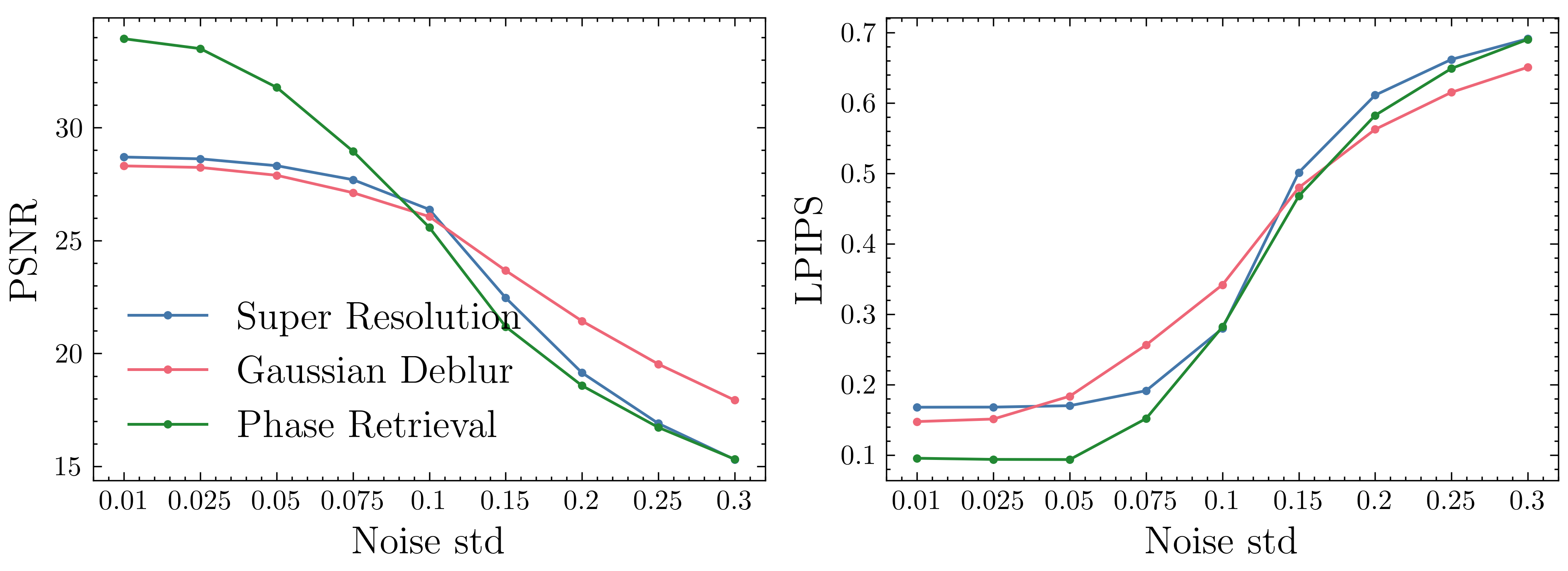}
    \caption{The effect of the \textbf{measurement noise level} $\beta_\rvy$.}
    \label{fig:meas_sigma}
\end{figure}

\paragraph{Different Measurement Noise Level}
When subjected to varying levels of measurement noise, the quality of solutions to inverse problems can differ significantly. To evaluate the performance of DAPS under different noise conditions, we present the results in \cref{fig:meas_sigma}. DAPS is robust to small noise levels ($\sigma < 0.05$) and degrades almost linearly as $\sigma$ continues to increase.

\begin{figure}[ht]
    \centering
    \includegraphics[width=\textwidth]{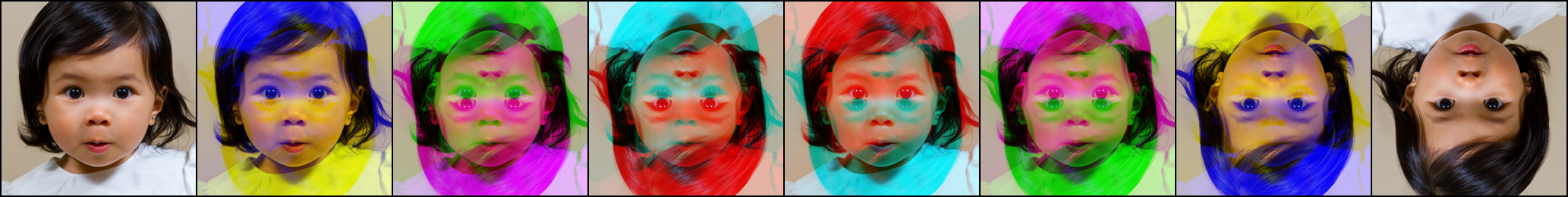}
    \caption{Eight images with exactly the \textbf{same measurement} for the phase retrieval task.}
    \label{fig:pr}
\end{figure}

\newpage
\subsection{More Discussion on Phase Retrieval}\label{sec:discuss-pr}

Compared to the baselines, DAPS exhibits significantly better sample quality and stability in the phase retrieval task. Unlike other selected tasks, phase retrieval is more ill-posed, meaning that images with the same measurements can appear quite different perceptually. Specifically, there are multiple disjoint modes with exactly the same measurement for phase retrieval, while for other tasks, such as super resolution and deblurring, the subset of images with low measurement error is a continuous set. We show in \cref{fig:pr} eight images with disparate perceptual features but with exactly the same measurement in phase retrieval.
To mitigate this issue, oversampling is often used to reduce the ill-posedness of the phase retrieval problem. We present quantitative results in \cref{tab:pr-over} using different oversampling ratios in phase retrieval. 
These results further demonstrate the strength of DAPS in addressing complex, ill-posed inverse problems. 

\begin{table}[ht]
\centering
\caption{Phase retrieval of \textbf{different oversampling ratios} with DAPS.}
\label{tab:pr-over}
\begin{tabular}{lccccc}
\toprule
Oversample  & 2.0 & 1.5& 1.0&0.5 & 0.0 \\
\midrule
LPIPS & 0.117 & 0.131 & 0.235 & 0.331 & 0.489 \\
PSNR & 30.26 & 29.17 & 24.87 & 21.60 & 16.02  \\
\bottomrule
\end{tabular}
\end{table}

\begin{figure}[ht]
    \centering
    \begin{subfigure}[b]{\textwidth}
    \includegraphics[width=\textwidth]{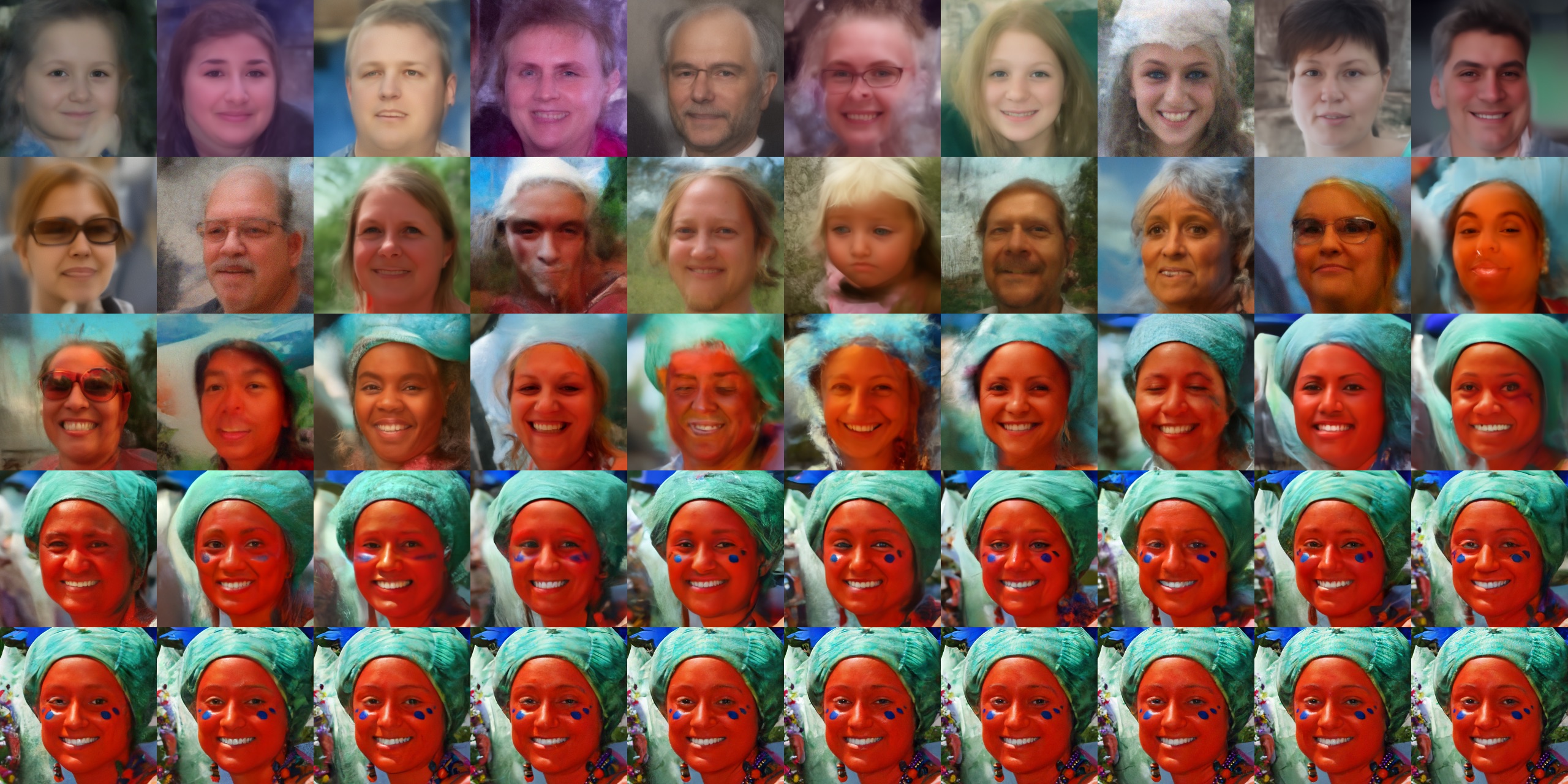}
    \caption{the estimated means of $p(\rvx_t\mid \rvx_0)$ as $\hat \rvx_0(\rvx_t)$}
    \end{subfigure}

    \begin{subfigure}[b]{\textwidth}
    \includegraphics[width=\textwidth]{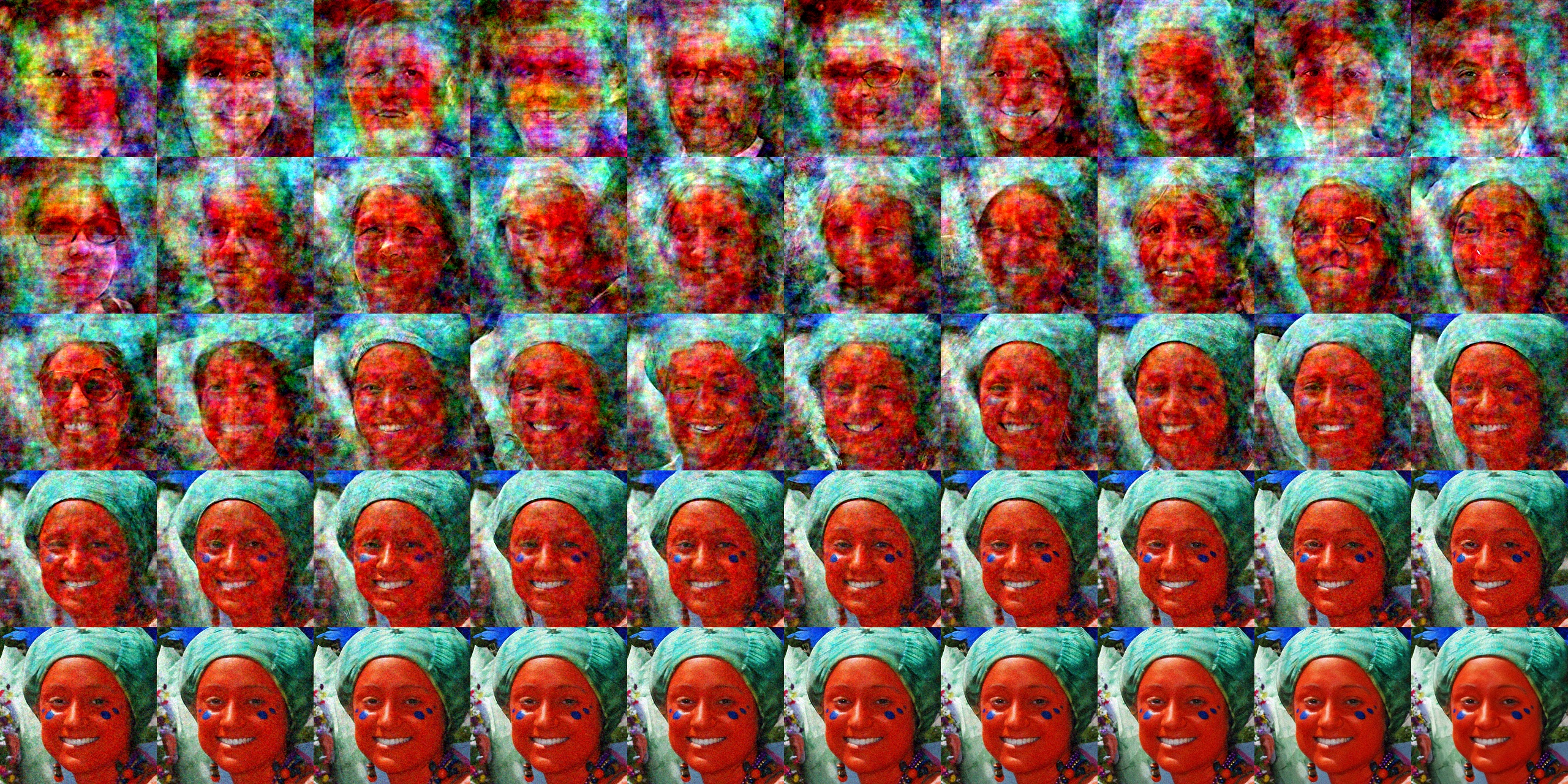}
     \caption{the samples $\rvx_{0\mid y}\sim p(\rvx_0\mid \rvx_t, \rvy)$}
    \end{subfigure}
    
    \caption{\textbf{DAPS trajectory for phase retrieval.} The images are selected from 200 annealing steps, evenly spaced in DAPS-1k configuration.}
    \label{fig:traj1}
\end{figure}

\subsection{More Analysis on Sampling Trajectory}
\label{appendix:traj}
\begin{figure}
    \centering
    \begin{subfigure}[b]{\textwidth}
    \includegraphics[width=\textwidth]{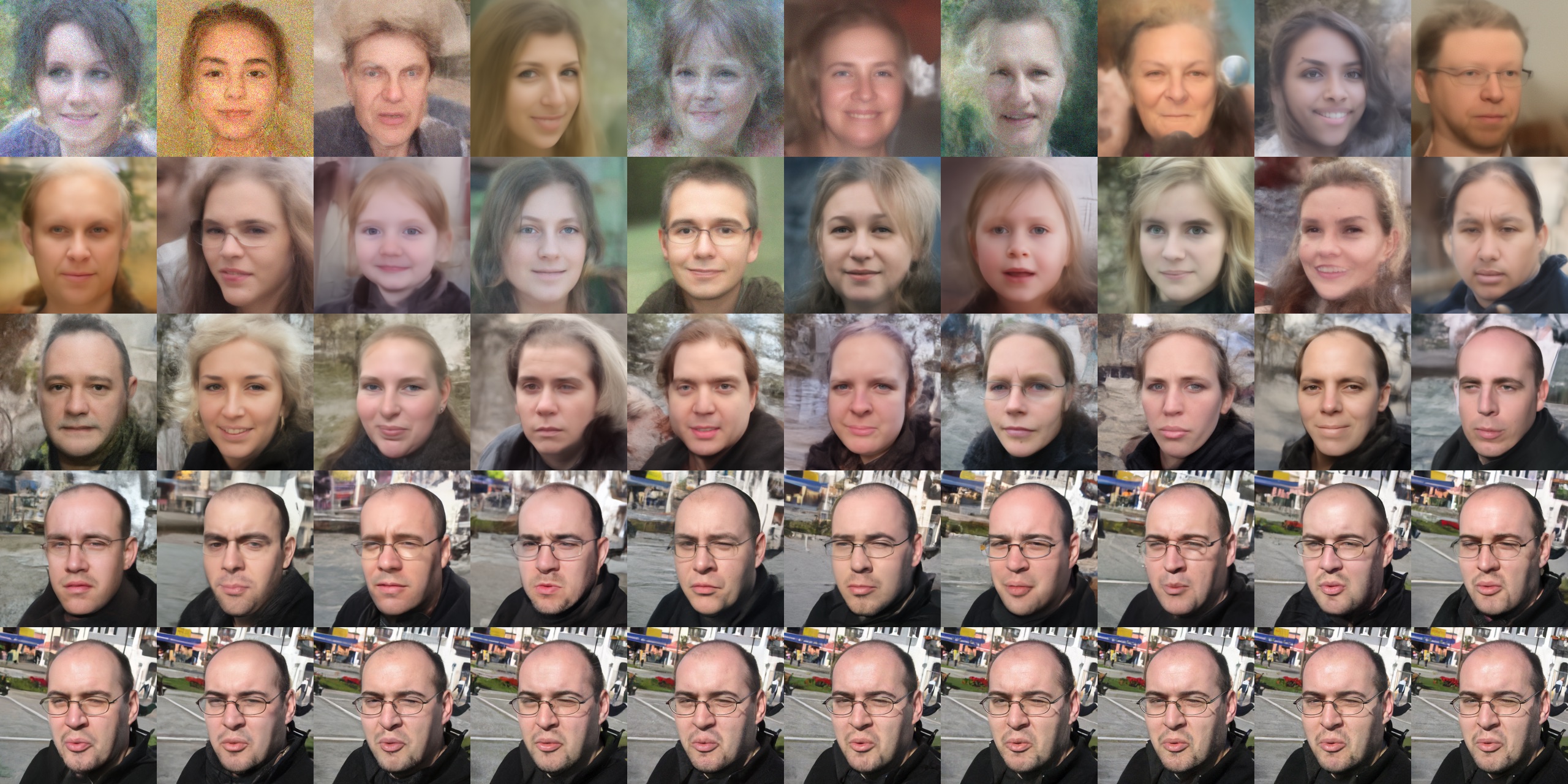}
    \caption{the estimated means of $p(\rvx_t\mid \rvx_0)$ as $\hat \rvx_0(\rvx_t)$}
    \end{subfigure}

    \begin{subfigure}[b]{\textwidth}
    \includegraphics[width=\textwidth]{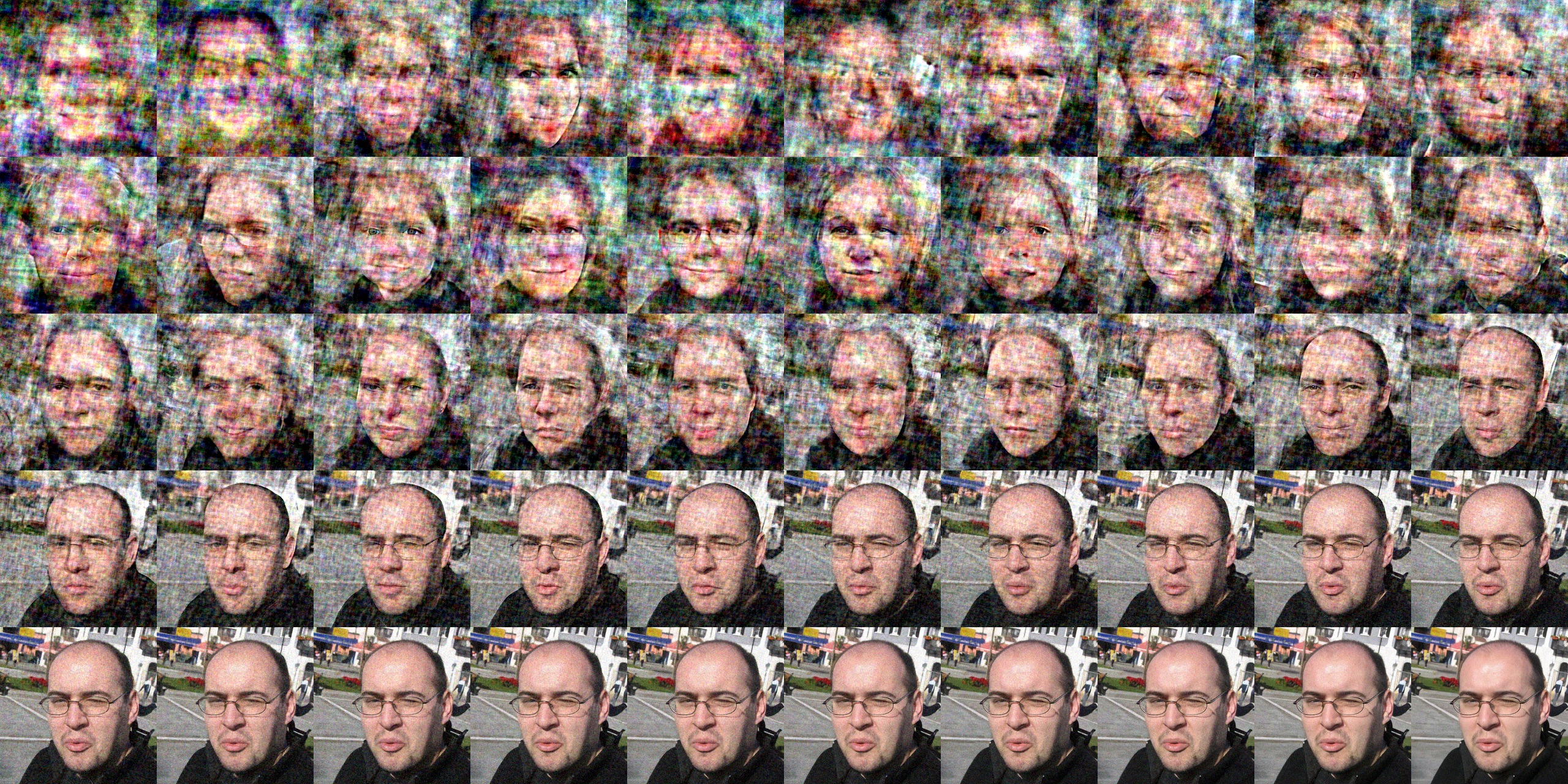}
     \caption{the samples $x_{0\mid y}\sim p(x_0\mid x_t, y)$}
    \end{subfigure}
    
    \caption{\textbf{DAPS trajectory for phase retrieval.} The images are selected from 200 annealing steps, evenly spaced in DAPS-1k configuration.}
    \label{fig:traj0}
\end{figure}

\begin{figure}
    \centering
    \begin{subfigure}[b]{\textwidth}
    \includegraphics[width=\textwidth]{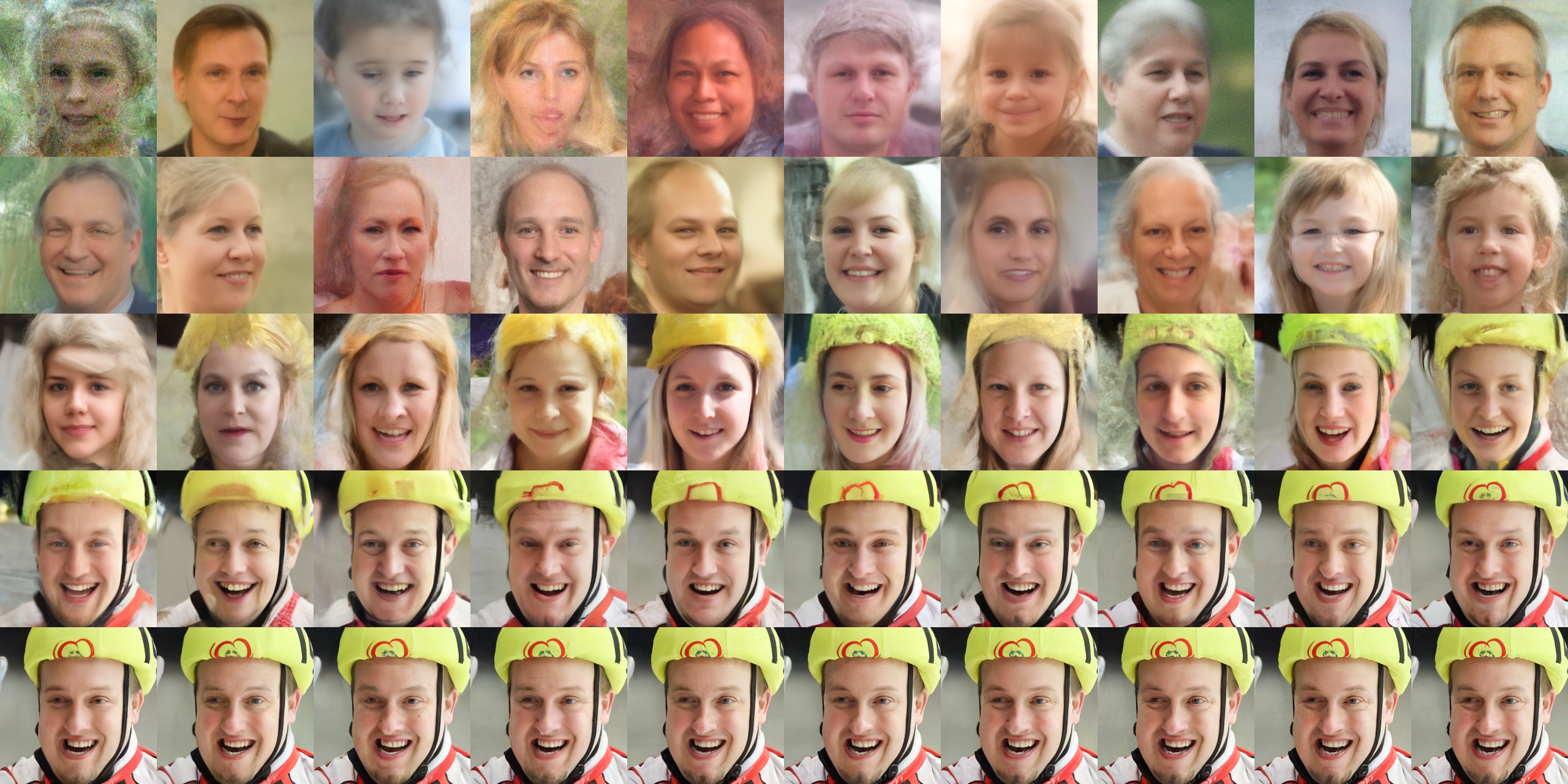}
    \caption{the estimated means of $p(\rvx_t\mid \rvx_0)$ as $\hat \rvx_0(\rvx_t)$}
    \end{subfigure}

    \begin{subfigure}[b]{\textwidth}
    \includegraphics[width=\textwidth]{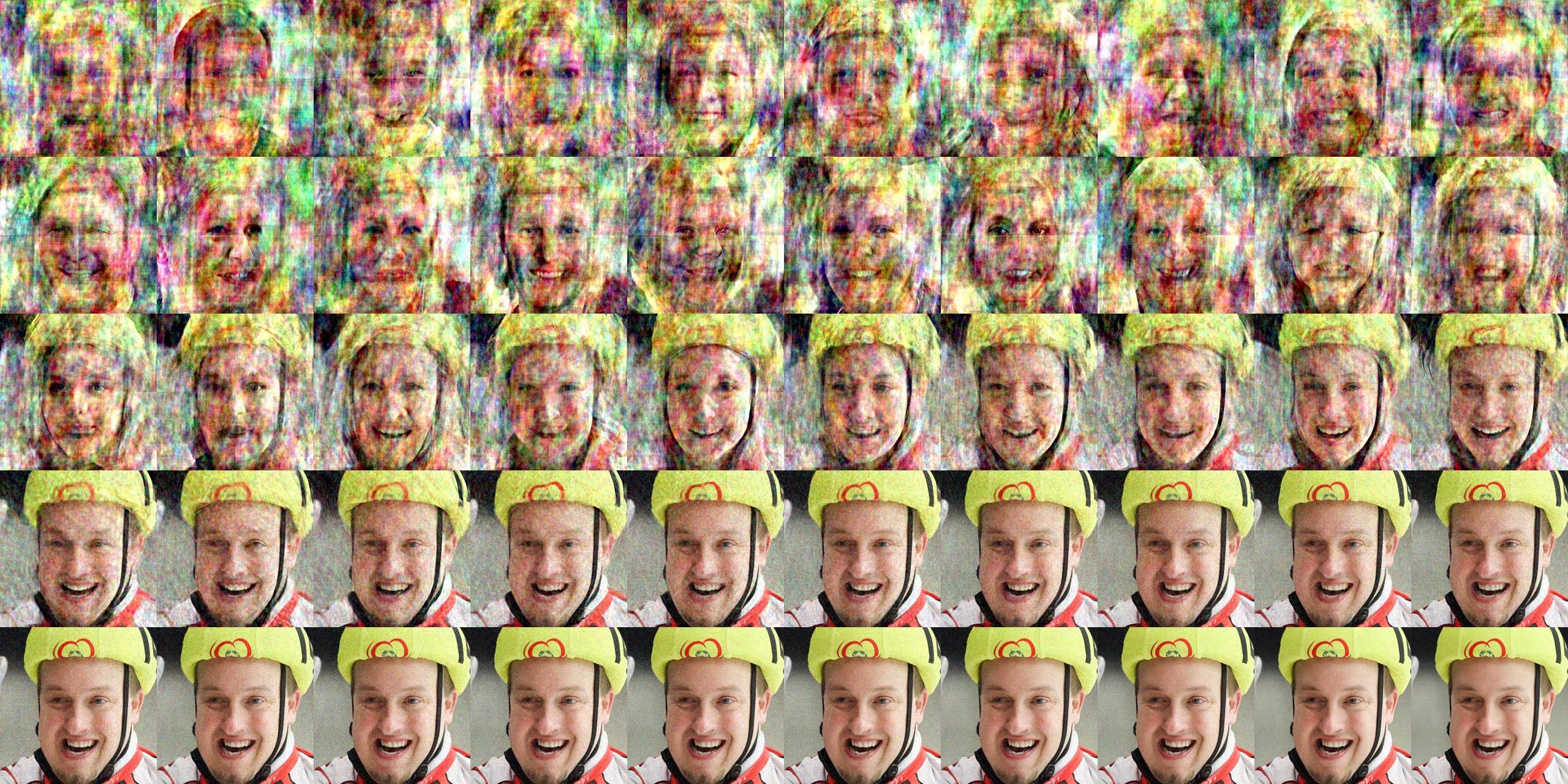}
     \caption{the samples $\rvx_{0\mid \rvy}\sim p(\rvx_0\mid \rvx_t, \rvy)$}
    \end{subfigure}
    
    \caption{\textbf{DAPS trajectory for phase retrieval.} The images are selected from 200 annealing steps, evenly spaced in DAPS-1k configuration.}
    \label{fig:traj2}
\end{figure}

Here we show a longer trajectory of phase retrieval in \cref{fig:traj0,fig:traj1,fig:traj2}. The $\hat \rvx_0 (\rvx_t)$ evolves from unconditional samples from the model to the posterior samples while $\rvx_{0\mid \rvy}$ evolves from noisy conditioned samples to the posterior samples. These two trajectories converge to the same sample as noise annealing down.

\subsection{More Qualitative Samples}
\label{appendix:qualitative}
\begin{figure}[t]
    \centering
\includegraphics[width=0.83\textwidth]{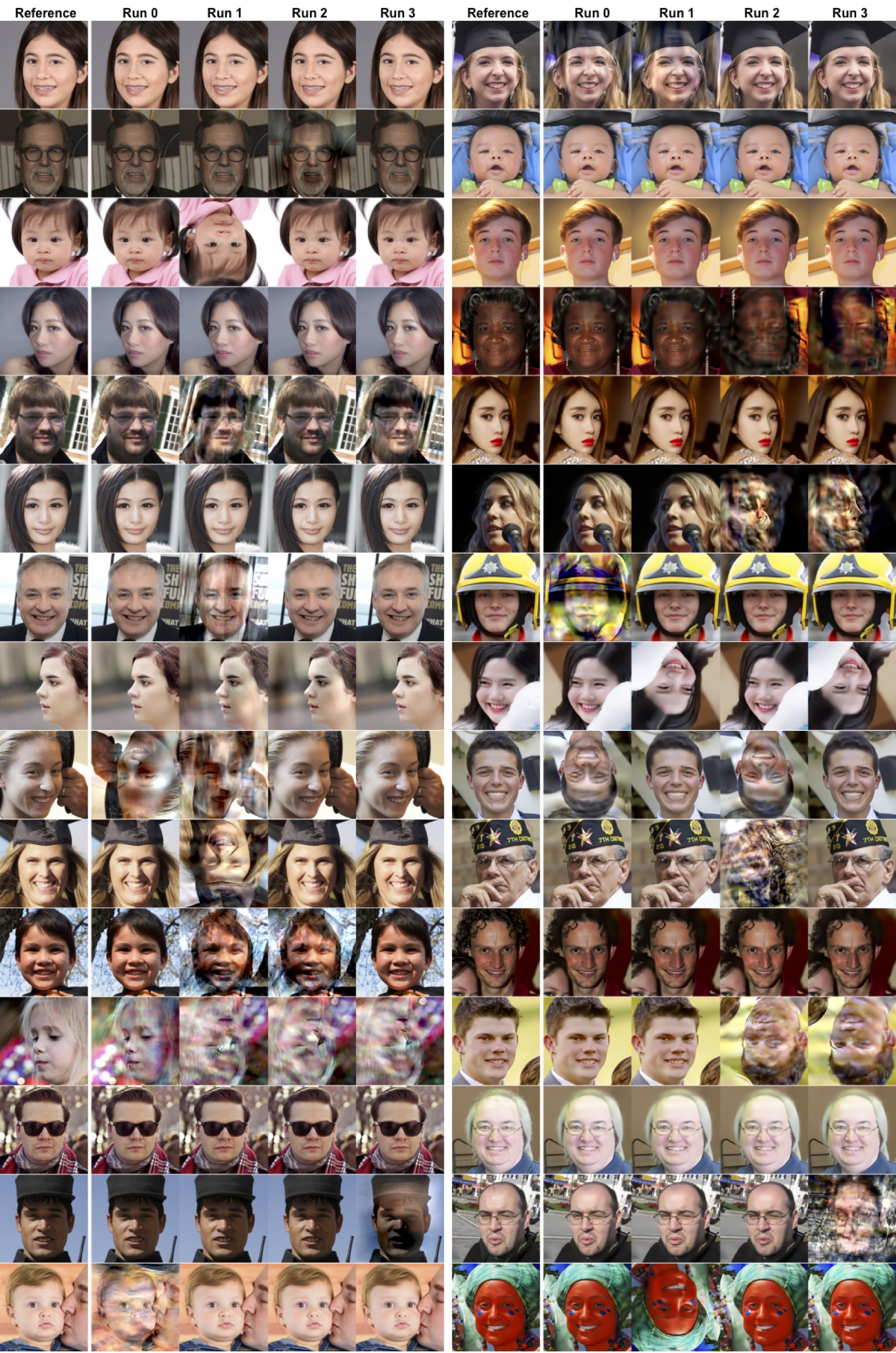}
    \caption{\textbf{Phase retrieval samples from DAPS (left) and LatentDAPS (right) in four independent runs on FFHQ.} No manual selection was performed to better visualize the success rate and sample quality.}
    \label{fig:app-pr}
\end{figure}

\begin{figure}[t]
    \centering
\includegraphics[width=0.87\textwidth]{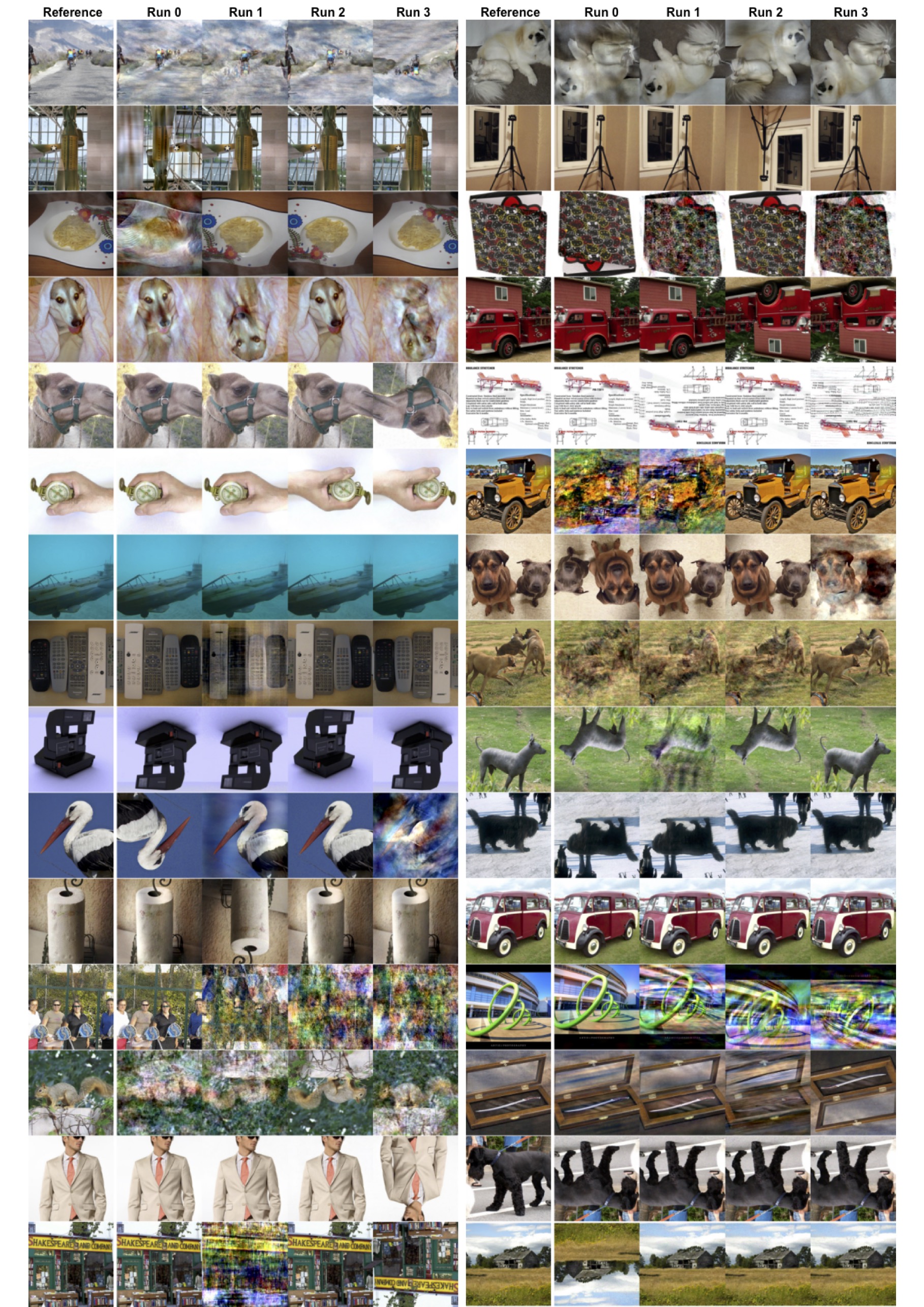}
    \caption{\textbf{Phase retrieval samples from DAPS in four independent runs on ImageNet.} The success rate of ImageNet is less than FFHQ and contains more samples with $180$ degree rotation. No manual selection was performed to better visual
    ize the success rate and sample quality.}
    \label{fig:app-pr-imagenet}
\end{figure}

\begin{figure}[t]
    \centering

    \begin{subfigure}[b]{0.37\textwidth}
        \centering
\includegraphics[width=\textwidth]{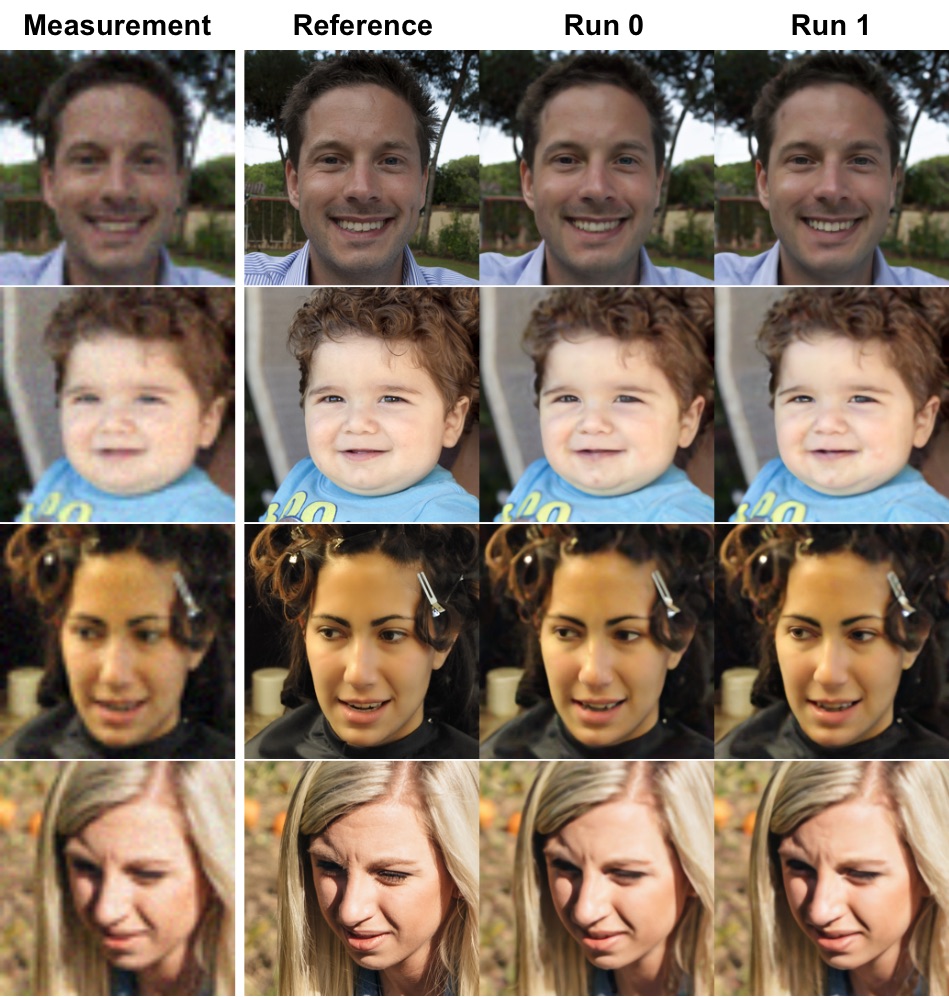}
        \caption{Super resolution 4×}
    \end{subfigure}
 \begin{subfigure}[b]{0.37\textwidth}
        \centering
\includegraphics[width=\textwidth]{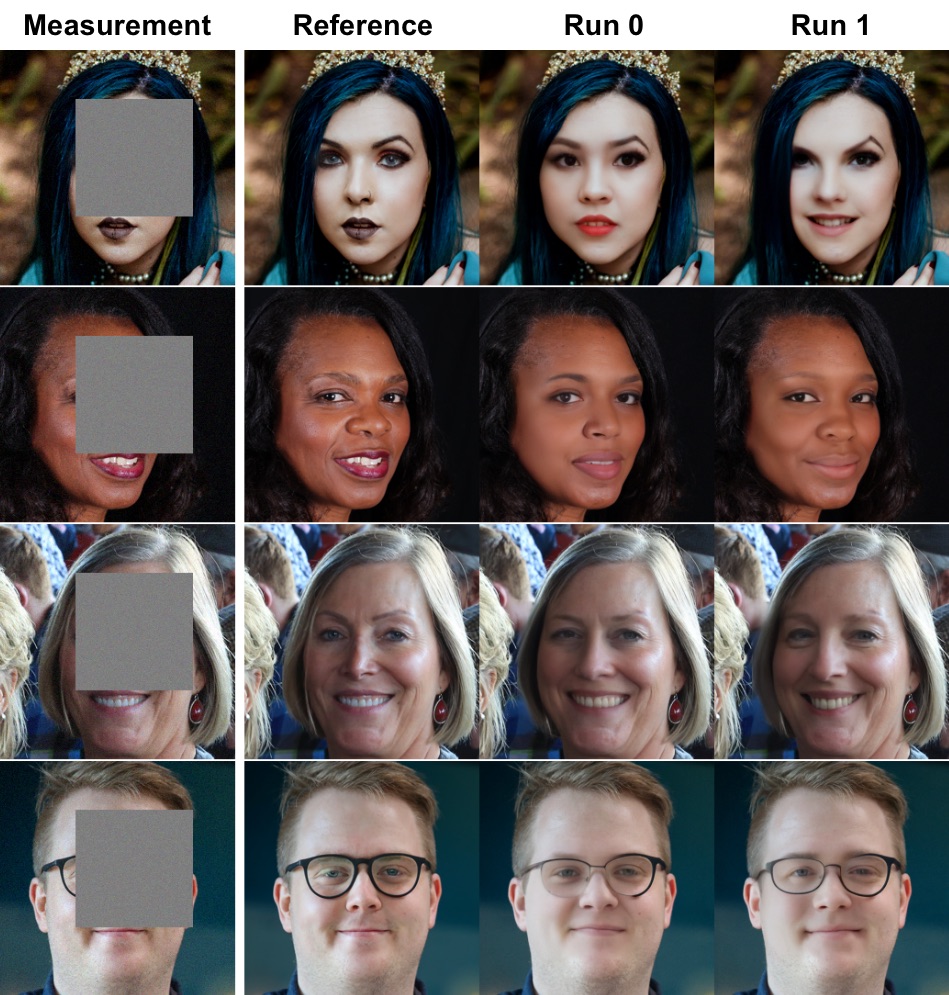}
        \caption{Inpaint (box)}
    \end{subfigure}\\
        \begin{subfigure}[b]{0.37\textwidth}
        \centering
\includegraphics[width=\textwidth]{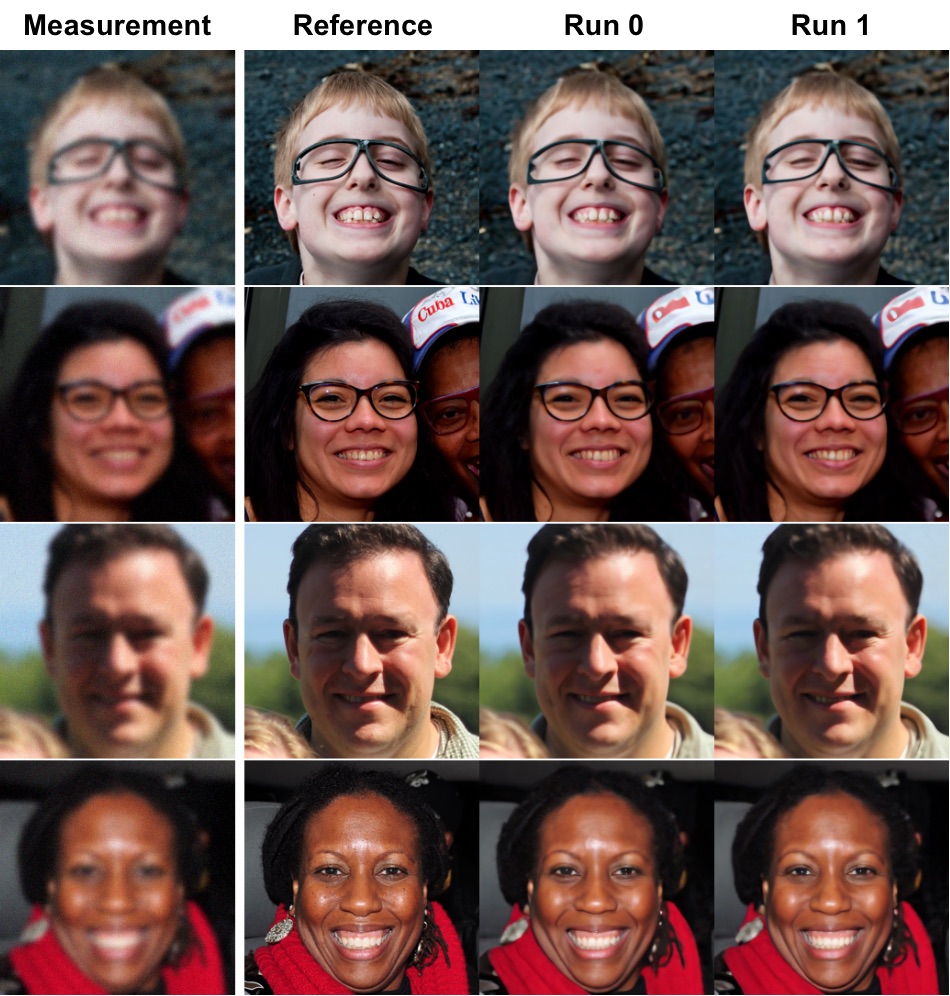}
        \caption{Gaussian deblurring}
    \end{subfigure}
 \begin{subfigure}[b]{0.37\textwidth}
        \centering
\includegraphics[width=\textwidth]{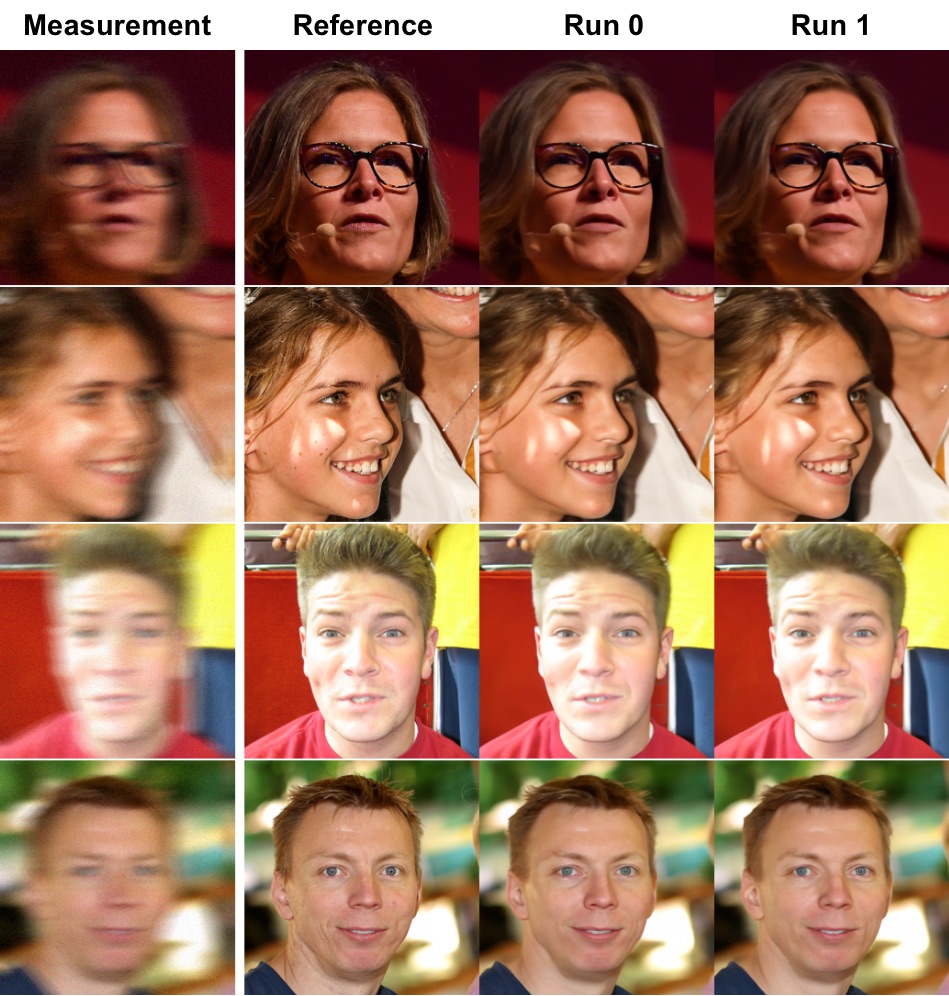}
        \caption{Motion deblurring}
    \end{subfigure}\\
        \begin{subfigure}[b]{0.37\textwidth}
        \centering
\includegraphics[width=\textwidth]{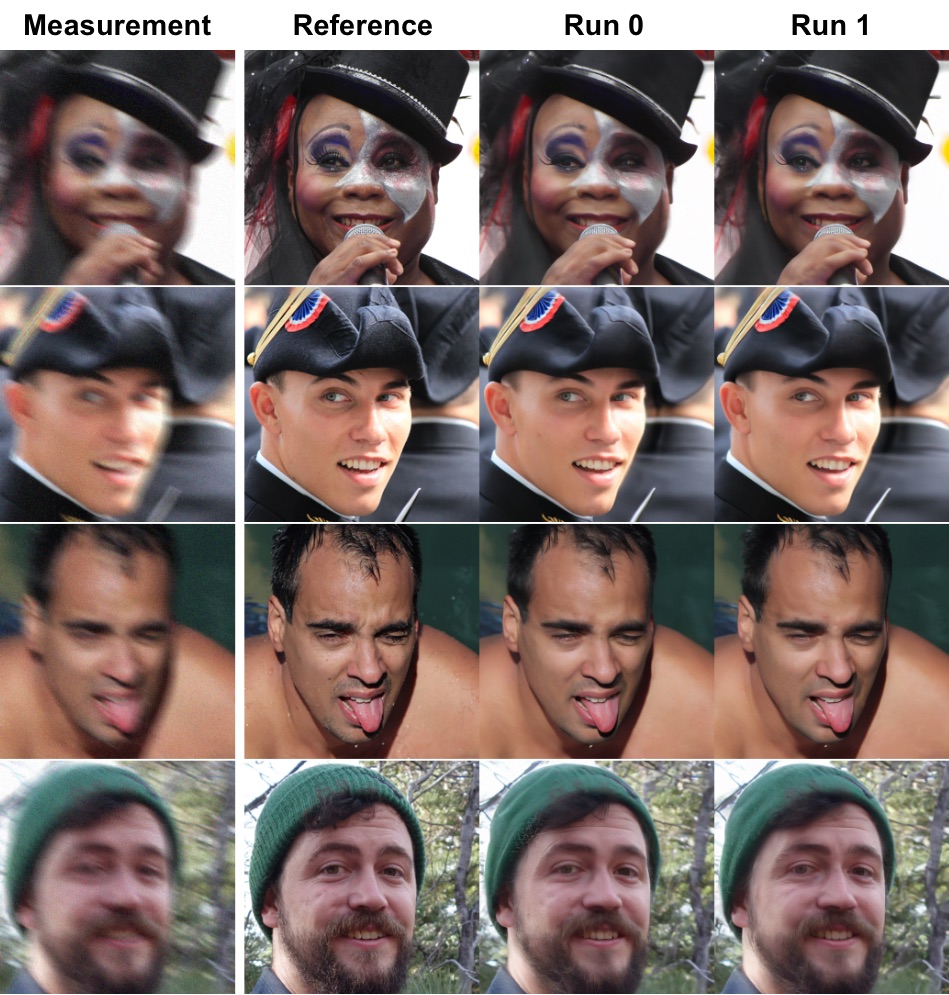}
        \caption{Nonlinear deblurring}
    \end{subfigure}
 \begin{subfigure}[b]{0.37\textwidth}
        \centering
\includegraphics[width=\textwidth]{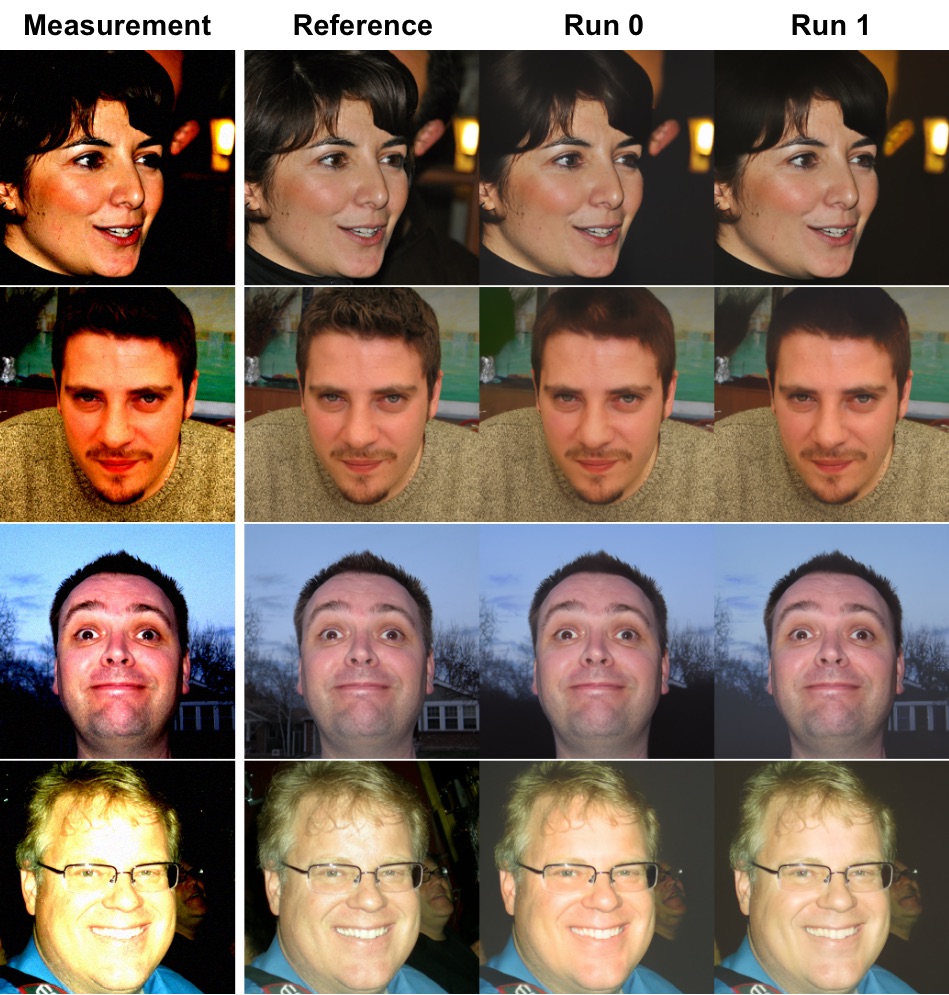}
        \caption{High Dynamic Range}
    \end{subfigure}\\
    \caption{\textbf{DAPS samples for various tasks on FFHQ.} DAPS can obtain visually better samples for the above linear and nonlinear tasks.}
    \label{fig:app-others}
\end{figure}

\begin{figure}[t]
    \centering

    \begin{subfigure}[b]{0.37\textwidth}
        \centering
\includegraphics[width=\textwidth]{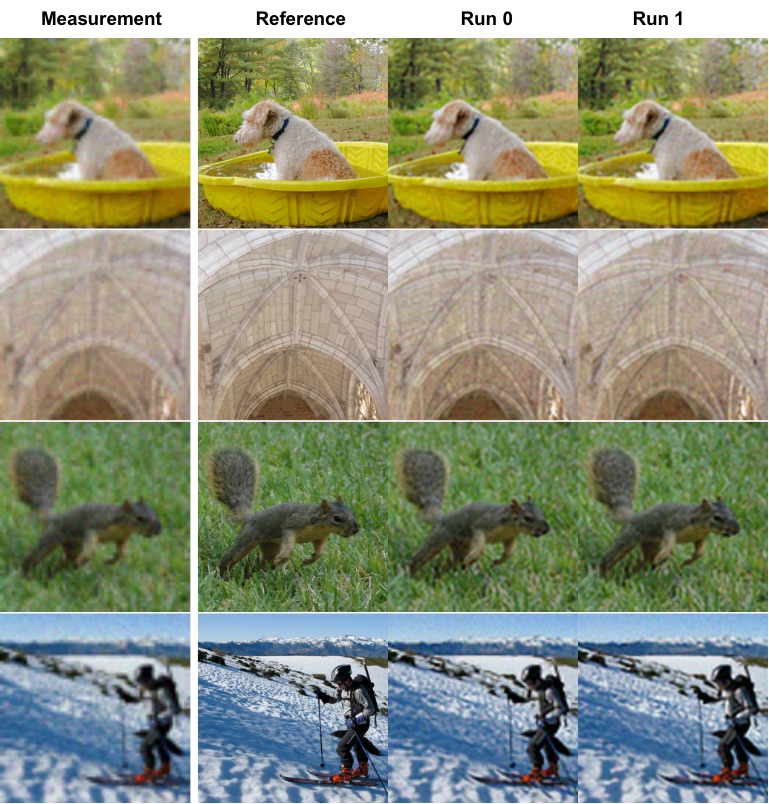}
        \caption{Super resolution 4×}
    \end{subfigure}
 \begin{subfigure}[b]{0.37\textwidth}
        \centering
\includegraphics[width=\textwidth]{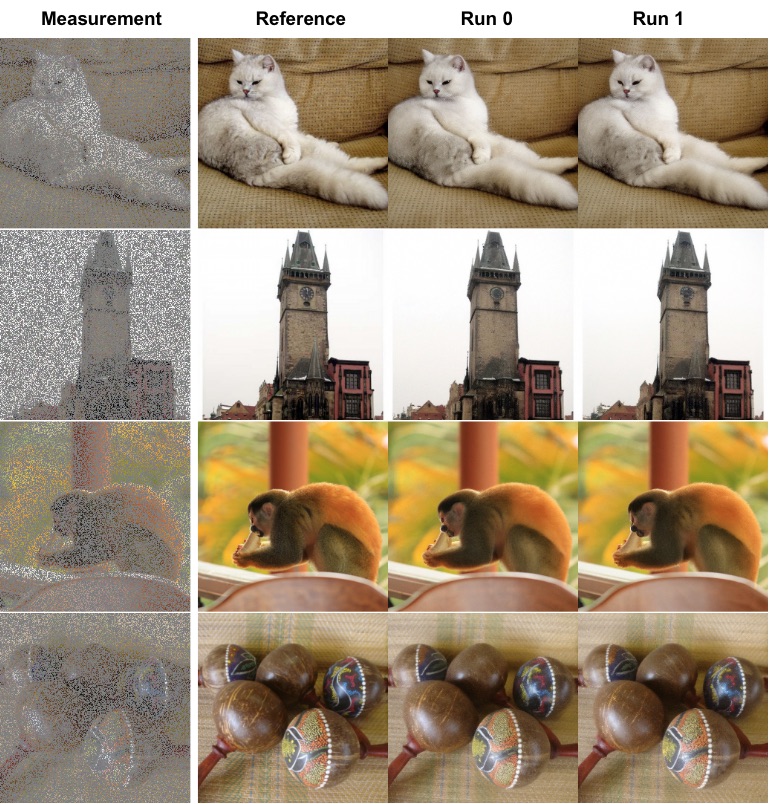}
        \caption{Inpaint (box)}
    \end{subfigure}\\
        \begin{subfigure}[b]{0.37\textwidth}
        \centering
\includegraphics[width=\textwidth]{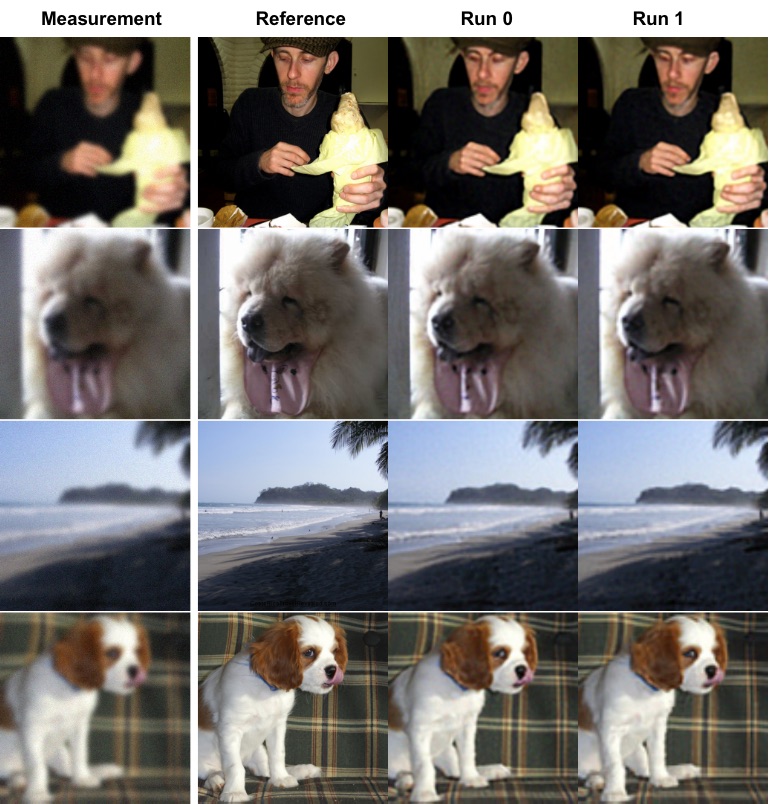}
        \caption{Gaussian deblurring}
    \end{subfigure}
 \begin{subfigure}[b]{0.37\textwidth}
        \centering
\includegraphics[width=\textwidth]{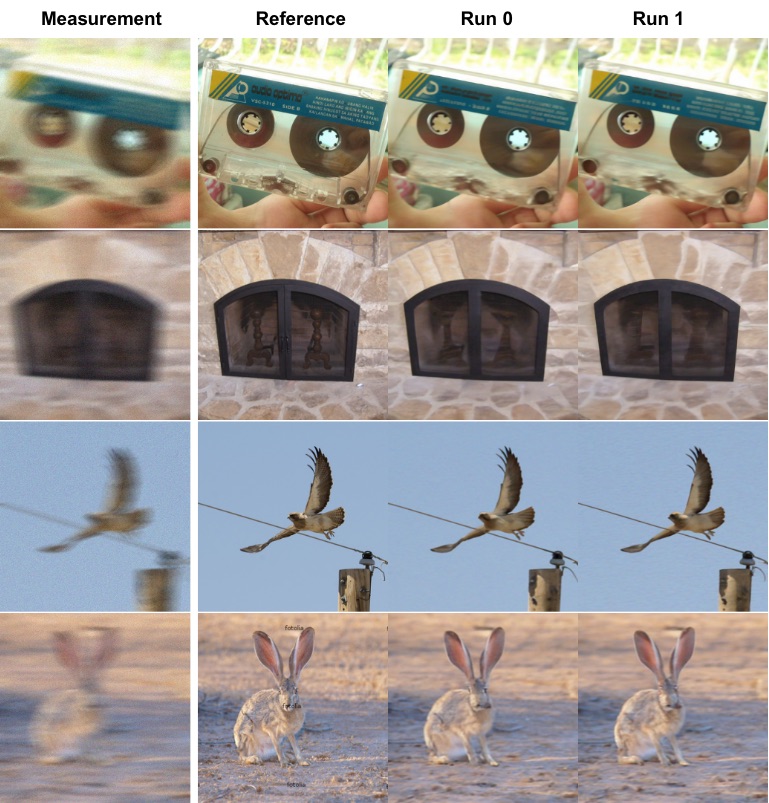}
        \caption{Motion deblurring}
    \end{subfigure}\\
        \begin{subfigure}[b]{0.37\textwidth}
        \centering
\includegraphics[width=\textwidth]{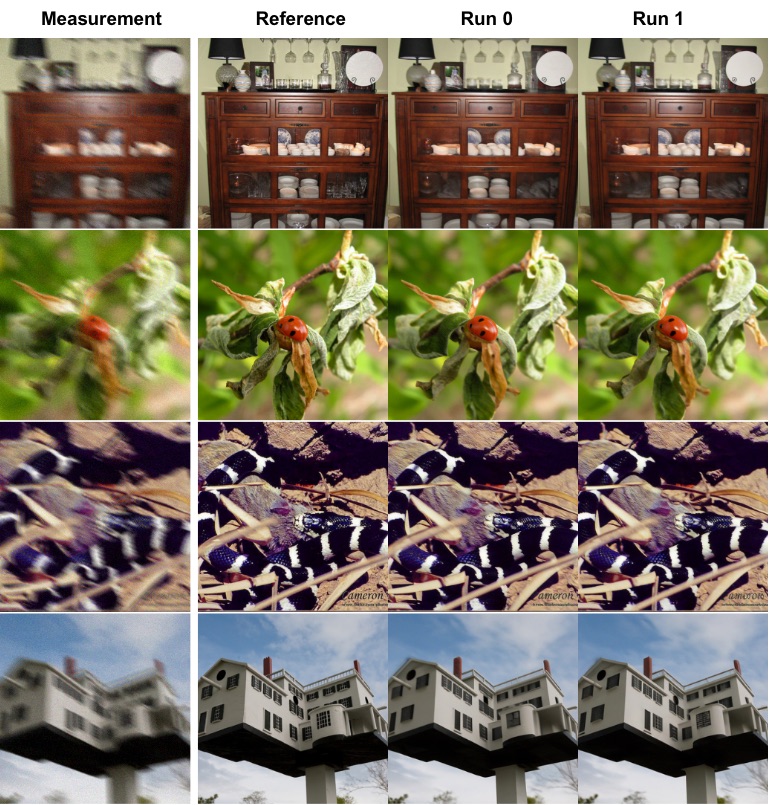}
        \caption{Nonlinear deblurring}
    \end{subfigure}
 \begin{subfigure}[b]{0.37\textwidth}
        \centering
\includegraphics[width=\textwidth]{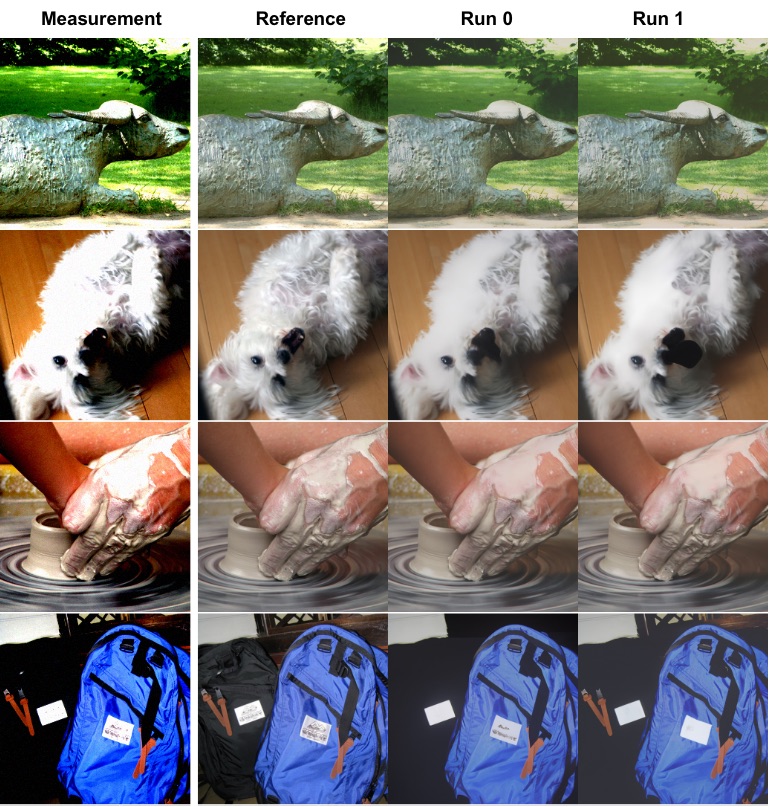}
        \caption{High Dynamic Range}
    \end{subfigure}\\
    \caption{\textbf{DAPS samples for various tasks on ImageNet.} DAPS can obtain visually better samples for the above linear and nonlinear tasks.}
    \label{fig:app-others-imagenet}
\end{figure}

\begin{figure}
    \centering
\includegraphics[width=0.73\textwidth]{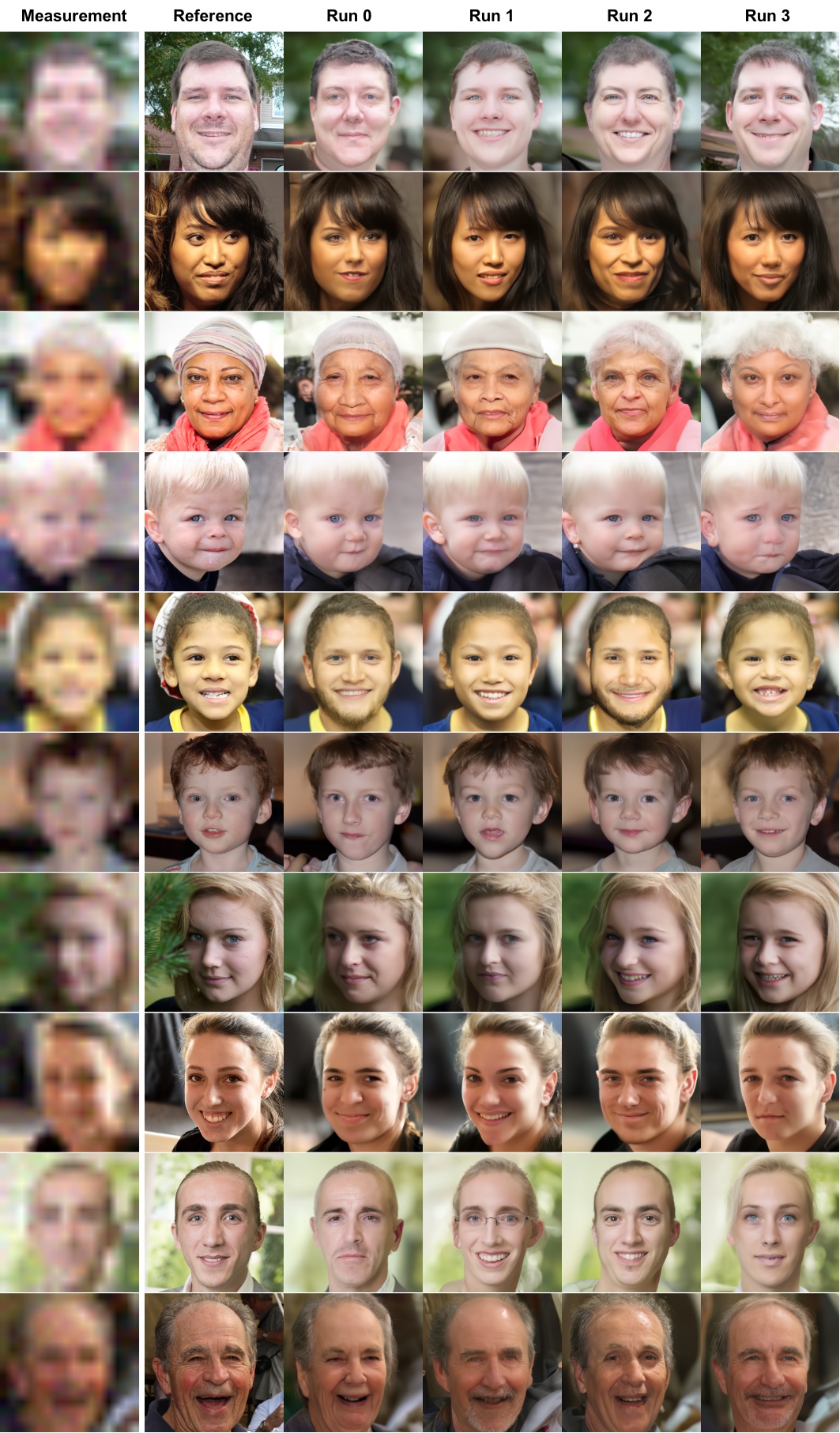}
    \caption{\textbf{More samples for super resolution 16×.} DAPS is able to generate diverse samples when the posterior distribution is multi-modal.}
    \label{fig:app-div-sr}
\end{figure}

\begin{figure}
    \centering
\includegraphics[width=0.73\textwidth]{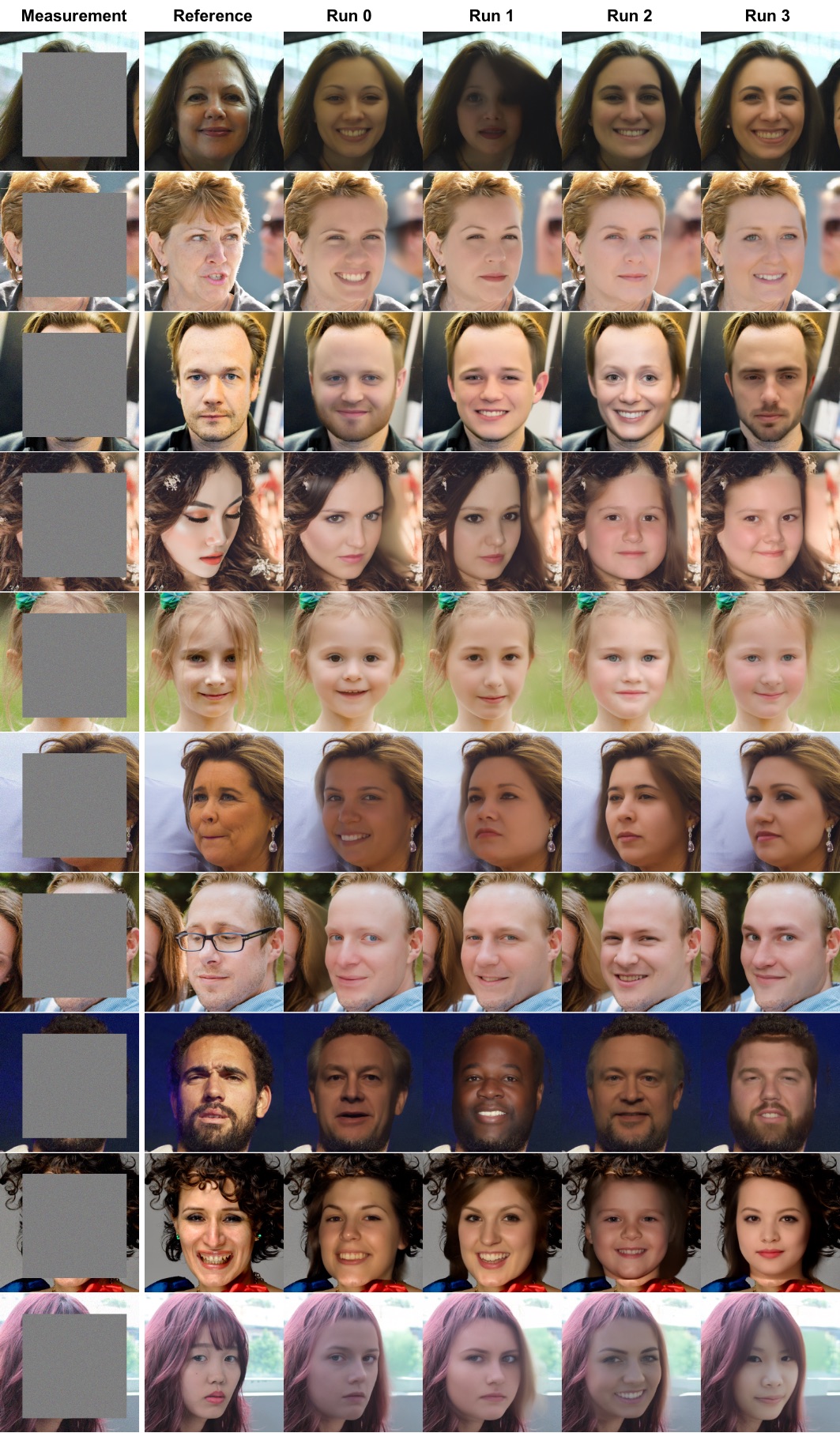}
    \caption{\textbf{More samples for inpainting of 192$\mathbf\times$192 box.} DAPS is able to generate diverse samples when the posterior distribution is multi-modal.}
    \label{fig:app-div-box}
\end{figure}

\begin{figure}
    \centering
    \includegraphics[width=.8\linewidth]{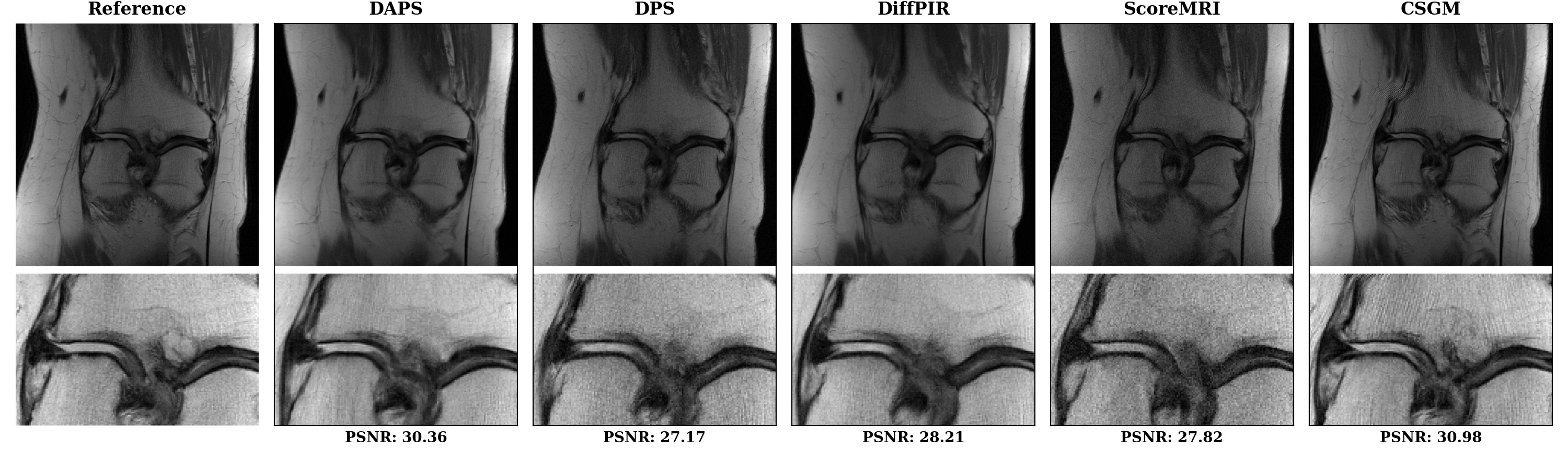}

    \includegraphics[width=.8\linewidth]{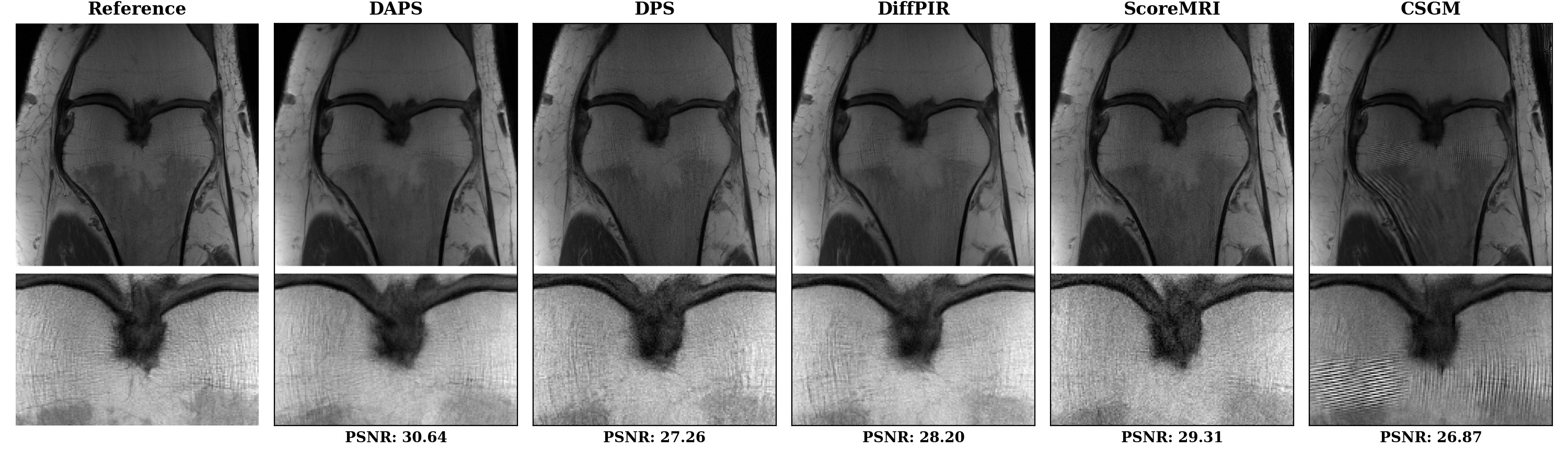}

    \includegraphics[width=.8\linewidth]{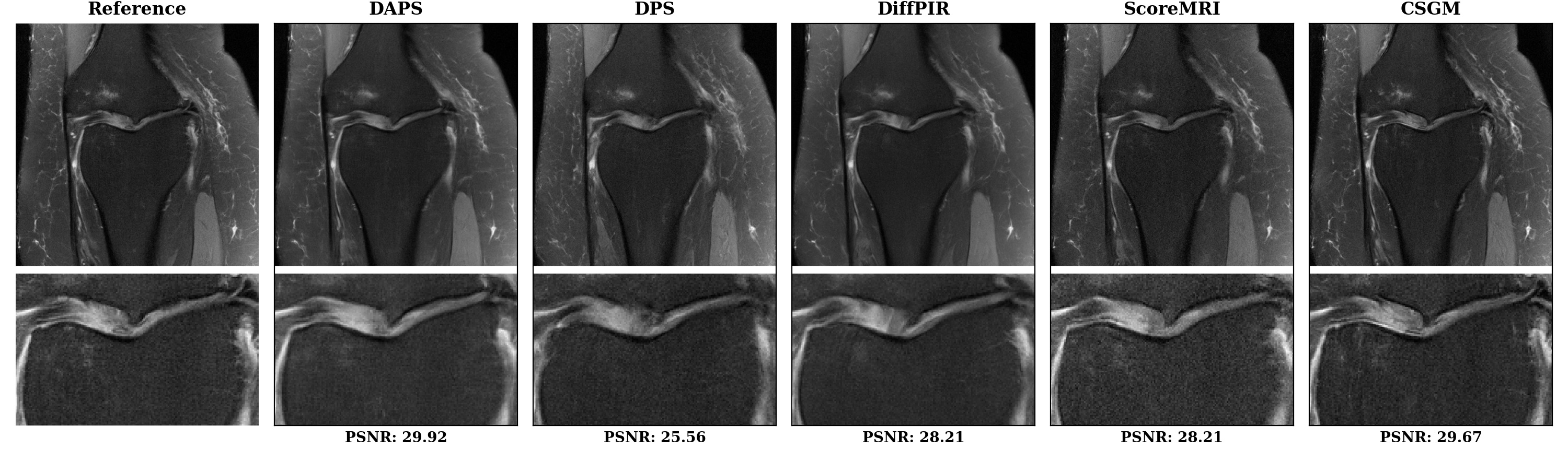}

    \includegraphics[width=.8\linewidth]{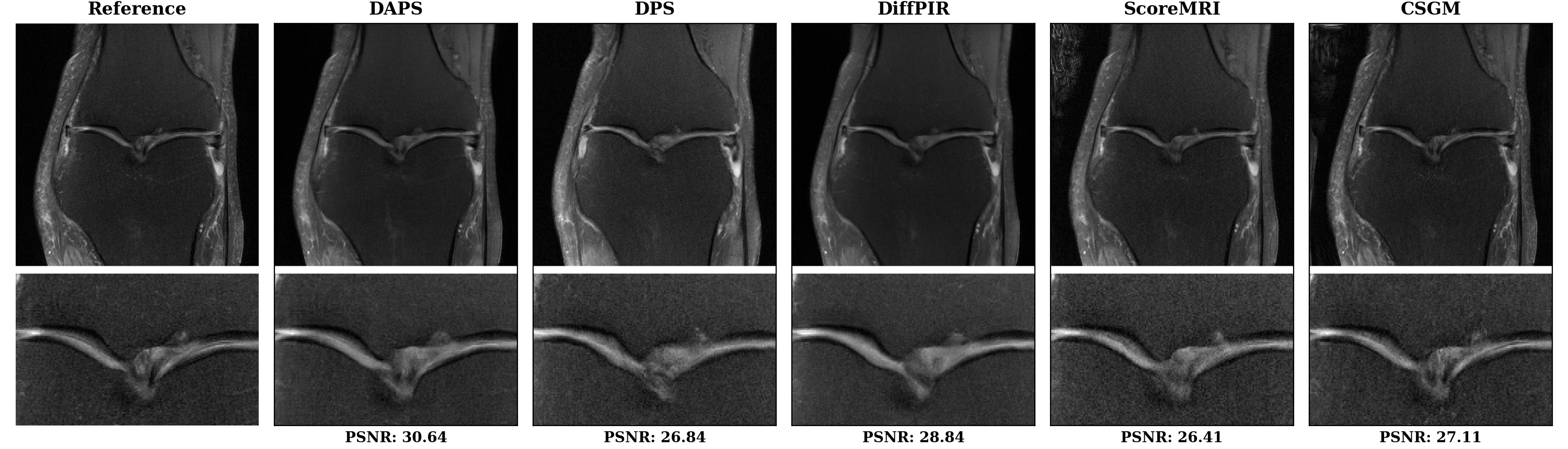}
    \includegraphics[width=.8\linewidth]{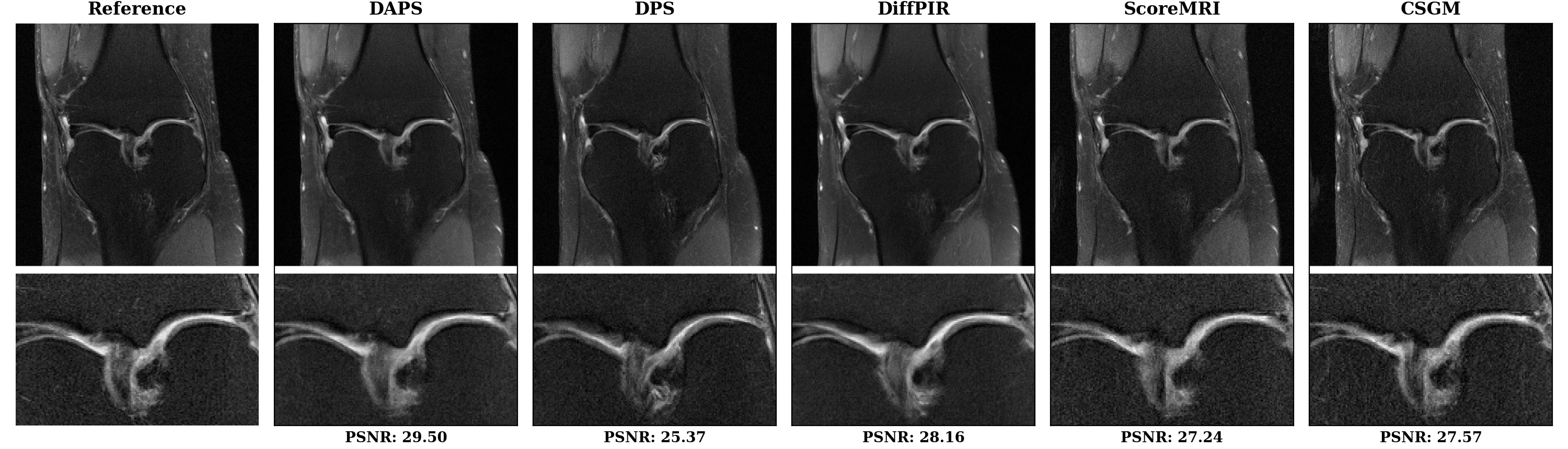}
    \caption{\textbf{More samples for CS-MRI.} DAPS is able to generate high-fidelity reconstructions for CS-MRI tasks.}
    \label{fig:mri_all}
\end{figure}

We show a full stack of phase retrieval samples in $4$ runs without manual post-selection in \cref{fig:app-pr,fig:app-pr-imagenet}. More samples for other tasks are shown in \cref{fig:app-others,fig:app-others-imagenet}. The more diverse samples from box inpainting of size $192\times192$ and super resolution of factor $16$ are shown in \cref{fig:app-div-box,fig:app-div-sr}. More samples for CS-MRI are shown in~\cref{fig:mri_all}.

\end{document}